\PassOptionsToPackage{table}{xcolor}
\documentclass{article}
\usepackage{iclr2027_conference,times}

\usepackage{hyperref}
\usepackage{url}
\usepackage{amsmath,amssymb}
\usepackage{booktabs}
\usepackage{colortbl}
\usepackage{enumitem}
\usepackage{graphicx}
\usepackage{xcolor}
\usepackage[most]{tcolorbox}
\usepackage{pifont}
\usepackage{marvosym}
\definecolor{citegreen}{HTML}{486ED7}
\hypersetup{
  colorlinks=true,
  citecolor=citegreen,
  linkcolor=red,
  urlcolor=citegreen
}
\usepackage{tikz}
\usepackage{wrapfig}
\usetikzlibrary{positioning,arrows.meta,shapes.geometric,fit,backgrounds,calc,patterns}

\newcommand{\cmark}{\ding{51}}\newcommand{\xmark}{\ding{55}}
\newcommand{\bmark}{\textcolor{gray!50}{\xmark}}

\title{Sensitivity or Criterion? Measuring and \\ Moving When Language Models Act}

\author{
\textbf{Sixun Dong}\textsuperscript{1}\hspace{0.4cm}
\textbf{Yebowen Hu}\textsuperscript{2}\hspace{0.4cm}
\textbf{Ming Yin}\textsuperscript{3}\hspace{0.4cm}
\textbf{Yiran Chen}\textsuperscript{3}\hspace{0.4cm}
\textbf{Fei Liu}\textsuperscript{4}\hspace{0.4cm}
\textbf{Chen Chen}\textsuperscript{1}$^\text{\Letter}$ \\
\textsuperscript{1}University of Central Florida\hspace{0.2cm}
\textsuperscript{2}Zoom Communications\hspace{0.2cm}
\textsuperscript{3}Duke University\hspace{0.2cm}
\textsuperscript{4}Emory University \\
{\footnotesize \texttt{sixundong@ucf.edu, yebowen.hu@zoom.us, chen.chen@crcv.ucf.edu}}
}

\iclrfinalcopy

\begin{document}

\maketitle
\lhead{Preprint.}
\renewcommand\thefootnote{}
\footnotetext{\Letter~Corresponding author.}
\renewcommand\thefootnote{\arabic{footnote}}

\begin{abstract}
Before an agent acts it must decide whether any action is warranted. Benchmarks that ask this question
report one accuracy, which cannot distinguish a model that fails to recognize a warranted action from one
that recognizes it but is reluctant to act. Signal-detection theory separates the two as sensitivity and
criterion, but a criterion estimated from authored binary labels depends on each benchmark's mix of
examples and cannot be compared across benchmarks. We introduce \textsc{FlipAct}, a diagnostic
built on a qualitative representation of what tools do to physical states. A rule over annotated tool
properties computes each question's answer and a graded evidence score, with the Yes/No boundary fixed
at zero for every model, so every answer is computed, not written by a model. Questions are built from $2\times2$ quads that hold the required
knowledge fixed while the state flips. The slope of a model's Yes-rate against the score is its
sensitivity, and the score where it crosses one half is its criterion. Across $44$ open models, sensitivity rises
with size within a family, and its ranking agrees with three of four external when-to-act
benchmarks. Models respond to the sign of the evidence rather than its size, accepting actions that
leave the state no better about as often as helpful ones. When Yes means act, most models are conservative and stay so when the question is
rephrased. In some models this is a preference for answering No, and the criterion is not set by size. Activation
steering moves the criterion with the weights fixed, and distilling a model from its own steered copy
moves it in the weights, without a significant change in sensitivity. The distilled shift carries to six
external benchmarks with sensitivity kept, including two without a Yes/No answer. Accuracy rises where
the model started conservative and falls where it started permissive.
Project page: \url{https://sixundong.com/projects/flipact}.
\end{abstract}

\section{Introduction}
\label{sec:intro}

Large language model (LLM) agents must decide whether to act before they decide how. Benchmarks for
agents mostly score the action once taken: whether the chosen function, its arguments and the
trajectory are correct \citep{patil2025bfcl,huang2024metatool,wang2025mllm,yin2026livemcp101stresstestingdiagnosing}. Those that study the decision itself, in
tool calling \citep{ross2025when2call,when2tool2026}, clarification \citep{questbench2025} and abstention
\citep{abstentionbench2025}, report a single accuracy, and that number mixes two ways to fail. A model
may not recognize whether a situation warrants action, a failure of its \emph{sensitivity}; or it may
recognize the situation and still be too reluctant or too ready to act, a failure of its
\emph{criterion}. Too conservative a criterion refuses benign requests
\citep{rottger2024xstest,cui2024orbench,shi2024overkill} or answers from memory where a tool was needed;
too permissive a criterion acts where nothing warrants it.

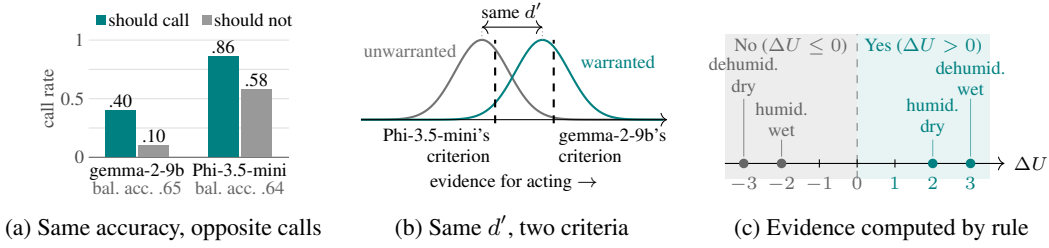
\begin{figure}[t]
\centering
\begin{minipage}[b]{0.30\linewidth}
\centering
\begin{tikzpicture}[x=1cm,y=1.55cm,font=\scriptsize]
  \foreach \y in {0.25,0.5,0.75,1.0} { \draw[black!12,thin] (0,\y) -- (2.65,\y); }
  \draw[black!65] (0,0) -- (2.65,0);
  \foreach \y/\l in {0/0,0.5/0.5,1.0/1} { \node[left,font=\scriptsize,black!70,inner sep=1.5pt] at (0,\y) {\l}; }
  \node[rotate=90,font=\scriptsize,black!70] at (-0.55,0.5) {call rate};
  \fill[teal] (0.22,0) rectangle (0.62,0.40);
  \fill[black!40]        (0.66,0) rectangle (1.06,0.10);
  \fill[teal] (1.58,0) rectangle (1.98,0.86);
  \fill[black!40]        (2.02,0) rectangle (2.42,0.58);
  \foreach \x/\v/\t in {0.42/0.40/.40, 0.86/0.10/.10, 1.78/0.86/.86, 2.22/0.58/.58} {
    \node[above,font=\scriptsize,inner sep=1pt] at (\x,\v) {\t}; }
  \node[below,font=\scriptsize,align=center,inner sep=2pt] at (0.64,0) {gemma-2-9b\\[-1pt]{\color{black!60}bal.\ acc.\ .65}};
  \node[below,font=\scriptsize,align=center,inner sep=2pt] at (2.00,0) {Phi-3.5-mini\\[-1pt]{\color{black!60}bal.\ acc.\ .64}};
  \fill[teal] (0.10,1.13) rectangle (0.22,1.21); \node[right,font=\scriptsize,inner sep=1pt] at (0.22,1.17) {should call};
  \fill[black!40]        (1.50,1.13) rectangle (1.62,1.21); \node[right,font=\scriptsize,inner sep=1pt] at (1.62,1.17) {should not};
\end{tikzpicture}\\[2pt]
{\footnotesize (a) Same accuracy, opposite calls}
\end{minipage}\hfill
\begin{minipage}[b]{0.34\linewidth}
\centering
\begin{tikzpicture}[x=0.5cm,y=0.48cm,font=\scriptsize]
  \def\dsep{1.6}
  \draw[thick,black!55] plot[domain=-3.2:4.8,samples=100] (\x,{2.2*exp(-(\x)^2/1.0)});
  \draw[thick,teal] plot[domain=-3.2:4.8,samples=100] (\x,{2.2*exp(-(\x-\dsep)^2/1.0)});
  \draw[->] (-3.2,0) -- (4.9,0);
  \node[black!55,font=\scriptsize] at (-1.9,1.9) {unwarranted};
  \node[teal,font=\scriptsize] at (3.6,1.6) {warranted};
  \draw[dashed,thick] (0.35,-0.1) -- (0.35,2.2);
  \draw[dashed,thick] (1.9,-0.1) -- (1.9,2.2);
  \node[anchor=north east,font=\scriptsize,inner sep=1pt,align=right] at (0.3,-0.15) {Phi-3.5-mini's\\criterion};
  \node[anchor=north west,font=\scriptsize,inner sep=1pt,align=left] at (1.95,-0.15) {gemma-2-9b's\\criterion};
  \node[below,font=\scriptsize] at (0.85,-1.25) {evidence for acting $\rightarrow$};
  \draw[black!45,densely dotted] (0,2.2) -- (0,2.55); \draw[black!45,densely dotted] (\dsep,2.2) -- (\dsep,2.55);
  \draw[<->] (0,2.55) -- (\dsep,2.55) node[midway,above,font=\scriptsize] {same $d'$};
\end{tikzpicture}\\[2pt]
{\footnotesize (b) Same $d'$, two criteria}
\end{minipage}\hfill
\begin{minipage}[b]{0.34\linewidth}
\centering
\begin{tikzpicture}[x=0.5cm,y=0.45cm,font=\scriptsize]
  \fill[black!7] (-3.5,-0.45) rectangle (0,3.8);
  \fill[teal!9]  (0,-0.45) rectangle (3.5,3.8);
  \draw[dashed,black!50] (0,-0.45) -- (0,3.8);
  \node[black!60,font=\scriptsize] at (-1.75,3.45) {No ($\Delta U\le0$)};
  \node[teal,font=\scriptsize] at (1.75,3.45) {Yes ($\Delta U>0$)};
  \draw[->] (-3.5,0) -- (3.9,0) node[right,font=\scriptsize] {$\Delta U$};
  \foreach \x in {-3,...,3} { \draw (\x,0.12) -- (\x,-0.12); }
  \foreach \x in {-3,-2,-1,0} { \node[below,font=\scriptsize,black!60] at (\x,-0.12) {$\x$}; }
  \foreach \x in {1,2,3} { \node[below,font=\scriptsize,teal] at (\x,-0.12) {$\x$}; }
  \draw[black!45,thin] (-3,0.2) -- (-3,1.9); \node[anchor=south,font=\scriptsize,black!60,inner sep=1pt,align=center] at (-3,1.9) {dehumid.\\dry};
  \draw[black!45,thin] (-2,0.2) -- (-2,0.7); \node[anchor=south,font=\scriptsize,black!60,inner sep=1pt,align=center] at (-2,0.7) {humid.\\wet};
  \draw[teal!60,thin]  (2,0.2)  -- (2,0.7);  \node[anchor=south,font=\scriptsize,teal,inner sep=1pt,align=center] at (2,0.7) {humid.\\dry};
  \draw[teal!60,thin]  (3,0.2)  -- (3,1.9);  \node[anchor=south,font=\scriptsize,teal,inner sep=1pt,align=center] at (3,1.9) {dehumid.\\wet};
  \fill[black!60] (-3,0) circle (1.8pt); \fill[black!60] (-2,0) circle (1.8pt);
  \fill[teal] (2,0) circle (1.8pt); \fill[teal] (3,0) circle (1.8pt);
\end{tikzpicture}\\[2pt]
{\footnotesize (c) Evidence computed by rule}
\end{minipage}
\caption{(a) Share of should-call and of should-not-call When2Call examples that each model calls.
(b) Schematic: at one sensitivity $d'$, the two models' criteria trade hits for false
alarms at the same accuracy. (c) The quad of Figure~\ref{fig:template} on the evidence scale $\Delta U$
that \textsc{FlipAct} computes by rule, with the Yes/No boundary fixed at zero.}
\label{fig:sdt}
\end{figure}

On When2Call~\citep{ross2025when2call}, gemma-2-9b and Phi-3.5-mini reach balanced accuracies of $0.65$
and $0.64$ with opposite behavior: the first rarely calls the tool, even where it should, and the second
calls it on most examples, many where it should not (Figure~\ref{fig:sdt}a). They separate the
two kinds of example about equally well, so on this benchmark the first should act more and the second
less.

Signal-detection theory \citep{greenswets1966} separates sensitivity ($d'$) from criterion
(Figure~\ref{fig:sdt}b; \S\ref{sec:sdt}), but on a benchmark with binary labels the criterion it recovers
is relative to that benchmark's mix of examples, and no zero is shared across benchmarks.
Comparing criteria needs a fixed boundary on a common evidence scale, which authored binary labels do not
provide \citep[cf.][]{jacobs2021measurement}.

\textsc{FlipAct} supplies such a scale by computing the evidence for acting. Its design draws on
qualitative process theory \citep{forbus1984qualitative}, which describes physical change by the sign of
each quantity's change and a coarse ordinal magnitude, enough to say what an action does to
a state without numbers. Each question shows a situation and a candidate tool and asks whether using the
tool is warranted. The state and the tool each have a sign and an ordinal magnitude on a physical
axis. A rule over these properties computes the evidence score $\Delta U$, how much closer the action
brings the state to its goal, with the Yes/No boundary at zero (Figure~\ref{fig:sdt}c). The design unit
is a $2\times2$ \emph{quad}: two tools, each shown in the state it fixes and in the state it worsens, so
the knowledge needed to answer stays the same and the state decides the answer (Figure~\ref{fig:template}).
Fitting a model's Yes-rate against $\Delta U$ gives its sensitivity as the slope $a$ and its
criterion as the score $\tau$ where it crosses one half.

Across $44$ open models (\S\ref{sec:coordinates}), sensitivity behaves as expected of
a capability: it rises with size inside a family, and its ranking agrees with external $d'$ on
three of four when-to-act benchmarks. Models read the sign of the evidence rather than its size,
accepting actions that leave the state no better about as often as helpful ones. When Yes means act,
the criterion is mostly conservative, in some models through a preference for answering No. It is a
stable per-model setting, not set by size: two releases of Mistral-Small-24B sit on opposite sides of
zero.

We move the criterion separately from sensitivity in two ways (\S\ref{sec:interventions}). At inference,
adding a direction read from the model's own activations \citep{zou2023representation,turner2023activation, ke2026neuronsrevealllmsmisuse}
shifts the criterion with the weights fixed. In the weights, distilling the model from its own steered copy
shifts the criterion on four models from three developers with no significant change in sensitivity, and
a random direction of the same norm does not. The shift carries beyond \textsc{FlipAct}'s Yes/No
format: on benchmarks without a Yes/No answer it exceeds the shift from a teacher that only favors Yes,
on both models tested. On every benchmark, accuracy rises or falls as the starting criterion predicts. Supervised training on the computed labels instead raises sensitivity and moves the
criterion with it.

\paragraph{Contributions.}
\begin{enumerate}[leftmargin=1.4em,itemsep=1pt,topsep=2pt]
\item A measurement scheme that splits when-to-act accuracy into a sensitivity and a criterion, read
against an evidence scale computed by rule with the Yes/No boundary fixed for every model.
\item \textsc{FlipAct}, built by an automated pipeline in which language models only annotate what a tool
does, while the states and every answer follow from the rule.
\item Evidence that the criterion moves separately from sensitivity: steering and distillation from a steered
copy move the criterion, while training on the computed labels raises sensitivity.
\end{enumerate}

\section{Related work}
\label{sec:related}

\textbf{When-to-act evaluation.} Benchmarks for deciding whether to act cover tool calling
\citep{ross2025when2call,when2tool2026,huang2024metatool,patil2025bfcl}, clarification
\citep{questbench2025}, abstention \citep{abstentionbench2025} and over-refusal
\citep{rottger2024xstest,cui2024orbench,shi2024overkill}. Most share three design choices: ground truth
is authored by humans or an LLM judge, the evidence for acting is a binary label, and performance is a
single rate that conflates a model's sensitivity with its criterion. \citet{questbench2025} instead
compute ground truth by constraint satisfaction, and \citet{when2tool2026} fix a controlled call/no-call
boundary with deliberate no-tool cases. Neither flips a world-state while holding the required knowledge
fixed, and none of these benchmarks grades the evidence for acting per question. We reuse four of them as
external tests (\S\ref{sec:sensitivity}) and measure transfer on six benchmarks (\S\ref{sec:transfer}).
App.~\ref{app:related-table} compares twelve on these design choices.

\textbf{Signal detection and language models.} Signal-detection theory separates an observer's
sensitivity from its criterion \citep{greenswets1966}. Reading a benchmark score as the measurement of a
latent quantity raises the question of whether it measures what it is read as
\citep{jacobs2021measurement,bean2025measuring}. \citet{cacioli2026signal} fits signal-detection models to
factual question answering with the model's answer log-probability as evidence, so the criterion sits on
each model's own confidence scale. \citet{shi2026tocall} read a call/no-call offset off sparse-autoencoder
features on When2Call and cancel it by steering. Their offset is an intercept on an internal margin,
compared within one model and without a separate sensitivity. \textsc{FlipAct} places the criterion on a
rule-computed evidence scale shared across questions and models, estimates it separately from sensitivity with
the required knowledge held fixed, and moves it into the weights.

\section{The \textsc{FlipAct} diagnostic}
\label{sec:diagnostic}

\subsection{Preliminaries: signal detection}
\label{sec:sdt}

Signal-detection theory \citep{greenswets1966} models an observer who reports a signal whenever a
scalar amount of evidence exceeds a threshold. The sensitivity $d'$ is the distance between the evidence
distributions of signal and noise trials in units of their spread, and the criterion $c$ is where the
threshold sits relative to their midpoint. From the hit rate $H$ and the false-alarm rate $\mathrm{FA}$
the equal-variance estimators, with $z$ the inverse of the standard normal distribution function, are
\begin{equation}
d' \;=\; z(H) - z(\mathrm{FA}), \qquad
c \;=\; -\tfrac12\bigl(z(H) + z(\mathrm{FA})\bigr).
\label{eq:sdt}
\end{equation}
In when-to-act terms a hit is acting when one should and a false alarm is acting when one should not.
A positive $c$ is conservative. \textsc{FlipAct} supplies an evidence score per question, and the same
two quantities become the slope $a$ and the crossing point $\tau$ of a model's Yes-rate against that
score (\S\ref{sec:fit}). Every comparison in the paper pairs $a$ with $d'$ and $\tau$ with $c$.
An interactive version is online.\footnote{\url{https://sixundong.com/projects/flipact}}

\subsection{Same knowledge, flipped state}
\label{sec:design}

\begin{figure}[t]
\centering
\newcommand{\sobj}[1]{\colorbox{blue!10}{\strut #1}}
\newcommand{\sstate}[1]{\colorbox{orange!22}{\strut #1}}
\newcommand{\stool}[1]{\colorbox{teal!14}{\strut #1}}
\setlength{\fboxsep}{1pt}
\begin{tcolorbox}[enhanced, sidebyside, sidebyside align=top, lefthand ratio=0.47, sidebyside gap=10pt, colback=black!2,
  colframe=black!35, boxrule=0.4pt, arc=1.5pt, left=5pt, right=5pt, top=4pt, bottom=4pt,
  fontupper=\small, fontlower=\small, segmentation style={black!25, dashed}]
\raggedright\textbf{Template}\\[3pt]
Situation: The \sobj{\texttt{<object>}} is \sstate{\texttt{<state>}}. {\color{black!55}[\texttt{<supply>}]} The goal is to bring the
\sobj{\texttt{<object>}} back to a normal state. Action: a person uses a \stool{\texttt{<tool>}}. Is
using the \stool{\texttt{<tool>}} a wise choice in this situation? Answer Yes or No.
\tcblower
\textbf{One quad}, object \sobj{room}\\[3pt]
\setlength{\tabcolsep}{4pt}
\begin{tabular}{@{}lcc@{}}
 & \stool{humidifier} & \stool{industrial dehumidifier} \\
 & \footnotesize wets by 2 & \footnotesize dries by 3 \\[1pt]
\sstate{very dry}         & \textcolor{teal}{Yes} \footnotesize$+2$ & \textcolor{black!60}{No} \footnotesize$-3$ \\
\sstate{dangerously wet}  & \textcolor{black!60}{No} \footnotesize$-2$ & \textcolor{teal}{Yes} \footnotesize$+3$ \\
\end{tabular}
\end{tcolorbox}
\caption{The question template (left) and one quad (right) with each question's answer and evidence score
$\Delta U$. The slot \texttt{[<supply>]} states that a prerequisite is met, for tools that have one.}
\label{fig:template}
\end{figure}

Each question shows a physical situation, the goal of returning it to normal and one candidate tool,
and asks whether using the tool is a wise choice. All questions share one minimal template
(Figure~\ref{fig:template}), as do the stimuli of template-based diagnostics
\citep{mccoy2019right,ribeiro2020checklist}, so only the tool, the state and their outcome vary. Physical
situations are used because a tool's effect on a state can be written as an explicit qualitative
transition: a humidifier helps a dry room and harms a wet one. In the quad of Figure~\ref{fig:template},
a humidifier that adds two steps of moisture and an industrial dehumidifier that removes three are
crossed with a very dry and a dangerously wet room. Every question of the quad needs the same two facts,
what the tool does and what the room needs. For a given tool only the state changes, and with it the
answer, so a model that knows what a humidifier is for must still read the state.

Quads and state flips are controls built into the set. The main set has $1{,}880$ questions: $840$ lie
in a complete quad and $1{,}256$ belong to a \emph{state flip}, the same tool with opposite answers in two
states. A state flip separates a model that ignores the state, which gives both questions about the same Yes--No logit
margin, from one that is merely permissive, which may answer Yes to both but still ranks the warranted one
higher (App.~\ref{app:train},~\ref{app:examples}). Every situation has as many Yes as No questions, so the description of the world does not predict the answer.

\subsection{Ground truth and evidence by rule}
\label{sec:gt}

Each situation fixes an integer state $v$ on a physical axis (thermal transfer, moisture, mechanical
cohesion, \dots) with the goal at $0$. Each tool carries a direction $\delta\in\{+1,-1\}$ and an ordinal
magnitude band $\mu\in\{1,2,3\}$ on that axis \citep[in qualitative process theory, a sign and a place in a
quantity space;][]{forbus1984qualitative}, optionally conditional on a prerequisite. Applying it
moves the state to $v+\delta\mu$, and the process behind that change is not modeled. The evidence score
is the utility of the action, how much closer to the goal it leaves the world,
\[
\Delta U \;=\; |v| \;-\; |v+\delta\mu|, \qquad \texttt{Yes} \iff \Delta U > 0 ,
\]
and a tool with unmet prerequisites carries $\Delta U=-2$ for each, a value set by
convention. The question type follows from the same computation: a wrong-direction tool
is \emph{destructive}, a right-direction tool that overshoots or acts on a state already at the goal is
\emph{redundant}, and a tool applied without its prerequisite is \emph{prerequisite}.
Because $v$ and $\mu$ are integers, $\Delta U$ takes seven values on $[-3,3]$. A question conveys $|v|$
only by the adverb \emph{slightly}, \emph{very} or \emph{dangerously}, and $\mu$ only by the tool itself. Language models
annotate each tool's axis, direction, magnitude and prerequisite, and the rule computes the answer from
those properties.

\subsection{An automated build}
\label{sec:pipeline}

\begin{figure}[t]
\centering
\begin{tikzpicture}[
  font=\scriptsize,
  >={Stealth[length=1.6mm]},
  node distance=3mm and 3.6mm,
  stp/.style={rectangle, rounded corners=2pt, line width=0.6pt, text width=25.5mm, minimum height=9.5mm,
               align=center, inner sep=2pt},
  src/.style={stp, draw=black!45, fill=white},
  llm/.style={stp, draw=orange!85!black, fill=orange!15},
  rule/.style={stp, draw=teal!75!black, fill=white},
  final/.style={stp, draw=teal!75!black, line width=1pt, fill=teal!10},
  arr/.style={->, draw=black!55, line width=0.6pt},
  lane/.style={font=\scriptsize\itshape, text=black!55, anchor=east, align=right},
]
\node[src] (pool) {Entity pool\\[1pt]{\color{black!60}$149{,}840$ entities}};
\node[llm, right=of pool] (tag) {Tool-property annotation\\[1pt]{\color{black!60}$2{,}876$ records}};
\node[llm, right=of tag] (screen) {Screens: axis, use,\\magnitude, deployability\\[1pt]{\color{black!60}$1{,}638$ tool records}};
\node[rule, right=of screen] (pair) {Pairing\\[1pt]{\color{black!60}$1{,}422$ tool pairs}};
\node[rule, below=6.5mm of pool] (situ) {Situation synthesis\\[1pt]{\color{black!60}states by rule}};
\node[rule, right=of situ] (gt) {Answer and $\Delta U$ by rule\\[1pt]{\color{black!60}$0$ LLM calls}};
\node[rule, right=of gt] (render) {Template rendering\\[1pt]{\color{black!60}$6{,}176$ questions}};
\node[final, right=of render] (core) {\textbf{Balance selection}\\[1pt]{\color{black!60}$1{,}880$ questions}};
\begin{scope}[on background layer]
  \node[fill=black!3, rounded corners=3pt, inner xsep=2mm, inner ysep=1.6mm, fit=(pool)(pair)] {};
  \node[fill=black!3, rounded corners=3pt, inner xsep=2mm, inner ysep=1.6mm, fit=(situ)(core)] {};
\end{scope}
\draw[arr] (pool) -- (tag);
\draw[arr] (tag) -- (screen);
\draw[arr] (screen) -- (pair);
\draw[arr] (pair.south) -- ++(0,-4mm) -| (situ.north);
\draw[arr] (situ) -- (gt);
\draw[arr] (gt) -- (render);
\draw[arr] (render) -- (core);
\node[lane, left=3mm of pool] {tool\\records};
\node[lane, left=3mm of situ] {questions};
\end{tikzpicture}
\vspace{-2mm}
\caption{The \textsc{FlipAct} construction pipeline. Language models are used only in the orange steps, which
annotate and screen tool properties; answers and evidence scores are computed by rule.}
\label{fig:pipeline}
\vspace{-2mm}
\end{figure}
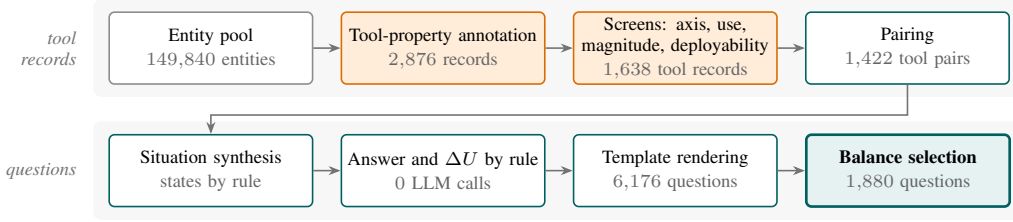

\textsc{FlipAct} is generated by a pipeline (Figure~\ref{fig:pipeline}) that needs no per-question human
annotation or adjudication. Its specification is written once: the physical axes, a closed vocabulary of
tool properties, the rule of \S\ref{sec:gt} and the template. Ten lexical, medical and visually grounded
resources supply $149{,}840$ candidate entities, and two annotator models, Qwen3.5-27B and gemma-4-31b,
fill the vocabulary for each. Where they disagree, both arbitrate against the specification, and
unresolved records are dropped. Tool records (one tool, one axis) are paired when they act in opposite directions on that axis
and target object, and states, answers, question types and text follow by rule. A balance selection keeps
$1{,}880$ of the $6{,}176$ rendered questions. Two matched-pair supplementary sets, $342$
already-satisfied and $224$ prerequisite-blocked questions, are kept outside the fits. A new axis, a new
tool inventory or a larger set runs through the same steps at the cost of annotator calls
(App.~\ref{app:synthesis}).

A lexical probe trained on the question text is at chance on held-out situations whose tools it has not
seen (App.~\ref{app:synth:probe}), and recomputing the ground truth independently of the build gives zero
mismatches. A model's judgment enters only where tools are annotated and screened, and a model family
outside the build reproduces every tool/non-tool judgment, nine in ten axis judgments and every magnitude band to
within one band, two in three exactly. Perturbing the annotations at that rate, or until a quarter of the
computed answers change, keeps the rankings by $a$ and $\tau$ intact and moves criteria toward acting, so it cannot
produce the conservative criteria (App.~\ref{app:noise}).

\section{Measuring sensitivity and criterion}
\label{sec:coordinates}

\textbf{Models.} To see how the coordinates $a$ and $\tau$ vary with size and family, we fit $44$ open models,
all but five with reasoning off (App.~\ref{app:thinking}), from $0.6$B to $70$B parameters
and fifteen families, among them Qwen3 and Qwen3.5, Gemma 2--4, Llama 3, Mistral and Phi
(App.~Tables~\ref{tab:coords}--\ref{tab:checkpoints}).

\textbf{External benchmarks.} To test whether the coordinates capture what when-to-act benchmarks
score, we compare them with four, each posing the decision in a different domain. Every example is one
where acting is right or one where withholding is, and $d'$ and $c$ follow from Eq.~\ref{eq:sdt} with
acting as the response. On BFCL acting is calling a listed function, which is right when one fits the query;
on When2Call it is choosing the tool call among four replies (call, ask, cannot answer, answer directly);
QuestBench and When2Tool ask Yes/No whether the model already has enough information to answer and
whether it would need a tool (App.~Table~\ref{tab:protocol}).

\subsection{The fit and the response profile}
\label{sec:fit}

On \textsc{FlipAct} every question carries a computed evidence score, so we fit the model's Yes-rate
as a logistic curve over that score \citep{greenswets1966}:
\begin{equation}
P(\text{Yes}\mid\Delta U) \;=\; \sigma\!\bigl(a\,(\Delta U - \tau)\bigr).
\label{eq:fit}
\end{equation}
Here $\sigma$ is the logistic function, the slope $a$ the sensitivity and the crossing point $\tau$ the criterion. A positive $\tau$ means
the fitted curve crosses one half above the rule's boundary, so the model answers Yes less often than
the rule warrants, which we call conservative. A criterion is \emph{estimable} when $a\ge0.15$ and $\mathrm{SE}(\tau)\le2.0$, and
\emph{well determined} when it also lies inside the score range; beyond $\pm3$ it is reported as a bound. Both coordinates have split-half
reliability above $0.9$ (App.~\ref{app:fitdetails}). On external benchmarks we use $d'$ and $c$
(Eq.~\ref{eq:sdt}); a model readier to act has lower $\tau$ and $c$.

\begin{wrapfigure}{r}{0.37\linewidth}
\vspace{-5mm}
\centering
\includegraphics[width=\linewidth]{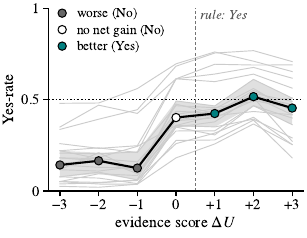}
\vspace{-8mm}
\caption{Yes-rate at each evidence score for the $21$ reasoning-off models with a well-determined criterion (gray), with
their median and interquartile band.}
\label{fig:response}
\vspace{-3mm}
\end{wrapfigure}
\textbf{Response profile.} Models read the sign of the evidence, not its size
(Figure~\ref{fig:response}). Yes-rates jump between $\Delta U=-1$ and $0$, significantly in $19$ of the
$21$ reasoning-off models with a well-determined criterion, and are flat on either side: no model answers Yes
reliably more often at $\Delta U=+3$ than at $+1$. The step sits one unit below the rule's
boundary: models accept actions that leave the state no better about as often as helpful ones. The
questions at $\Delta U=0$ are all redundant slight overshoots, so this set cannot tell whether that
placement belongs to the models or to the rule. Models do withhold when the state is already at its goal
or a prerequisite is unmet (App.~\ref{app:fitdetails}), so the overshoot is the one redundancy they accept. The sign of $\tau$ is set by the height of the Yes
plateau, not by the step: for the median model the Yes-rate on helpful actions stays near one half
(Figure~\ref{fig:response}).

\textbf{Reading $\tau$.} A step function at the sign of the evidence predicts held-out answers as well as the logistic.
Still, $\tau$ tracks the criterion divided by sensitivity: with $c/d'$ of Eq.~\ref{eq:sdt} read from hits
on $\Delta U>0$ and false alarms on $\Delta U<0$, $\tau\approx-0.09+3.22\,c/d'$ ($r=0.999$), and both
find the same models conservative (App.~\ref{app:fitdetails}--\ref{app:graded}). In $9$ of the $21$
models the Yes-rate never reaches one half, so $\tau$ is where the fitted curve, not the data, crosses
it. We therefore read $\tau$ by sign and rank.

\subsection{Sensitivity grows with size and agrees with external benchmarks}
\label{sec:sensitivity}

\begin{figure}[t]
\centering
\includegraphics[width=\linewidth]{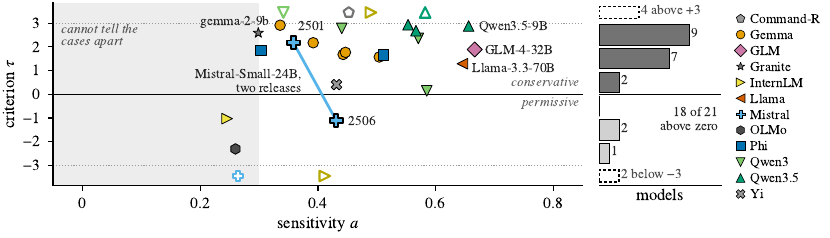}%
\vspace{-8mm}
\caption{Sensitivity against criterion for the $27$ reasoning-off models with an estimable criterion,
with the distribution of the well-determined criteria on the right. Hollow markers at the edges are
criteria beyond $\pm3$. The line joins the two same-size Mistral-Small-24B releases.}
\label{fig:map}
\end{figure}

Within a family, $a$ rises with size: Qwen3 goes from $0.34$ at 4B to $0.58$ at 32B, and the one
exception is Qwen3.5, which peaks at 9B. Twelve models fit below $0.30$, among them widely deployed
small instruct models, and their Yes-rates span almost the whole range: a model can be permissive or
conservative about a tool and still be unable to tell when using it is warranted (Figure~\ref{fig:map}).

\begin{figure}[t]
\centering
\includegraphics[width=0.68\linewidth]{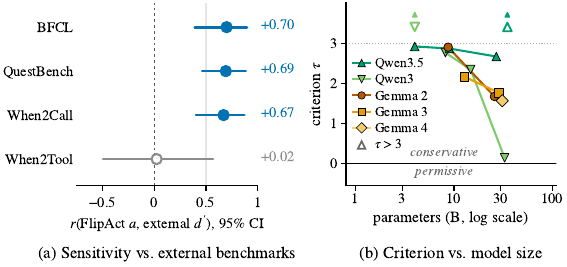}
\vspace{-4mm}
\caption{(a) Correlation across the panel between \textsc{FlipAct} sensitivity $a$ and external $d'$,
with bootstrap intervals; the hollow point is not significant. (b) Criterion against model size for the families with three or more estimable
members, one line per generation. Hollow markers with an arrow are criteria beyond $+3$.}
\label{fig:coords}
\vspace{-4mm}
\end{figure}

\textsc{FlipAct} sensitivity predicts external $d'$ on three of the four benchmarks
(Figure~\ref{fig:coords}a). The comparison uses a \emph{panel}: of the $39$ reasoning-off models, $27$ have an
estimable criterion and $21$ a well-determined one, and the $18$ of these with $a\ge0.30$ and usable
external $d'$ and $c$ form the panel, so that
both sides of each comparison are estimated reliably (App.~\ref{app:check} varies the cutoff on $a$). All
three relations hold with model size partialed out (App.~\ref{app:capability}). This agreement across
three domains and formats supports reading $a$ as the sensitivity these benchmarks score. On When2Tool the relation is not significant.

\subsection{Criterion: a per-model setting, conservative by default}
\label{sec:criterion}

Eighteen of the $21$ reasoning-off models with a well-determined criterion are positive on the canonical wording of
Figure~\ref{fig:template}: they answer Yes less often than the rule warrants (Figure~\ref{fig:map}), and
nearly all tested stay positive under five rephrasings (App.~\ref{app:polarity}). The
criterion is one setting across question types: models that accept more destructive actions also accept
more slight overshoots on other questions ($\rho=0.88$, $27$ models,
$a\ge0.30$), though most accept overshoots more readily.

As in signal detection, the criterion collects every bias in the response, including a preference for
one answer word. We also ask each question in three reversed wordings, such as ``Is using the tool a
mistake?'', under which No means act. After recoding, a preference for No raises the canonical criterion
and lowers the reversed one, so writing them as $\tau_d+b$ and $\tau_d-b$ separates a threshold on the
evidence, $\tau_d$, from the preference, $b$ (App.~\ref{app:polarity}). For the three Qwen3.5 models and
GLM-4-32B, $b$ is $+3$ to $+4$ and $\tau_d$ within a unit of zero; the other twelve of the $16$ models
tested stay conservative without it. JEV \citep{typesafe2026docs}, a model that returns a probability, not text,
and so has no answer word to prefer, is conservative too and steps at the sign of the evidence
(App.~\ref{app:jev}).

Where the criterion sits is a property of the model more than of its size. The two Mistral-Small-24B
releases are the same size and land on opposite sides of zero, at $+2.18$ (2501) and $-1.10$ (3.2, 2506)
(Figure~\ref{fig:map}). Inside a family $\tau$ drifts toward zero as models grow,
in Qwen3, in the dense Qwen3.5 models and across the Gemma generations (Figure~\ref{fig:coords}b). Part
of this drift follows from rising sensitivity, since $\tau$ tracks the criterion divided by sensitivity.
Families differ in their typical $\tau$, so pooled over all models the relation with size vanishes.

The criterion measured on \textsc{FlipAct} does not consistently predict a model's $c$ on other
benchmarks: across the panel the correlation is near zero on When2Call (Pearson $r=-0.01$), QuestBench and
When2Tool, and positive on BFCL ($r=+0.51$), where it rests on the three models with the lowest $\tau$
(App.~Table~\ref{tab:benchmarks}). Section~\ref{sec:transfer} asks instead whether moving $\tau$ moves $c$.

\section{Moving the criterion}
\label{sec:interventions}

We ask whether the criterion can be moved while sensitivity stays fixed, at inference (\S\ref{sec:steer})
and in the weights (\S\ref{sec:train}), and whether the moved criterion carries to external benchmarks
(\S\ref{sec:transfer}). Such a move trades misses for
false alarms and cannot raise accuracy beyond what sensitivity allows.
\subsection{Steering moves the criterion with the weights fixed}
\label{sec:steer}

\begin{table}[t]
\caption{Steering on When2Call, unsteered and at the selected strength, both measured on the same $600$ examples.
Arrows mark a move toward acting, and bold a change in that direction.}
\label{tab:steer}
\begin{center}\small
\setlength{\tabcolsep}{7pt}
\begin{tabular}{lccccr}
\toprule
 & \multicolumn{2}{c}{call rate $\uparrow$} & \multicolumn{2}{c}{criterion $c$ $\downarrow$} & \\
\cmidrule(lr){2-3}\cmidrule(lr){4-5}
Model & unsteered & $+$ steering & unsteered & $+$ steering & $\Delta$ acc. \\
\midrule
Qwen3-8B          & $0.56$ & $\mathbf{0.60}$ & $-0.18$ & $\mathbf{-0.32}$ & $-0.022$ \\
Qwen3.5-27B       & $0.47$ & $\mathbf{0.50}$ & $+0.10$ & $\mathbf{+0.01}$ & $+0.005$ \\
gemma-2-9b        & $0.24$ & $\mathbf{0.29}$ & $+0.78$ & $\mathbf{+0.59}$ & $+0.002$ \\
Mistral-24B       & $0.37$ & $\mathbf{0.55}$ & $+0.39$ & $\mathbf{-0.13}$ & $-0.032$ \\
phi-4             & $0.57$ & $\mathbf{0.59}$ & $-0.30$ & $\mathbf{-0.33}$ & $-0.038$ \\
\bottomrule
\end{tabular}
\end{center}
\vspace{-5mm}
\end{table}

\textbf{Models and data.} We steer five open models, Qwen3-8B, Qwen3.5-27B, gemma-2-9b,
Mistral-Small-24B (the 2501 release, Mistral-24B below) and phi-4, and measure the effect on
When2Call \citep{ross2025when2call}, on $600$ examples, half call and half withhold (eleven more models in App.~\ref{app:s5figs}).

\textbf{Method.} Steering follows the standard recipe
\citep{turner2023activation,zou2023representation,rimsky2024steering,arditi2024refusal}. A logistic probe
from the final-token residual stream at relative depth $0.6$ to the model's own Yes/No answers on the
\textsc{FlipAct} questions gives the criterion direction $\hat w$, which is added during generation to the residual-stream
activation $h$ at that layer as $h \leftarrow h + \lambda\hat w$. For each model the strength $\lambda^\star$ is the one with the lowest
When2Call criterion among those that keep $d'$ within $25\%$ of its unsteered value.

\textbf{Results.} The direction moves the When2Call criterion toward acting on all five models
(Table~\ref{tab:steer}). On Qwen3-8B, gemma-2-9b and Mistral-24B the shift is at least three times that
of same-norm random and orthogonal directions and keeps over two thirds of its size when the refusal,
sycophancy and engagement directions are projected out of $\hat w$; over sixteen models, eleven move
significantly toward acting and four pass both controls (App.~\ref{app:s5figs}). The shift is small on Qwen3.5-27B, which starts near zero on When2Call and whose
\textsc{FlipAct} criterion is mostly a preference for No, and there it does not exceed random directions of the same norm.
On phi-4, which starts permissive and moves least, it does not hold when the strength is chosen on one half
of the examples and measured on the other (App.~\ref{app:s5figs}). On four models steering gains hits at the
cost of correct withholds, and on all five, accuracy changes by at most four points. With the direction fit
on one half of \textsc{FlipAct} and measured on the other, it also moves the criterion of Qwen3.5-9B and
Qwen3.5-27B toward zero and raises their accuracy by $2.6$ and $1.6$ points.

\subsection{Distilling from a steered copy puts the criterion into the weights}
\label{sec:train}

\textbf{Models and data.} As an existence proof that the criterion can be placed in the weights, we
distill four students from three developers, Qwen3-8B, Qwen3.5-9B, gemma-3-12b and Mistral-24B, each from
its own steered copy. The \textsc{FlipAct} main set is split in half so that no situation or tool record
appears on both sides. Students train on one half ($940$ questions) with three seeds per condition, and every result is measured
on the held-out half, so baselines differ from the full-set values of \S\ref{sec:coordinates}.

\begin{table}[t]
\caption{Held-out coordinates and accuracy of each student before training, after criterion distillation
and after sensitivity learning (\S\ref{sec:link}), seed means; shaded rows give the change under each control.
Bold marks a change in the targeted direction (arrows). --: not estimable in every seed. $^\dagger$Beyond $\pm3$.}
\label{tab:train}
\begin{center}\footnotesize
\setlength{\tabcolsep}{4.2pt}
\begin{tabular}{lrrrrrrrrr}
\toprule
 & \multicolumn{3}{c}{Baseline} & \multicolumn{3}{c}{Criterion distillation} & \multicolumn{3}{c}{Sensitivity learning} \\
\cmidrule(lr){2-4}\cmidrule(lr){5-7}\cmidrule(lr){8-10}
Student & $a$ & $\tau$ & Acc. & $a$ & $\tau$ $\downarrow$ & Acc.\ $\uparrow$ & $a$ $\uparrow$ & $\tau$ & Acc.\ $\uparrow$ \\
\midrule
Qwen3-8B     & $0.47$ & $+2.33$ & $0.60$ & $0.55$ & $\mathbf{-2.11}$ & $\mathbf{0.64}$ & $\mathbf{1.09}$ & $-0.02$ & $\mathbf{0.78}$ \\
\rowcolor{black!4}\quad{control} & & & & $+0.02$ & $+0.04$ & $+0.00$ & $-0.19$ & -- & $-0.04$ \\
\arrayrulecolor{black!25}\specialrule{0.3pt}{1pt}{1pt}\arrayrulecolor{black}
Qwen3.5-9B   & $0.75$ & $+1.88$ & $0.65$ & $0.73$ & $\mathbf{+0.86}$ & $\mathbf{0.70}$ & $\mathbf{1.22}$ & $+0.18$ & $\mathbf{0.83}$ \\
\rowcolor{black!4}\quad{control} & & & & $+0.00$ & $+0.20$ & $-0.01$ & $-0.30$ & -- & $-0.04$ \\
\arrayrulecolor{black!25}\specialrule{0.3pt}{1pt}{1pt}\arrayrulecolor{black}
gemma-3-12b  & $0.36$ & $+2.17$ & $0.59$ & $0.35$ & $\mathbf{+0.30}$ & $\mathbf{0.61}$ & $\mathbf{1.56}$ & $-0.41$ & $\mathbf{0.83}$ \\
\rowcolor{black!4}\quad{control} & & & & $+0.00$ & $-0.11$ & $+0.00$ & $-0.20$ & -- & $-0.07$ \\
\arrayrulecolor{black!25}\specialrule{0.3pt}{1pt}{1pt}\arrayrulecolor{black}
Mistral-24B  & $0.59$ & $+3.47$\rlap{$^\dagger$} & $0.59$ & $0.42$ & $\mathbf{+0.19}$ & $\mathbf{0.62}$ & $\mathbf{1.16}$ & $+0.08$ & $\mathbf{0.79}$ \\
\rowcolor{black!4}\quad{control} & & & & $+0.07$ & $-0.10$ & $+0.00$ & $-0.61$ & -- & $-0.06$ \\
\bottomrule
\end{tabular}
\end{center}
\end{table}

\textbf{Method.} In \emph{criterion distillation}, the teacher $T$ is the student with steering on. Its direction is a difference
of means over the student's own Yes and No answers on the training half, and its strength is the one on a
sweep that moves $\tau$ furthest while $a$ and $d'$ stay close to their unsteered values. The student $S$
has the same weights, a LoRA adapter and no steering, and is trained to match the teacher's next-token
distribution at the answer position of each training question $q$,
\begin{equation}
\mathcal{L} \;=\; \sum_{q} \mathrm{KL}\bigl(p_T(\cdot\mid q)\,\big\Vert\,p_S(\cdot\mid q)\bigr).
\label{eq:distil}
\end{equation}
The control replaces the direction with a random one of the same norm. A change is significant when it
exceeds twice its paired bootstrap error over situations (settings in App.~\ref{app:train}).

\textbf{Results.} Criterion distillation moves $\tau$ toward acting on all four students
(Table~\ref{tab:train}), by more than twice its error on three and just under twice on gemma-3-12b.
No student's sensitivity changes significantly in any seed; the largest change, $-0.17\pm0.10$ on
Mistral-24B, is under twice its error (App.~Table~\ref{tab:trainseeds}). An equivalence test bounds the change
within $\pm0.12$ on Qwen3.5-9B and gemma-3-12b and cannot decide on Qwen3-8B and Mistral-24B. On
When2Call, $d'$ stays within $20\%$ of its baseline for all four. Held-out accuracy rises by one to five
points, as a criterion shift should when the model starts conservative, as all four do. The random
direction moves $\tau$ at most a fifth as far. Injected at a strength that perturbs the answers as much as
the teacher's direction, a random direction either leaves $\tau$ in place or collapses the model onto one
answer (App.~\ref{app:train}).

\subsection{The distilled criterion moves the same way on six external benchmarks}
\label{sec:transfer}

\textbf{Benchmarks.} To test whether the shift generalizes beyond \textsc{FlipAct}'s physical
questions and its Yes/No format, each student is tested on six unseen benchmarks, scored with $d'$ and $c$.
Four ask whether to act now: When2Call and When2Tool \citep{when2tool2026} (call a tool), QuestBench
\citep{questbench2025} (ask or solve) and BFCL-miss-func \citep{patil2025bfcl} (the needed function is
removed from the list). Two ask a planning judgment: ACPBench \citep{kokel2025acpbench} (does an action
apply) and CaT-Bench \citep{lal2024catbench} (must one step precede another). When2Call offers four
replies, BFCL-miss-func takes a free-form call or refusal, and the other four take Yes or No.

\begin{figure*}[t]
\centering
\includegraphics[width=\textwidth]{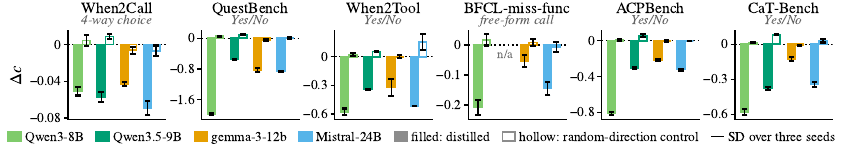}%
\vspace{-5mm}
\caption{Criterion shift on six external benchmarks after criterion distillation (filled) and under the
random-direction control (hollow), seed means $\pm$ SD. Each benchmark has its own scale; negative is toward
acting. Qwen3.5-9B on BFCL-miss-func is left out because its baseline does not reproduce.}
\label{fig:transfer}
\vspace{-2mm}
\end{figure*}

\begin{table}[t]
\begin{minipage}[t]{0.52\linewidth}
\caption{Change in balanced accuracy after criterion distillation for the two Qwen students, on the three
benchmarks where at least one starts conservative; seed means, bold marks a rise.}
\label{tab:acc}
\vspace{2mm}
\centering\footnotesize
\setlength{\tabcolsep}{3pt}
\begin{tabular}{@{}llrr@{}}
\toprule
Student & Benchmark & Start $c$ & $\Delta$ bal.\ acc.\ $\uparrow$ \\
\midrule
Qwen3-8B   & When2Tool  & $+0.42$ & $\mathbf{+0.014}$ \\
           & QuestBench & $+0.71$ & $\mathbf{+0.041}$ \\
           & ACPBench   & $+0.40$ & $\mathbf{+0.025}$ \\
\addlinespace[2pt]
Qwen3.5-9B & When2Tool  & $+0.73$ & $\mathbf{+0.046}$ \\
           & QuestBench & $-0.06$ & $-0.011$ \\
           & ACPBench   & $+0.65$ & $\mathbf{+0.052}$ \\
\bottomrule
\end{tabular}
\end{minipage}\hfill
\begin{minipage}[t]{0.45\linewidth}
\caption{Change in criterion-free AUC after criterion distillation, seed means. Every change is
equivalent to zero within $\pm0.02$ (App.~\ref{app:readout}).}
\vspace{2mm}
\label{tab:auc}
\centering\footnotesize
\setlength{\tabcolsep}{2pt}
\begin{tabular}{@{}lrrrr@{}}
\toprule
 & Qwen3 & Qwen3.5 & gemma-3 & Mistral \\
 & 8B & 9B & 12b & 24B \\
\midrule
When2Call  & $-.008$ & $-.006$ & $+.001$ & $-.002$ \\
QuestBench & $-.012$ & $-.002$ & $-.005$ & $+.002$ \\
When2Tool  & $+.004$ & $-.001$ & $+.011$ & $+.001$ \\
ACPBench   & $+.001$ & $+.001$ & $+.004$ & $-.003$ \\
CaT-Bench  & $+.001$ & $+.008$ & $-.003$ & $+.002$ \\
\bottomrule
\end{tabular}
\end{minipage}
\vspace{-2mm}
\end{table}

\textbf{Results.} Moving the \textsc{FlipAct} criterion moves $c$ the same way on every benchmark,
although across models it does not consistently predict a benchmark's $c$ (\S\ref{sec:criterion}). On all $23$ scored
model--benchmark pairs, $c$ moves toward acting in every seed, and every control moves less than the
distilled student beside it (Figure~\ref{fig:transfer}). Sensitivity is kept: the criterion-free
AUC from answer log-probabilities, available on five benchmarks, stays unchanged for every student
(Table~\ref{tab:auc}).

\textbf{Answer preference.} On a Yes/No format a shift toward acting and a shift toward Yes coincide.
On \textsc{FlipAct} most of the distilled shift flips sign under the reversed wordings of
\S\ref{sec:criterion}, so there it shifts the preference for Yes, the form the decision to act
takes on a Yes/No format (App.~\ref{app:yesbias}). Whether more than that preference carries over shows
only where the answer has no Yes: if a Yes preference were all that transfers, a student distilled from a
teacher whose Yes logits are raised by a constant tuned to the steered teacher's Yes-rate would match the criterion-distilled one there. On Qwen3-8B and
gemma-3-12b it does not: on When2Call and BFCL-miss-func the criterion-distilled student moves $c$ further
in every seed, though on Qwen3-8B the Yes-bias student reaches half to three quarters of it. On the
four Yes/No benchmarks it moves $c$ as far or further.

\textbf{Accuracy.} Accuracy follows the starting point, as a pure criterion shift predicts. Where a
student starts conservative, balanced accuracy rises in every case, by up to six points
(Table~\ref{tab:acc}; all students in App.~Table~\ref{tab:acc_full}). Where it starts permissive, accuracy falls by up to
four points or stays within its error, except Mistral-24B on When2Tool: its hit rate is already
at ceiling, so a shift toward acting only adds false alarms, and accuracy falls about ten points. A
criterion shift should thus take its sign from the target benchmark. Steering the five
models of \S\ref{sec:steer} toward acting on five benchmarks lowers balanced accuracy on $12$ of the
$25$ model--benchmark pairs; choosing the sign and strength from fifty labeled examples of each
benchmark instead leaves only Mistral-24B on When2Call worse (App.~\ref{app:calib}). Where the benchmark has an answer token, a scalar bias on that token chosen the
same way does as well ($19$ of $20$ pairs), as expected if steering is one offset on the criterion.

\subsection{Labels move sensitivity, and the criterion with it}
\label{sec:link}

\emph{Sensitivity learning}, supervised training on the computed labels, tests the converse: whether
labels move sensitivity. It uses the same students, split, adapter and seeds, adds $263$ supplementary
questions on the same tool records to the training half, and shuffles the labels within matched groups of
questions as its control. It raises $a$ on every student, by about $0.5$ to $1.2$, and held-out accuracy
by $17$ to $24$ points, while the shuffled-label control lowers accuracy (Table~\ref{tab:train}). It also
brings $\tau$ close to zero, so labels move both coordinates, whereas distillation moves the
criterion alone. The gain in sensitivity stays on \textsc{FlipAct}: on the external benchmarks its
changes in $d'$ and $c$ differ in sign across students and are matched by the shuffled-label control. Limitations are discussed in
App.~\ref{app:limits}.

\section{Conclusion}
\label{sec:conclusion}

We treated the decision to act as a measurement problem. Two models with the same accuracy can need opposite
fixes, so a when-to-act benchmark says more when it reports $d'$ and $c$ beside accuracy. The criterion
moves, at inference or in the weights, with a shift whose sign the target benchmark sets, while sensitivity
on \textsc{FlipAct} rises with training on labels. On a Yes/No format the criterion also holds a preference between answer
words, which reversed wordings separate from the rest.

\subsection*{Ethics statement}
This work evaluates open-weight models on public benchmarks and involves no human subjects. Moving a
model's criterion toward acting makes it act on thinner evidence, a trade-off practitioners should weigh
for their application.

\subsection*{Reproducibility statement}
The paper specifies the full experimental setup: the construction of \textsc{FlipAct} and its rule
(\S\ref{sec:diagnostic}, App.~\ref{app:synthesis}), the estimators (\S\ref{sec:fit}), and the steering and
training procedures with all hyperparameters (\S\ref{sec:interventions}, App.~\ref{app:train}). Every
training condition is run with three seeds and reported as the seed mean with its bootstrap error, and
every reported change comes with a significance test; steering effects carry bootstrap intervals and are
checked on held-out splits. The dataset, code and evaluation logs will~be~released.

\subsection*{AI use statement}
We used generative AI tools to generate synthetic data sets: in the \textsc{FlipAct} pipeline, language
models annotate tool properties (\S\ref{sec:pipeline}, App.~\ref{app:synthesis}), while every answer and
evidence score is computed by rule. We used generative AI tools to implement methods: the evaluation,
steering, training and analysis code was written with an AI coding assistant. We also used generative AI
tools to aid and polish the writing. We did not use generative AI tools for the other tasks with required
disclosure. The authors reviewed all AI-assisted code, data and text and take responsibility for the
final content of this work.

\bibliography{references}
\bibliographystyle{iclr2027_conference}

\clearpage
\appendix
\raggedbottom%
\section{Construction of \textsc{FlipAct}}
\label{app:synthesis}

This section details the construction summarized in \S\ref{sec:diagnostic}: the annotation schema, the
rule that turns a tool into its questions, a worked quad, the build with its counts, and the checks on
data quality and on annotation noise.

\subsection{Annotation schema}
\label{app:synth:schema}

Each annotator model (App.~\ref{app:synth:build}) fills one record per candidate entity against the
closed vocabulary of Table~\ref{tab:schema}. A value outside the vocabulary is a parse failure and is
never mapped to a neighboring value. A tool that acts on two axes becomes one record
per axis, and each record carries a single direction.

\begin{table}[!ht]
\caption{The annotation schema: the six physical axes, each with its two problem poles as the questions
word them and an example tool in each direction, and the other fields of a tool record.}
\label{tab:schema}
\begin{center}\footnotesize
\setlength{\tabcolsep}{3.0pt}
\resizebox{\linewidth}{!}{%
\begin{tabular}{lllll}
\toprule
 & & & \multicolumn{2}{c}{Example tool} \\
\cmidrule(lr){4-5}
Axis & Quantity & Problem poles & adds ($\delta=+1$) & removes ($\delta=-1$) \\
\midrule
Thermal transfer    & temperature        & cold / hot                  & pocket hand warmer   & ice pack \\
Radiant emission    & visible light      & dark / glaringly bright     & electric lamp        & blackout curtain \\
Mechanical cohesion & connectedness      & loose / jammed              & thread lock compound & puller \\
Spatial containment & seal               & exposed / sealed shut       & lid                  & cap opener \\
Electromagnetism    & circuit continuity & without power / overloaded  & ac adapter           & emergency stop button \\
Moisture level      & moisture           & dry / wet                   & humidifier           & industrial dehumidifier \\
\rowcolor{black!10}\multicolumn{5}{l}{\textbf{Other fields of a record}} \\
is-tool            & \multicolumn{4}{>{\raggedright\arraybackslash}p{10.5cm}@{}}{yes / no; a record that is not a tool leaves the build} \\
Direction $\delta$ & \multicolumn{4}{>{\raggedright\arraybackslash}p{10.5cm}@{}}{$+1$ or $-1$} \\
Magnitude $\mu$    & \multicolumn{4}{>{\raggedright\arraybackslash}p{10.5cm}@{}}{$1$ local, on one object or spot (hot water bottle, paper towel); $2$ room or person scale (hair dryer, space heater); $3$ building, vehicle or industrial scale (furnace, industrial dehumidifier)} \\
Target object      & \multicolumn{4}{>{\raggedright\arraybackslash}p{10.5cm}@{}}{$16$ classes and \emph{other}, e.g.\ a person's body, a room, a container, a rigid object or joint, a circuit or device, soil or a plant} \\
Prerequisite   & \multicolumn{4}{>{\raggedright\arraybackslash}p{10.5cm}@{}}{mains electricity, running water, fuel} \\
Side effect        & \multicolumn{4}{>{\raggedright\arraybackslash}p{10.5cm}@{}}{any of the six axes, or electronic integrity, where the tool has one; not used by the rule} \\
\bottomrule
\end{tabular}%
}
\end{center}
\end{table}

\subsection{From a tool to its questions}
\label{app:synth:rule}

The state of every question is computed from the tool, with the goal at $0$. A solution question places
the state at $v=-\delta\mu$, the deficit the tool closes. A destructive question uses the deficit of the
paired tool, which points the other way, so the tool widens the gap. A redundant question uses a deficit
of one met by a magnitude of at least two, so the tool crosses the goal and leaves the state no closer. A
prerequisite question removes the supply a tool depends on from a solution state. Two tools are paired
only when they act on the same axis and target object in opposite directions and differ in magnitude by
at most one band, so each tool's solution state is its partner's destructive state. A pocket hand warmer
pairs with an ice pack on a person, never with an air conditioner in a room. Prerequisite questions and questions
whose state is already at the goal are not part of the main set. They go to the two supplementary sets
of \S\ref{sec:pipeline}.

Every question is rendered from the template of Figure~\ref{fig:template}, with the state adverb
\emph{slightly}, \emph{very} or \emph{dangerously} for $|v|=1,2,3$. For a tool with a prerequisite,
a sentence between the state and the goal reads ``The supplies this tool depends on (mains electricity)
are available nearby'' or ``None of the supplies this tool depends on (mains electricity) is available
anywhere nearby.'' It appears in every question of such a tool, including those whose supply is present,
so its presence carries no information about the answer.

\subsection{A worked quad}
\label{app:synth:worked}

Table~\ref{tab:workedquad} shows one tool pair on the thermal-transfer axis with a person's body as the
target. Tool $X$ is a blanket, which warms ($\delta=+1$, $\mu=1$), and tool $Y$ is an ice bath, which
cools ($\delta=-1$, $\mu=2$). State $A$ is \emph{slightly cold} ($v=-1$) and state $B$ is
\emph{very hot} ($v=+2$). The two off-diagonal questions are destructive: cooling a body that is already
cold, and wrapping one that is already hot. The fifth question keeps the ice bath and lowers the deficit
to one (\emph{slightly hot}). The ice bath still points the right way, but its magnitude of two now
overshoots the goal, so magnitude alone separates this No from the ice bath's Yes in state $B$.

\begin{table}[!ht]
\caption{One tool pair: the four questions of the quad and the ice bath's redundant question, with the
rule's computation for each.}
\label{tab:workedquad}
\begin{center}\footnotesize
\begin{tabular}{@{}lclcccclc@{}}
\toprule
\multicolumn{2}{@{}c}{State} & \multicolumn{3}{c}{Tool} & \multicolumn{4}{c@{}}{Outcome} \\
\cmidrule(r){1-2}\cmidrule(lr){3-5}\cmidrule(l){6-9}
 & $v$ & & $\delta$ & $\mu$ & $v+\delta\mu$ & $\Delta U$ & Type & Answer \\
\midrule
$A$: slightly cold & $-1$ & blanket  & $+1$ & $1$ & $0$  & $+1$ & solution    & Yes \\
                   & $-1$ & ice bath & $-1$ & $2$ & $-3$ & $-2$ & destructive & No  \\
\arrayrulecolor{black!25}\specialrule{0.3pt}{1pt}{1pt}\arrayrulecolor{black}
$B$: very hot      & $+2$ & ice bath & $-1$ & $2$ & $0$  & $+2$ & solution    & Yes \\
                   & $+2$ & blanket  & $+1$ & $1$ & $+3$ & $-1$ & destructive & No  \\
\arrayrulecolor{black!30}\midrule\arrayrulecolor{black}
slightly hot       & $+1$ & ice bath & $-1$ & $2$ & $-1$ & $0$  & redundant   & No  \\
\bottomrule
\end{tabular}
\end{center}
\end{table}

\subsection{The build}
\label{app:synth:build}

\begin{figure*}[!ht]
\centering
\includegraphics[width=\linewidth]{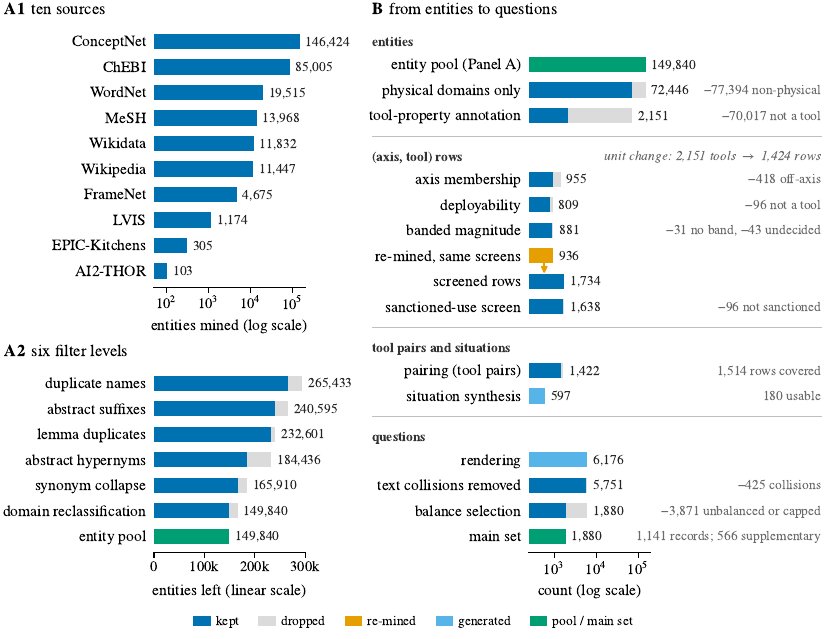}%
\caption{Counts at each stage of the \textsc{FlipAct} construction. A1: the ten entity sources. A2: the
six filter levels that produce the entity pool. B: the pool through tool-property annotation and the
screens to tool records, then tool pairs, situations and questions. A1 and B use a log scale, A2 a linear
one.}
\label{fig:build_funnel}
\end{figure*}

Ten lexical, medical and visually grounded resources supply candidate entities: ConceptNet
\citep{speer2017conceptnet}, Wikipedia \citep{wikipedia} and Wikidata \citep{vrandecic2014wikidata} for
breadth; WordNet \citep{miller1995wordnet}, LVIS \citep{gupta2019lvis}, EPIC-Kitchens
\citep{damen2022rescaling} and AI2-THOR \citep{kolve2017ai2thor} for objects a person handles; MeSH
\citep{lipscomb2000mesh}, ChEBI \citep{hastings2016chebi} and FrameNet \citep{baker1998framenet} for
specialist vocabulary. Figure~\ref{fig:build_funnel} gives the counts at each step. Six
filter levels remove duplicate names, abstract suffixes, lemma duplicates, abstract hypernym paths and
synonyms, and reclassify domains, leaving $149{,}840$ entities. No language model proposes an entity.

Tool-property annotation runs on the $72{,}446$ entities in domains that can hold a physical tool.
Qwen3.5-27B and gemma-4-31b each annotate every property, and each annotator's reading is the majority
over three differently worded prompts at temperature zero. Where the two disagree, both arbitrate
against the written specification, and a row on which the arbiters also disagree is dropped. Most
entities are not tools: the annotators reject $70{,}017$ of $73{,}171$ annotation rows. Screens under the
same protocol check that each remaining tool acts on its axis, is deployable in the situations the
template renders, and has a magnitude band. Tools mined a second time for under-supplied groups pass the
same screens (orange in Figure~\ref{fig:build_funnel}). A last screen removes $96$ rows whose tool is not
the standard instrument of any legitimate trade (a weapon, for example), leaving $1{,}638$ tool records.

Pairing yields $1{,}422$ tool pairs, and the grid of states over them yields $6{,}176$ questions. Removing
texts that collide with each other or with an evaluation set used in the paper leaves $5{,}751$. The
final selection drops every situation whose questions all have the same answer: a situation recurs
across tools, so its answer could otherwise be memorized. It then balances Yes and No within each
remaining situation and caps any one axis, leaving the $1{,}880$ questions of the main set
(Table~\ref{tab:composition}).

\begin{table}[!ht]
\caption{Number of questions in the main set by question type, physical axis and evidence score.}
\label{tab:composition}
\begin{center}\footnotesize
\begin{tabular}{@{}lc@{\hspace{2.5em}}lc@{\hspace{2.5em}}>{\centering\arraybackslash}p{3.6em}>{\centering\arraybackslash}p{3.6em}@{}}
\toprule
\multicolumn{2}{@{}c}{Question type} & \multicolumn{2}{c}{Physical axis} & \multicolumn{2}{c@{}}{Evidence score $|\Delta U|$} \\
\cmidrule(r{2.5em}){1-2}\cmidrule(r{2.5em}){3-4}\cmidrule{5-6}
solution    & $940$ & Mechanical cohesion & $564$ & $0$ & $231$ \\
destructive & $705$ & Thermal transfer    & $476$ & $1$ & $798$ \\
redundant   & $235$ & Electromagnetism    & $234$ & $2$ & $719$ \\
            &       & Spatial containment & $226$ & $3$ & $132$ \\
            &       & Moisture level      & $212$ &     &       \\
            &       & Radiant emission    & $168$ &     &       \\
\bottomrule
\end{tabular}
\end{center}
\end{table}

\subsection{Data quality}
\label{app:synth:probe}

The answer cannot be read from the wording. A lexical probe (TF-IDF features, logistic regression) is
trained on the question text of some situations and read on the others. Every question of a tool that
appears in the held-out situations is removed from training. Because of the state flips, such a tool
would otherwise sit in training in the opposite state with the opposite answer, and the probe would
learn the tool's name with the wrong sign. The probe reaches an AUC of $0.498\pm0.005$ on the main set,
inside the range of $0.485$ to $0.524$ it reaches with permuted labels, and a lexical proxy for the
answer correlates with the label at $|r|=0.022$.

The set is balanced and agrees with the rule. Yes and No are exactly balanced in each of the $180$
situations, with no imbalance per axis or per state level, and $3.9\%$ of the main set's tool records
have Yes and No counts that differ by more than one. Recomputing every answer from the annotated
properties, independently of the build, gives zero mismatches, and no question text is duplicated or
shared with an evaluation set used in the paper.

The tool annotations, the one step where a model's judgment enters, are reproduced by a model family
that took no part in the build. GLM-4-32B re-annotated a stratified sample of $150$ rows with the same
prompts and matched the recorded value on $100\%$ of is-tool judgments and $92.0\%$ of axis judgments. On
magnitude it placed all $150$ tools within one band of the record, $67.3\%$ in the same band. Each of its
other $49$ judgments moves a tool from the lowest or the highest band to the middle one, so it never reverses
the order of two tools by magnitude. Perturbing the annotations at this rate keeps both rankings of the
models above $0.98$ and moves every criterion toward acting (App.~\ref{app:noise}), so annotation error
does not produce the conservative criteria.

\subsection{Robustness to annotation noise}
\label{app:noise}

We test whether annotation error at the rates of App.~\ref{app:synth:probe} could reorder the models or
produce the conservative criteria. The perturbation acts on the magnitude band, the annotated property
that sets the size of the evidence score. We hold every model's answers fixed, perturb the magnitude of
a random share of tool records, recompute the evidence score and answer of every affected question, and
refit $(a,\tau)$, $300$ times per condition. N1 moves $32.7\%$ of records by one band at random, the rate
at which GLM-4-32B disagrees on magnitude. N2 follows its disagreement pattern, one band
toward the middle. N3 scales N2 until $26\%$ of computed answers change, a stress level above both. A synthetic observer with each model's fitted $a$ and $\tau=0$ or $+2.5$ goes through the
same perturbation. Rankings are compared by Spearman correlation with the unperturbed fit, and a model's
criterion shift is the median over replicates minus its unperturbed value.

\begin{table}[!ht]
\caption{Coordinates under three levels of annotation noise, $300$ replicates per condition. The last
block is the criterion recovered for a synthetic observer.}
\label{tab:noise}
\begin{center}\footnotesize
\begin{tabular}{lccc}
\toprule
 & N1 uniform & N2 empirical & N3 stress \\
\midrule
Answers flipped & $10.7\%$ & $15.7\%$ & $26.0\%$ \\
\rowcolor{black!10}\multicolumn{4}{l}{\textbf{Sensitivity and ranking}} \\
\quad $\rho(a)$, median & $0.992$ & $0.991$ & $0.984$ \\
\quad $\rho(\tau)$, median & $0.996$ & $0.995$ & $0.990$ \\
\quad $a$ ratio, perturbed / unperturbed, median & $0.980$ & $0.965$ & $0.976$ \\
\rowcolor{black!10}\multicolumn{4}{l}{\textbf{Criterion}} \\
\quad Shift, mean over models & $-0.19$ & $-0.21$ & $-0.39$ \\
\quad Models shifting by more than $\pm1$ & $0$ & $0$ & $0$ \\
\rowcolor{black!10}\multicolumn{4}{l}{\textbf{Synthetic observer, recovered $\tau$}} \\
\quad $\tau_{\mathrm{true}}=0$, median & $-0.21$ & $-0.27$ & $-0.44$ \\
\quad $\tau_{\mathrm{true}}=0$, 95th percentile & $0.00$ & $-0.06$ & $-0.22$ \\
\quad $\tau_{\mathrm{true}}=+2.5$, median & $+2.38$ & $+2.41$ & $+2.28$ \\
\bottomrule
\end{tabular}
\end{center}
{\footnotesize $\rho(\tau)$ is over the $25$ entries whose criterion lies inside the score range, and
$\rho(a)$ over $30$ entries whose slope fit is stable.}
\end{table}

Noise at the observed rate lowers sensitivity by two to four percent, keeps both rankings above $0.98$,
and moves every criterion toward acting (Table~\ref{tab:noise}). The dominant error relabels a
magnitude-$1$ tool as magnitude $2$, which moves a Yes the model gave at $\Delta U=+1$ to $\Delta U=0$
and pulls the fitted crossing point down. An observer with $\tau=0$ is never recovered at $+1$ or above
in $7{,}500$ replicates, and one at $+2.5$ is recovered within a quarter of a unit. Annotation error
therefore does not produce the conservative criteria, and if anything it understates them. The one model
within $0.15$ of zero, Qwen3-32B, crosses to a small negative value, and every other positive criterion
stays positive in every replicate. Refitting on the $994$ tool records whose magnitude band the two
annotators agreed on without arbitration gives $\rho(a)=0.994$ and $\rho(\tau)=0.995$ and moves no
criterion by more than twice its bootstrap error.

\section{Per-model coordinates and measurement checks}
\label{app:coordtables}

Table~\ref{tab:coords} gives the coordinates of all $44$ models and Table~\ref{tab:checkpoints} the
checkpoint behind each. The evaluation protocol comes first (App.~\ref{app:protocol}), and the other
subsections follow the order of \S\ref{sec:coordinates}: the fit and its readouts
(App.~\ref{app:fitdetails} and~\ref{app:graded}), the convergence of sensitivity with external
benchmarks (App.~\ref{app:convergence} and~\ref{app:capability}), and the criterion across generations
and reasoning modes within a model line (App.~\ref{app:generations} and~\ref{app:thinking}).

\begin{table}[p]
\caption{Coordinates of all $44$ models on \textsc{FlipAct} and their When2Call $d'$ and $c$, grouped
by family with the reasoning-on entries last, and two constant policies for reference. Checkpoints are
in Table~\ref{tab:checkpoints}.}
\label{tab:coords}
\label{tab:whyacc}
\begin{center}\footnotesize
\setlength{\tabcolsep}{3.15pt}\renewcommand{\arraystretch}{0.86}\setlength{\aboverulesep}{0.25ex}\setlength{\belowrulesep}{0.25ex}
\begin{tabular}{lrrrrrrrrrc}
\toprule
 & \multicolumn{2}{c}{Coordinates} & \multicolumn{5}{c}{\textsc{FlipAct} scores} & \multicolumn{2}{c}{When2Call} & \\
\cmidrule(lr){2-3}\cmidrule(lr){4-8}\cmidrule(lr){9-10}
Model & $a$ & $\tau$ & ACC & SQPR & $F_1$ & TNR & MCC & $d'$ & $c$ & Panel \\
\midrule
\rowcolor{black!10}\multicolumn{11}{l}{\textbf{Constant policies}}\\
Always-No  & n/a & n/a & $0.50$ & $0.18$ & n/a & $1.00$ & n/a & n/a & n/a & \\
Always-Yes & n/a & n/a & $0.50$ & $0.35$ & $0.67$ & $0.00$ & n/a & n/a & n/a & \\
\rowcolor{black!10}\multicolumn{11}{l}{\textbf{Qwen3.5}}\\
Qwen3.5-0.8B & $-0.52$\rlap{\smash{$^\dagger$}} & n/a & $0.50$ & $0.19$ & $0.07$ & $0.95$ & $-0.02$ & $0.27$ & $+0.22$ & \\
Qwen3.5-2B & $0.40$\rlap{\smash{$^\dagger$}} & n/a & $0.51$ & $0.21$ & $0.16$ & $0.92$ & $0.02$ & $0.41$ & $+0.39$ & \\
Qwen3.5-4B & $0.55$ & $+2.93$ & $0.61$ & $0.37$ & $0.45$ & $0.90$ & $0.27$ & $1.31$ & $-0.71$ & $\bullet$ \\
Qwen3.5-9B & $0.66$ & $+2.88$ & $0.62$ & $0.39$ & $0.44$ & $0.93$ & $0.30$ & $1.67$ & $-0.52$ & $\bullet$ \\
Qwen3.5-27B & $0.57$ & $+2.68$ & $0.62$ & $0.40$ & $0.48$ & $0.90$ & $0.30$ & $2.00$ & $+0.06$ & $\bullet$ \\
Qwen3.5-35B-A3B & $0.58$ & ${>}{+}3$ & $0.60$ & $0.35$ & $0.39$ & $0.93$ & $0.26$ & $1.43$ & $-0.68$ & \\
\rowcolor{black!10}\multicolumn{11}{l}{\textbf{Qwen3}}\\
Qwen3-0.6B & n/a & n/a & $0.50$ & $0.35$ & $0.67$ & $0.00$ & n/a & n/a & n/a & \\
Qwen3-1.7B & $0.46$\rlap{\smash{$^\dagger$}} & n/a & $0.50$ & $0.35$ & $0.66$ & $0.06$ & $0.02$ & n/a & n/a & \\
Qwen3-4B & $0.34$ & ${>}{+}3$ & $0.54$ & $0.26$ & $0.27$ & $0.91$ & $0.12$ & $1.44$ & $-0.22$ & \\
Qwen3-8B & $0.44$ & $+2.78$ & $0.61$ & $0.39$ & $0.48$ & $0.85$ & $0.24$ & $1.46$ & $-0.26$ & $\bullet$ \\
Qwen3-14B & $0.57$ & $+2.36$ & $0.62$ & $0.40$ & $0.50$ & $0.87$ & $0.28$ & $1.46$ & $-0.40$ & $\bullet$ \\
Qwen3-32B & $0.58$ & $+0.14$ & $0.69$ & $0.54$ & $0.69$ & $0.69$ & $0.38$ & $1.63$ & $+0.01$ & $\bullet$ \\
\rowcolor{black!10}\multicolumn{11}{l}{\textbf{Gemma}}\\
gemma-2-2b & $0.05$\rlap{\smash{$^\dagger$}} & n/a & $0.51$ & $0.27$ & $0.46$ & $0.59$ & $0.01$ & $0.60$ & $+0.27$ & \\
gemma-2-9b & $0.34$ & $+2.91$ & $0.59$ & $0.38$ & $0.49$ & $0.80$ & $0.21$ & $0.99$ & $+0.76$ & $\bullet$ \\
gemma-2-27b & $0.44$ & $+1.68$ & $0.63$ & $0.42$ & $0.56$ & $0.77$ & $0.26$ & $1.21$ & $-0.00$ & $\bullet$ \\
\addlinespace[2pt]
gemma-3-1b & $0.11$\rlap{\smash{$^\dagger$}} & n/a & $0.51$ & $0.31$ & $0.58$ & $0.34$ & $0.03$ & n/a & n/a & \\
gemma-3-4b & $0.12$\rlap{\smash{$^\dagger$}} & n/a & $0.54$ & $0.31$ & $0.47$ & $0.69$ & $0.09$ & n/a & n/a & \\
gemma-3-12b & $0.39$ & $+2.17$ & $0.62$ & $0.41$ & $0.54$ & $0.79$ & $0.25$ & $1.32$ & $-0.37$ & $\bullet$ \\
gemma-3-27b & $0.45$ & $+1.77$ & $0.63$ & $0.43$ & $0.56$ & $0.79$ & $0.28$ & $1.46$ & $-0.38$ & $\bullet$ \\
\addlinespace[2pt]
gemma-4-31b & $0.50$ & $+1.57$ & $0.65$ & $0.45$ & $0.58$ & $0.80$ & $0.31$ & $2.18$ & $-0.41$ & $\bullet$ \\
\rowcolor{black!10}\multicolumn{11}{l}{\textbf{InternLM}}\\
InternLM2.5-1.8B & $1.27$\rlap{\smash{$^\dagger$}} & n/a & $0.50$ & $0.35$ & $0.67$ & $0.00$ & $0.01$ & n/a & n/a & \\
InternLM2.5-7B & $0.25$ & $-1.03$ & $0.59$ & $0.40$ & $0.61$ & $0.52$ & $0.17$ & $0.85$ & $-0.17$ & \\
InternLM2.5-20B & $0.41$ & ${<}{-}3$ & $0.60$ & $0.44$ & $0.69$ & $0.33$ & $0.25$ & $1.00$ & $+0.00$ & \\
InternLM3-8B & $0.49$ & ${>}{+}3$ & $0.57$ & $0.31$ & $0.34$ & $0.93$ & $0.21$ & n/a & n/a & \\
\rowcolor{black!10}\multicolumn{11}{l}{\textbf{Yi}}\\
Yi-1.5-6B & $0.15$\rlap{\smash{$^\dagger$}} & n/a & $0.53$ & $0.30$ & $0.41$ & $0.74$ & $0.08$ & n/a & n/a & \\
Yi-1.5-9B & $0.14$\rlap{\smash{$^\dagger$}} & n/a & $0.55$ & $0.38$ & $0.62$ & $0.37$ & $0.12$ & $0.63$ & $-0.05$ & \\
Yi-1.5-34B & $0.43$ & $+0.41$ & $0.64$ & $0.47$ & $0.63$ & $0.67$ & $0.28$ & $1.01$ & $+0.13$ & $\bullet$ \\
\rowcolor{black!10}\multicolumn{11}{l}{\textbf{Mistral}}\\
Mistral-7B-v0.3 & $0.26$ & ${<}{-}3$ & $0.54$ & $0.40$ & $0.66$ & $0.21$ & $0.11$ & $0.31$ & $-0.16$ & \\
Mistral-Small-24B (2501) & $0.36$ & $+2.18$ & $0.61$ & $0.41$ & $0.54$ & $0.78$ & $0.24$ & $1.18$ & $-0.04$ & $\bullet$ \\
Mistral-Small-3.2-24B (2506) & $0.43$ & $-1.10$ & $0.64$ & $0.50$ & $0.68$ & $0.52$ & $0.29$ & $1.10$ & $-0.33$ & $\bullet$ \\
\rowcolor{black!10}\multicolumn{11}{l}{\textbf{Llama}}\\
Llama-3.1-8B & $0.10$\rlap{\smash{$^\dagger$}} & n/a & $0.55$ & $0.34$ & $0.57$ & $0.49$ & $0.09$ & $0.29$ & $-1.07$ & \\
Llama-3.3-70B & $0.65$ & $+1.29$ & $0.67$ & $0.49$ & $0.62$ & $0.82$ & $0.36$ & $1.54$ & $-0.39$ & $\bullet$ \\
\rowcolor{black!10}\multicolumn{11}{l}{\textbf{Phi}}\\
Phi-3.5-mini & $0.30$ & $+1.85$ & $0.59$ & $0.38$ & $0.54$ & $0.72$ & $0.20$ & $0.88$ & $-0.64$ & $\bullet$ \\
phi-4 & $0.51$ & $+1.67$ & $0.64$ & $0.44$ & $0.57$ & $0.80$ & $0.29$ & $1.80$ & $-0.28$ & $\bullet$ \\
\rowcolor{black!10}\multicolumn{11}{l}{\textbf{Other families}}\\
GLM-4-32B-0414 & $0.67$ & $+1.89$ & $0.66$ & $0.45$ & $0.56$ & $0.87$ & $0.35$ & $1.70$ & $+0.14$ & $\bullet$ \\
Command-R-08-2024 & $0.45$ & ${>}{+}3$ & $0.57$ & $0.31$ & $0.34$ & $0.92$ & $0.20$ & $0.92$ & $-0.56$ & \\
Granite-3.3-8B & $0.30$ & $+2.59$ & $0.59$ & $0.37$ & $0.51$ & $0.76$ & $0.19$ & $0.77$ & $-0.16$ & \\
OLMo-2-1124-13B & $0.26$ & $-2.31$ & $0.59$ & $0.41$ & $0.64$ & $0.44$ & $0.18$ & $0.64$ & $+1.18$ & \\
Nemotron-Nano-9B-v2 & $1.07$\rlap{\smash{$^\dagger$}} & n/a & $0.50$ & $0.18$ & $0.01$ & $1.00$ & $0.02$ & n/a & n/a & \\
\rowcolor{black!10}\multicolumn{11}{l}{\textbf{Reasoning on}}\\
Qwen3-30B-A3B-Thinking-2507 & $0.55$ & ${>}{+}3$ & $0.62$ & $0.39$ & $0.45$ & $0.92$ & $0.30$ & n/a & n/a & \\
DeepSeek-R1-Distill-Llama-8B & $0.30$ & $+1.47$ & $0.61$ & $0.44$ & $0.57$ & $0.70$ & $0.22$ & n/a & n/a & \\
DeepSeek-R1-Distill-Qwen-32B & $0.48$ & $+1.48$ & $0.65$ & $0.47$ & $0.59$ & $0.79$ & $0.31$ & n/a & n/a & \\
DeepSeek-R1-Distill-Llama-70B & $0.48$ & $+0.59$ & $0.66$ & $0.50$ & $0.64$ & $0.71$ & $0.32$ & n/a & n/a & \\
gpt-oss-20b & $0.53$ & $+1.73$ & $0.65$ & $0.46$ & $0.58$ & $0.82$ & $0.32$ & n/a & n/a & \\
\bottomrule
\end{tabular}
\end{center}
{\footnotesize $a$ is sensitivity and $\tau$ the criterion, positive when conservative. A $\tau$ beyond
the score range is shown as a bound, and $^\dagger$ marks a fit whose criterion is not estimable.
ACC, $F_1$ (Yes class), true-negative rate (TNR) and Matthews correlation (MCC) are over the $1{,}880$ main-set questions, and SQPR is the share of
its $1{,}141$ tool records with every question answered correctly. $\bullet$ marks the panel of
\S\ref{sec:sensitivity}, and n/a a value that is undefined or not available.\par}
\end{table}

\begin{table}[p]
\caption{The Hugging Face checkpoint behind each entry and the precision of its weights, grouped by
family with the family's reference. The reasoning-mode comparison of App.~\ref{app:thinking} also
runs the listed checkpoints with reasoning on.}
\label{tab:checkpoints}
\begin{center}\scriptsize
\setlength{\tabcolsep}{3.0pt}\renewcommand{\arraystretch}{0.9}\setlength{\aboverulesep}{0.25ex}\setlength{\belowrulesep}{0.25ex}
\begin{tabular}{llll}
\toprule
Entry & Checkpoint & Weights & Note \\
\midrule
\rowcolor{black!10}\multicolumn{4}{l}{\textbf{Qwen3.5} \citep{qwen35}}\\
Qwen3.5-0.8B & \texttt{Qwen/Qwen3.5-0.8B} & bf16 & \\
Qwen3.5-2B & \texttt{Qwen/Qwen3.5-2B} & bf16 & \\
Qwen3.5-4B & \texttt{Qwen/Qwen3.5-4B} & bf16 & \\
Qwen3.5-9B & \texttt{Qwen/Qwen3.5-9B} & bf16 & \\
Qwen3.5-27B & \texttt{Qwen/Qwen3.5-27B} & bf16 & \\
Qwen3.5-35B-A3B & \texttt{Qwen/Qwen3.5-35B-A3B} & bf16 & \\
\rowcolor{black!10}\multicolumn{4}{l}{\textbf{Qwen3} \citep{yang2025qwen3}}\\
Qwen3-0.6B & \texttt{Qwen/Qwen3-0.6B} & bf16 & \\
Qwen3-1.7B & \texttt{Qwen/Qwen3-1.7B} & bf16 & \\
Qwen3-4B & \texttt{Qwen/Qwen3-4B} & bf16 & \\
Qwen3-8B & \texttt{Qwen/Qwen3-8B} & bf16 & \\
Qwen3-14B & \texttt{Qwen/Qwen3-14B} & bf16 & \\
Qwen3-32B & \texttt{Qwen/Qwen3-32B} & bf16 & \\
Qwen3-30B-A3B-Thinking-2507 & \texttt{Qwen/Qwen3-30B-A3B-Thinking-2507} & bf16 & reasoning on \\
\rowcolor{black!10}\multicolumn{4}{l}{\textbf{Gemma} \citep{gemma2024gemma2,gemma2025gemma3,gemma4}}\\
gemma-2-2b & \texttt{google/gemma-2-2b-it} & bf16 & \\
gemma-2-9b & \texttt{google/gemma-2-9b-it} & bf16 & \\
gemma-2-27b & \texttt{google/gemma-2-27b-it} & bf16 & \\
gemma-3-1b & \texttt{google/gemma-3-1b-it} & bf16 & \\
gemma-3-4b & \texttt{google/gemma-3-4b-it} & bf16 & \\
gemma-3-12b & \texttt{google/gemma-3-12b-it} & bf16 & \\
gemma-3-27b & \texttt{google/gemma-3-27b-it} & bf16 & \\
gemma-4-31b & \texttt{google/gemma-4-31b-it} & bf16 & \\
\rowcolor{black!10}\multicolumn{4}{l}{\textbf{InternLM} \citep{cai2024internlm2}}\\
InternLM2.5-1.8B & \path{internlm/internlm2_5-1_8b-chat} & bf16 & \\
InternLM2.5-7B & \path{internlm/internlm2_5-7b-chat} & bf16 & \\
InternLM2.5-20B & \path{internlm/internlm2_5-20b-chat} & bf16 & \\
InternLM3-8B & \texttt{internlm/internlm3-8b-instruct} & bf16 & \\
\rowcolor{black!10}\multicolumn{4}{l}{\textbf{Yi} \citep{young2024yi}}\\
Yi-1.5-6B & \texttt{01-ai/Yi-1.5-6B-Chat} & bf16 & \\
Yi-1.5-9B & \texttt{01-ai/Yi-1.5-9B-Chat} & bf16 & \\
Yi-1.5-34B & \texttt{01-ai/Yi-1.5-34B-Chat} & bf16 & \\
\rowcolor{black!10}\multicolumn{4}{l}{\textbf{Mistral} \citep{jiang2023mistral,mistralsmall2025}}\\
Mistral-7B-v0.3 & \texttt{mistralai/Mistral-7B-Instruct-v0.3} & bf16 & \\
Mistral-Small-24B (2501) & \texttt{mistralai/Mistral-Small-24B-Instruct-2501} & bf16 & \\
Mistral-Small-3.2-24B (2506) & \texttt{mistralai/Mistral-Small-3.2-24B-Instruct-2506} & bf16 & \\
\rowcolor{black!10}\multicolumn{4}{l}{\textbf{Llama} \citep{grattafiori2024llama3}}\\
Llama-3.1-8B & \texttt{meta-llama/Llama-3.1-8B-Instruct} & bf16 & \\
Llama-3.3-70B & \texttt{kosbu/Llama-3.3-70B-Instruct-AWQ} & 4-bit AWQ & \\
\rowcolor{black!10}\multicolumn{4}{l}{\textbf{Phi} \citep{abdin2024phi3,abdin2024phi4}}\\
Phi-3.5-mini & \texttt{microsoft/Phi-3.5-mini-instruct} & bf16 & \\
phi-4 & \texttt{microsoft/phi-4} & bf16 & \\
\rowcolor{black!10}\multicolumn{4}{l}{\textbf{GLM} \citep{glm2024chatglm}}\\
GLM-4-32B-0414 & \texttt{zai-org/GLM-4-32B-0414} & bf16 & \\
\rowcolor{black!10}\multicolumn{4}{l}{\textbf{Command R} \citep{cohere2024commandr}}\\
Command-R-08-2024 & \texttt{AMead10/c4ai-command-r-08-2024-awq} & 4-bit AWQ & \\
\rowcolor{black!10}\multicolumn{4}{l}{\textbf{Granite} \citep{granite2024granite3}}\\
Granite-3.3-8B & \texttt{ibm-granite/granite-3.3-8b-instruct} & bf16 & \\
\rowcolor{black!10}\multicolumn{4}{l}{\textbf{OLMo} \citep{olmo2024olmo2}}\\
OLMo-2-1124-13B & \texttt{allenai/OLMo-2-1124-13B-Instruct} & bf16 & \\
\rowcolor{black!10}\multicolumn{4}{l}{\textbf{Nemotron} \citep{nvidia2025nemotronnano2,nemotron3nano4b}}\\
Nemotron-Nano-9B-v2 & \texttt{nvidia/NVIDIA-Nemotron-Nano-9B-v2} & bf16 & \\
Nemotron-3-Nano-4B & \texttt{nvidia/NVIDIA-Nemotron-3-Nano-4B-BF16} & bf16 & App.~\ref{app:thinking} only \\
\rowcolor{black!10}\multicolumn{4}{l}{\textbf{DeepSeek-R1 distillations} \citep{deepseek2025r1}}\\
DeepSeek-R1-Distill-Llama-8B & \texttt{deepseek-ai/DeepSeek-R1-Distill-Llama-8B} & bf16 & reasoning on \\
DeepSeek-R1-Distill-Qwen-32B & \texttt{deepseek-ai/DeepSeek-R1-Distill-Qwen-32B} & bf16 & reasoning on \\
DeepSeek-R1-Distill-Llama-70B & \texttt{casperhansen/deepseek-r1-distill-llama-70b-awq} & 4-bit AWQ & reasoning on \\
\rowcolor{black!10}\multicolumn{4}{l}{\textbf{gpt-oss} \citep{openai2025gptoss120bgptoss20bmodel}}\\
gpt-oss-20b & \texttt{openai/gpt-oss-20b} & MXFP4 (native) & reasoning on \\
\bottomrule
\end{tabular}
\end{center}
\end{table}

\subsection{Evaluation protocol}
\label{app:protocol}

\paragraph{\textsc{FlipAct}.} Each question is one user turn, rendered with the model's own chat template
and its generation prompt, with no system message added. After the template's closing ``Answer Yes or
No.'' the prompt adds a new line, ``Answer with exactly one word: Yes or No.'' With reasoning off, every
question of the main and supplementary sets is answered once with vLLM, sampling at temperature $0.7$,
top-$p$ $0.8$ and top-$k$ $20$ with seed $0$ and at most $8$ new tokens, the same settings for every
model except gemma-4-31b, which is run with transformers and greedy decoding. The answer is the first
standalone \emph{yes} or \emph{no} in the output, ignoring case. An
output with neither is unparsed and left out of the fit. No entry has more than one unparsed main-set
question. Two further conditions guard the panel against unreadable output: at most $30\%$ unparsed
\textsc{FlipAct} answers and a When2Call $d'$ of at least $0.5$. In practice neither removes a model that
the conditions of \S\ref{sec:sensitivity} admit. The fit of Eq.~\ref{eq:fit} is a maximum-likelihood logistic regression on these binary answers,
so every Yes-rate and $P(\text{Yes})$ on \textsc{FlipAct} is a proportion of sampled answers, not a token
probability. $\mathrm{SE}(\tau)$ is half the width of the profile-likelihood interval within which the
deviance rises by at most one. Nine entries were also run with seed $1$. The two runs give the same
answer on $93$ to $100\%$ of main-set questions, and for the six of the nine with a well-determined
criterion both coordinates of the second run fall inside the $95\%$ bootstrap intervals of the first.
The steering and training readouts of \S\ref{sec:interventions} use the same prompt with reasoning off
but do not sample: an answer is Yes when the logit of the first token of ``Yes'' exceeds that of ``No''
at the answer position.

\paragraph{Reasoning on.} The five reasoning-on entries of Table~\ref{tab:coords} and the reasoning-on
condition of App.~\ref{app:thinking} receive the same prompt, one-word instruction included. Hybrid models
reason through their template's switch (\texttt{enable\_thinking}, or \texttt{thinking} for
Granite-3.3-8B), and gpt-oss-20b runs at its template's default reasoning effort, medium. Decoding
samples at temperature $0.6$, top-$p$ $0.95$ and top-$k$ $20$ with seed $0$, within a budget of
$8{,}192$ new tokens ($4{,}096$ for Granite-3.3-8B and Qwen3.5-35B-A3B). The answer is the first
standalone \emph{yes} or \emph{no} after the end of the reasoning trace, which is the last
\texttt{</think>} or, for gpt-oss-20b, the start of its final channel. An answer is unparsed when its trace reaches the
budget without closing or is followed by no Yes or No, and no answer is read from inside a trace. A model is not fit in a mode where more than $10\%$ of its answers are unparsed,
which removes Qwen3.5-0.8B and Qwen3.5-2B with reasoning on ($78\%$ and $23\%$). Every other condition stays
below $6\%$, and the five reasoning-on entries of Table~\ref{tab:coords} have no unparsed answer.
External benchmarks are run with reasoning off only, so reasoning-on entries have no external readout
and are not in the panel.

\paragraph{External benchmarks.} Table~\ref{tab:protocol} describes the seven benchmarks. Each example
is one user turn through the chat template with reasoning off, decoded greedily with vLLM (with
transformers for gemma-4-31b), with at most $8$ new tokens on QuestBench, $64$ on When2Call, $128$ on
BFCL and $256$ on the other four. Yes/No replies are parsed as on \textsc{FlipAct}. A When2Call reply is
read as the letter it starts with, else a letter after a cue such as ``answer'' or ``option'' in its
first $80$ characters, else any standalone letter A--D there. On both BFCL sets a reply acts when it
contains \texttt{CALL: f(\dots)} or calls a target function by name without that prefix. A reply with
no parsable answer counts as withholding and stays in the denominators. No reply in the transfer test
is unparsed. In the panel, at most $0.3\%$ of replies are unparsed on the Yes/No benchmarks, and on
When2Call the highest rate is $12.5\%$, for Mistral-Small-3.2-24B. Table~\ref{tab:coords} shows an
entry's When2Call values as n/a when more than $30\%$ of its replies are unparsed, which applies only to
Nemotron-Nano-9B-v2.
Runs with a steering hook (App.~\ref{app:s5figs} and~\ref{app:crossds}) use transformers with greedy
decoding and at most $8$ new tokens ($64$ on BFCL-miss-func), on a balanced sample of $600$ examples of
When2Call, When2Tool and CaT-Bench and on all examples of ACPBench and BFCL-miss-func.

The BFCL set of the panel draws $80$ single-turn queries at random from BFCL's live-simple category, each
answered by its one listed function, and $36$ from its live-irrelevance category, where none of the one to
ten listed functions fits the query. Each function is restated as its name, description and parameter names,
under an instruction to reply with one \texttt{CALL:} line if a tool is needed and to answer directly
otherwise.

\begin{table}[h]
\caption{The external benchmarks, grouped by the comparison that uses them: source split, number of
examples in each class, the act and withhold examples, and the reply format.}
\label{tab:protocol}
\begin{center}\footnotesize
\setlength{\tabcolsep}{3pt}
\begin{tabular}{>{\raggedright\arraybackslash}p{2.75cm}r>{\raggedright\arraybackslash}p{5.9cm}>{\raggedright\arraybackslash}p{2.6cm}}
\toprule
Benchmark & Act / withhold & Act example (withhold example) & Reply format \\
\midrule
\rowcolor{black!10}\multicolumn{4}{l}{\textbf{Panel comparison (\S\ref{sec:sensitivity})}} \\
BFCL\newline{\scriptsize{v3 live simple, live irrelevance}}
  & $80$ / $36$
  & a query its one listed function answers (a query no listed function fits)
  & free-form; \texttt{CALL: f(\dots)} acts \\
\rowcolor{black!10}\multicolumn{4}{l}{\textbf{Panel comparison and transfer test (\S\ref{sec:sensitivity}, \S\ref{sec:transfer})}} \\
QuestBench\newline{\scriptsize{Logic-Q ($500$ of $1{,}150$ tasks), GSM-Q CSP ($151$), GSM-Q verbal ($288$)}}
  & $939$ / $939$
  & a fully specified task (the same task with one variable removed); acting is Yes to ``Do you already
    have enough information to give a single correct final answer \dots?''
  & Yes/No \\
\addlinespace[2pt]
When2Call\newline{\scriptsize{test, multiple choice}}
  & $1{,}295$ / $2{,}357$
  & a request whose correct reply is a tool call (one whose correct reply asks for missing information,
    $1{,}062$, or says no tool can answer, $1{,}295$); acting is choosing the tool-call reply
  & one letter for four replies: call, ask, cannot answer, answer directly \\
\addlinespace[2pt]
When2Tool\newline{\scriptsize{single-hop test, $15$ environments}}
  & $750$ / $750$
  & a hard-tier task, built to need a tool (an easy-tier task, solvable without one; the $750$
    medium-tier tasks are not scored); acting is Yes to ``would you need to use one of the tools?''
  & Yes/No \\
\rowcolor{black!10}\multicolumn{4}{l}{\textbf{Transfer test (\S\ref{sec:transfer})}} \\
BFCL-miss-func\newline{\scriptsize{v3 multi-turn miss-func, $200$ scenarios}}
  & $200$ / $200$
  & a request with the one to three functions it needs listed (the same request with those functions removed)
  & free-form; \texttt{CALL: f(\dots)} acts \\
\addlinespace[2pt]
ACPBench\newline{\scriptsize{applicability, test and validation}}
  & $88$ / $82$
  & an action applicable in the stated planning state (an inapplicable action); acting is Yes
  & Yes/No \\
\addlinespace[2pt]
CaT-Bench\newline{\scriptsize{test, $1{,}420$ step pairs in both directions}}
  & $1{,}420$ / $1{,}420$
  & two recipe steps of which one must precede the other (two independent steps); acting is Yes
  & Yes/No \\
\bottomrule
\end{tabular}
\end{center}
\end{table}

On every benchmark, and in the signal-detection readout of App.~\ref{app:graded}, $d'$ and $c$ follow
Eq.~\ref{eq:sdt} with the log-linear correction $H=(\text{hits}+\tfrac12)/(n_{\text{act}}+1)$ and
$\mathrm{FA}=(\text{false alarms}+\tfrac12)/(n_{\text{withhold}}+1)$, applied to every rate so that $z$
stays finite for a model that always or never acts.

For the criterion-free AUC (App.~\ref{app:check} and~\ref{app:readout}), the benchmarks it covers are
run again with the same greedy decoding, recording the $20$ most likely tokens and their
log-probabilities at the first four generated positions. At the first position whose generated token is
an answer, the margin is the log-probability of acting (Yes, or the tool-call letter) minus that of the
strongest alternative (No, or the most likely other letter), each answer taking its most likely spelling
among the recorded tokens. A model--benchmark pair is scored only if the sign of its margin reproduces at
least $99\%$ of its decoded answers. Examples with no answer token among the recorded positions, and
When2Call replies with no parsable letter, are left out; Mistral-Small-3.2-24B loses the most ($959$ of
$1{,}878$ QuestBench and $463$ of $3{,}652$ When2Call examples). gemma-4-31b, run with transformers, has
no recorded log-probabilities and is absent from these readouts.

\subsection{Reliability and response profile}
\label{app:fitdetails}
\label{sec:robustness}

Both coordinates are reliable. Refitting on $200$ random halves of the tool records with the
Spearman--Brown correction gives a split-half reliability of $0.959$ for $a$ and $0.991$ for $\tau$.
Tool records share question text across situations, so the stricter unit is the situation cluster, all
questions that share one situation text. It is the resampling unit for every interval in the paper.
Split by cluster, the two reliabilities are $0.906$ and $0.973$.

Models respond to the sign of the evidence score and ignore its size (Table~\ref{tab:profile} and
Figure~\ref{fig:response}). The jump in Yes-rate between $\Delta U=-1$ and $0$ is significant in $19$
of the $21$ models with a well-determined criterion, and the jump between $0$ and $+1$ in none. Within
either side, no model answers Yes reliably more often at the larger magnitude. The step therefore
sits one unit below the rule's boundary. Every question at $\Delta U=0$ is a slight overshoot, and
models accept these overshoots about as often as helpful actions: the overshoot Yes-rate lies nearer the
hit rate than the false-alarm rate in most models, and its interval includes the hit rate in $20$ of
$21$.

\begin{table}[!ht]
\caption{Median Yes-rate at each evidence score over the $21$ models with a well-determined criterion,
the median line of Figure~\ref{fig:response}.}
\label{tab:profile}
\begin{center}\footnotesize
\setlength{\tabcolsep}{6pt}
\begin{tabular}{@{}lccccccc@{}}
\toprule
$\Delta U$ & $-3$ & $-2$ & $-1$ & $0$ & $+1$ & $+2$ & $+3$ \\
\midrule
Median Yes-rate & $0.14$ & $0.17$ & $0.13$ & $0.40$ & $0.43$ & $0.52$ & $0.46$ \\
\bottomrule
\end{tabular}
\end{center}
\end{table}

A step fits held-out answers as well as the logistic. In five-fold cross-validation with folds by
situation, the logistic in $\Delta U$ improves on a step with one Yes-rate per side by a median
$0.010$ nats per question $[-0.003,+0.024]$, reliably in $3$ of $21$ models. A step between $-1$
and $0$ fits best in $20$ of $21$, although its location was read off these same curves. In $9$ of
the $21$ models the observed Yes-rate, smoothed to be monotone, never reaches one half, so their $\tau$ is
where the fitted curve crosses it.

Refitting with $\Delta U$ collapsed to $\{-1,0,+1\}$ reproduces the sensitivity ranking
($\rho=0.985$). Three monotone re-expressions of the score (centered ranks, destructive scores halved, a
per-axis $z$-score) keep $\rho(a)\ge0.98$ over all fitted models and $\rho(\tau)\ge0.995$, and leave
$r(a,d')$ on When2Call between $+0.66$ and $+0.67$.

The two supplementary sets show that the overshoot is the one redundancy models accept
(Table~\ref{tab:annex}). Where the state is already at its goal, every model answers Yes less often than
at the main set's slight overshoots, by a median of $0.27$. Where a tool's supply is missing, every model
answers Yes less often than on the same question with the supply present, by a median of $0.43$. Models
thus withhold when an action is plainly unneeded or cannot work, and accept it when it points the right
way but goes one step too far.

\begin{table}[!ht]
\caption{Yes-rate on the main set by evidence sign and on the two supplementary sets, median and range
over the $21$ models with a well-determined criterion. Each supplementary question is paired with a
matched question whose answer is Yes.}
\label{tab:annex}
\begin{center}\footnotesize
\setlength{\tabcolsep}{6pt}
\begin{tabular}{llcc}
\toprule
Questions & Answer & Median Yes-rate & Range \\
\midrule
\rowcolor{black!10}\multicolumn{4}{l}{\textbf{Main set}} \\
destructive, $\Delta U<0$            & No  & $0.15$ & $[0.03, 0.51]$ \\
slight overshoot, $\Delta U=0$       & No  & $0.40$ & $[0.18, 0.70]$ \\
helpful, $\Delta U>0$                & Yes & $0.47$ & $[0.30, 0.76]$ \\
\rowcolor{black!10}\multicolumn{4}{l}{\textbf{Already satisfied}} \\
state at its goal                    & No  & $0.08$ & $[0.01, 0.68]$ \\
matched question                     & Yes & $0.23$ & $[0.09, 0.58]$ \\
\rowcolor{black!10}\multicolumn{4}{l}{\textbf{Prerequisite}} \\
supply missing                       & No  & $0.29$ & $[0.00, 0.51]$ \\
matched question, supply present     & Yes & $0.66$ & $[0.46, 0.90]$ \\
\bottomrule
\end{tabular}
\end{center}
\end{table}

\subsection{A signal-detection readout}
\label{app:graded}

Because models read the sign of the score, Table~\ref{tab:graded} sets the logistic coordinates beside a
signal-detection readout: hits on $\Delta U>0$ and false alarms on $\Delta U<0$ enter
Eq.~\ref{eq:sdt}, and the overshoot questions at $\Delta U=0$ are left out. Sensitivity ranks the
models the same under every readout. So does the criterion when it is divided by sensitivity, as
$c/d'$: $\tau$ is close to a linear function of it ($r=0.999$), $\tau\approx-0.09+3.22\,c/d'$, where
$-0.09$ is the midpoint of the mean negative and the mean positive score of the main set and $3.22$ the
distance between them. The absolute $c$ ranks models less closely with $\tau$ ($\rho=0.77$) because
$\tau$, like $c/d'$, is divided by sensitivity and $c$ is not. The number of positive criteria, the
opposite signs of the two Mistral-Small-24B releases and the convergence of sensitivity with external
$d'$ are the same under every readout.

\begin{table}[!ht]
\caption{The logistic coordinates beside a signal-detection readout that leaves out $\Delta U=0$. Rank
agreement with the logistic fit and the count of criteria with an interval above zero are over the $21$
models with a well-determined criterion, and $r$ is over the panel. Shaded: the fit of the main text.}
\label{tab:graded}
\begin{center}\footnotesize
\setlength{\tabcolsep}{4pt}
\begin{tabular}{lcccccc}
\toprule
 & \multicolumn{2}{c}{Rank agreement} & Criterion & \multicolumn{3}{c}{$r$(sensitivity, external $d'$)} \\
\cmidrule(lr){2-3}\cmidrule(lr){4-4}\cmidrule(lr){5-7}
Readout & $\rho$(sens.) & $\rho$(crit.) & interval $>0$ & BFCL & QuestBench & When2Call \\
\midrule
\rowcolor{black!4}{Logistic, all questions ($a$, $\tau$)} & $1.00$    & $1.00$    & $16$ & $+0.70$ & $+0.69$ & $+0.67$ \\
Logistic, $\Delta U=0$ left out          & $0.98$    & $1.00$    & $17$ & $+0.69$ & $+0.65$ & $+0.65$ \\
\arrayrulecolor{black!25}\specialrule{0.3pt}{1pt}{1pt}\arrayrulecolor{black}
Signal detection, $d'$ and $c$               & $0.99$    & $0.77$    & $18$ & $+0.70$ & $+0.65$ & $+0.65$ \\
Signal detection, $d'$ and $c/d'$            & $0.99$    & $0.99$    & $18$ & $+0.70$ & $+0.65$ & $+0.65$ \\
\bottomrule
\end{tabular}
\end{center}
\end{table}

\subsection{Convergence with external benchmarks}
\label{app:check}
\label{app:convergence}

\begin{figure}[!ht]
\centering
\includegraphics[width=\linewidth]{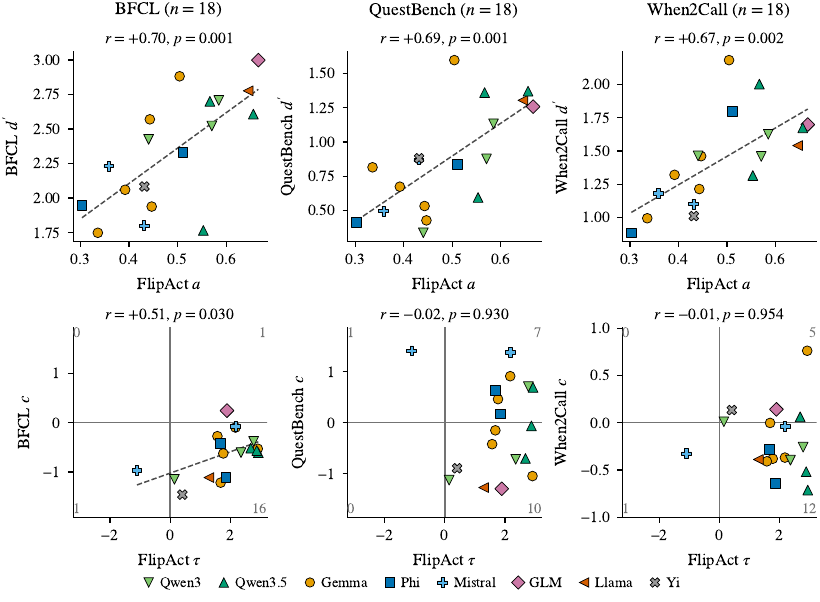}%
\caption{The two coordinates against their external counterparts for the panel models, on the three
benchmarks where sensitivity converges. Top: sensitivity against $d'$. Bottom: criterion against $c$,
with the number of models in each quadrant in gray. A dashed line marks a significant correlation.}
\label{fig:mtmm_all}
\end{figure}

This subsection gathers the evidence behind \S\ref{sec:sensitivity}. Table~\ref{tab:benchmarks} gives
each coordinate's correlation with its counterpart on the four external benchmarks, and
Figure~\ref{fig:mtmm_all} plots the panel models on the three where sensitivity converges.

\begin{table}[!ht]
\caption{Each \textsc{FlipAct} coordinate against its counterpart on the four external benchmarks of
\S\ref{sec:sensitivity}, over the panel of $18$ models.}
\label{tab:benchmarks}
\begin{center}\footnotesize
\setlength{\tabcolsep}{5pt}
\begin{tabular}{@{}lccccc@{}}
\toprule
 & \multicolumn{3}{c}{Sensitivity $a$ vs.\ $d'$} & \multicolumn{2}{c@{}}{Criterion $\tau$ vs.\ $c$} \\
\cmidrule(lr){2-4}\cmidrule(l){5-6}
Benchmark & $r$ & $p$ & 95\% interval & $r$ & $p$ \\
\midrule
BFCL \citep{patil2025bfcl} & $+0.70$ & $0.0013$ & $[+0.39,+0.90]$ & $+0.51$ & $0.030$ \\
QuestBench \citep{questbench2025}      & $+0.69$ & $0.0015$ & $[+0.46,+0.88]$ & $-0.02$ & $0.930$ \\
When2Call \citep{ross2025when2call}    & $+0.67$ & $0.0024$ & $[+0.40,+0.87]$ & $-0.01$ & $0.954$ \\
When2Tool \citep{when2tool2026}        & $+0.02$ & $0.9324$ & $[-0.50,+0.57]$ & $+0.19$ & $0.439$ \\
\bottomrule
\end{tabular}
\end{center}
{\footnotesize What each benchmark asks and where its answer key comes from are in
Table~\ref{tab:related-benchmarks}. Collapsing When2Call's choice among several replies to call or
withhold adds an offset to $c$ (App.~\ref{app:jev}), so external $c$ is not comparable across
benchmarks. The criterion correlation on BFCL rests on the three models with the lowest $\tau$: without
them $r=+0.21$. Its Spearman correlation is $+0.46$ ($p=0.058$), and Holm-adjusted over the four
benchmarks its $p$ is $0.12$.\par}
\end{table}

\paragraph{Panel cutoff.} The panel of \S\ref{sec:sensitivity} admits models with $a\ge0.30$. When that cutoff moves with the
other panel conditions fixed (Table~\ref{tab:check}), the correlation of $a$ with When2Call $d'$ stays
positive and significant at every cutoff. The criterion correlation is not stable: it is negative and
significant at $0.25$ and not significant at $0.30$ or $0.35$, so no direction is read from it.

\begin{table}[!ht]
\caption{The When2Call correlations of both coordinates at three panel cutoffs, over nested
panels. Shaded: the cutoff of the main text.}
\label{tab:check}
\begin{center}\footnotesize
\setlength{\tabcolsep}{5pt}
\begin{tabular}{cccccc}
\toprule
 & & \multicolumn{2}{c}{$a$ vs.\ $d'$} & \multicolumn{2}{c}{$\tau$ vs.\ $c$} \\
\cmidrule(lr){3-4}\cmidrule(lr){5-6}
Cutoff $a\ge$ & $n$ & $r$ & $p$ & $r$ & $p$ \\
\midrule
$0.25$ & $20$ & $+0.77$ & $0.0001$ & $-0.46$ & $0.041$ \\
\rowcolor{black!4}{$0.30$} & $18$ & $+0.67$ & $0.0024$ & $-0.01$ & $0.95$ \\
$0.35$ & $16$ & $+0.53$ & $0.033$  & $-0.29$ & $0.28$ \\
\bottomrule
\end{tabular}
\end{center}
\end{table}

\paragraph{A criterion-free readout.} Scoring each external example by its answer-token log-probability
gives an AUC with no criterion term (App.~\ref{app:readout}). Over the panel models with readable
log-probabilities, $a$ also correlates with AUC on QuestBench and When2Call, with $r=+0.70$ and $+0.81$,
so sensitivity converges with both benchmarks when the external measure carries no criterion term.

\subsection{Sensitivity and general capability}
\label{app:capability}

Separating warranted from unwarranted action on \textsc{FlipAct} takes physical knowledge as well as
reading the state, the tool and the question. Table~\ref{tab:capability} therefore checks the
convergence of \S\ref{sec:sensitivity} against four other indicators on the panel. Two are general: log
parameter count, and MMLU on a $570$-question subset, available for $15$ of the $18$ models. The other
two come from the answers $a$ is fitted on, so partialing them out removes part of $a$ itself:
\textsc{FlipAct} accuracy, which on the balanced main set equals balanced accuracy, and a
\textsc{FlipAct} $d'$ computed with Eq.~\ref{eq:sdt}. The convergence holds with size partialed out:
on the three benchmarks where $a$ converges, the partial correlation of $a$ with external $d'$ given log
parameters is $0.62$ to $0.64$. Given MMLU it is $0.52$ to $0.55$ ($p$ between $0.04$ and $0.06$ at
$n=15$). The \textsc{FlipAct} $d'$ correlates with $a$ at $0.96$, so the convergence does not depend on
how \textsc{FlipAct}'s sensitivity is summarized.

\begin{table}[!ht]
\caption{Correlation of $a$ with external $d'$ on the panel of $18$ models, raw and with each indicator
partialed out. The shaded row repeats the raw correlation on the $15$ models with MMLU.}
\label{tab:capability}
\begin{center}\footnotesize
\setlength{\tabcolsep}{6pt}
\begin{tabular}{lcccc}
\toprule
Partialed out & BFCL & QuestBench & When2Call & When2Tool \\
\midrule
none                      & $+0.70$ & $+0.69$ & $+0.67$ & $+0.02$ \\
\rowcolor{black!4}{none, $15$ models with MMLU} & $+0.71$ & $+0.75$ & $+0.72$ & $-0.27$ \\
\arrayrulecolor{black!25}\specialrule{0.3pt}{1pt}{1pt}\arrayrulecolor{black}
log parameters            & $+0.64$ & $+0.63$ & $+0.62$ & $-0.18$ \\
MMLU ($n=15$)             & $+0.53$ & $+0.55$ & $+0.52$ & $-0.67$ \\
\arrayrulecolor{black!25}\specialrule{0.3pt}{1pt}{1pt}\arrayrulecolor{black}
\textsc{FlipAct} accuracy & $+0.54$ & $+0.54$ & $+0.56$ & $-0.20$ \\
\textsc{FlipAct} $d'$     & $+0.19$ & $-0.11$ & $+0.08$ & $-0.37$ \\
\bottomrule
\end{tabular}
\end{center}
\end{table}

\subsection{Generations within the Qwen and Gemma lines}
\label{app:generations}

The generation, like size (\S\ref{sec:criterion}), does not set the criterion. The Qwen and Gemma
lines each span two or three generations at comparable sizes, released between mid-2024 and 2026
(Table~\ref{tab:generations}). From one generation to the next, sensitivity rises at small and middle
sizes, most from Qwen3 to Qwen3.5, and changes little at the largest sizes. The criterion stays put from
Qwen3-8B to Qwen3.5-9B and across the three Gemma generations at 27--31B, falls from gemma-2-9b to
gemma-3-12b and rises from Qwen3-32B to Qwen3.5-27B.

\begin{table}[!ht]
\caption{Coordinates of the Qwen and Gemma generations at comparable sizes, with the month the weights
were released. A $\tau$ beyond the score range is shown as a bound.}
\label{tab:generations}
\begin{center}\footnotesize
\setlength{\tabcolsep}{6pt}
\begin{tabular}{@{}lllccc@{}}
\toprule
Line & Size & Model & Released & $a$ & $\tau$ \\
\midrule
Qwen  & 4B       & Qwen3-4B     & 2025-04 & $0.34$ & ${>}{+}3$ \\
      &          & Qwen3.5-4B   & 2026-03 & $0.55$ & $+2.93$ \\
\arrayrulecolor{black!25}\cmidrule[0.3pt](l){2-6}\arrayrulecolor{black}
      & 8--9B    & Qwen3-8B     & 2025-04 & $0.44$ & $+2.78$ \\
      &          & Qwen3.5-9B   & 2026-03 & $0.66$ & $+2.88$ \\
\arrayrulecolor{black!25}\cmidrule[0.3pt](l){2-6}\arrayrulecolor{black}
      & 27--32B  & Qwen3-32B    & 2025-04 & $0.58$ & $+0.14$ \\
      &          & Qwen3.5-27B  & 2026-02 & $0.57$ & $+2.68$ \\
\arrayrulecolor{black!30}\midrule\arrayrulecolor{black}
Gemma & 9--12B   & gemma-2-9b   & 2024-06 & $0.34$ & $+2.91$ \\
      &          & gemma-3-12b  & 2025-03 & $0.39$ & $+2.17$ \\
\arrayrulecolor{black!25}\cmidrule[0.3pt](l){2-6}\arrayrulecolor{black}
      & 27--31B  & gemma-2-27b  & 2024-06 & $0.44$ & $+1.68$ \\
      &          & gemma-3-27b  & 2025-03 & $0.45$ & $+1.77$ \\
      &          & gemma-4-31b  & 2026-04 & $0.50$ & $+1.57$ \\
\bottomrule
\end{tabular}
\end{center}
\end{table}

\subsection{Reasoning mode}
\label{app:thinking}

Reasoning mode raises sensitivity only in the smallest models
\citep{gong2026think}. In larger ones it moves the criterion
toward acting. We compare reasoning off and on for every model that has both modes, in one paired run
with bootstrap intervals (Table~\ref{tab:thinking}), so the values with reasoning off can differ from
Table~\ref{tab:coords}, most for Qwen3-1.7B, whose fit there is flat. Reasoning raises $a$ significantly on
Qwen3-1.7B, Qwen3-4B, Nemotron-3-Nano-4B and Qwen3.5-4B, the models of 4B and below, and on no larger
model. It moves $\tau$ significantly toward zero on Qwen3.5-4B, Qwen3-8B, Qwen3.5-9B and
Qwen3.5-27B. The criterion falls less on Qwen3-14B, stays put on Granite-3.3-8B and rises slightly on
Qwen3-32B.
Qwen3.5-0.8B, Qwen3.5-2B, Qwen3.5-35B-A3B and Qwen3-0.6B have no usable fit in one of the two modes and
are omitted.

\begin{table}[!ht]
\caption{Reasoning mode off and on. Bold marks a significant rise in $a$ or fall in $\tau$.}
\label{tab:thinking}
\begin{center}\footnotesize
\setlength{\tabcolsep}{5pt}
\begin{tabular}{lcccccc}
\toprule
 & \multicolumn{3}{c}{$a$} & \multicolumn{3}{c}{$\tau$} \\
\cmidrule(lr){2-4}\cmidrule(lr){5-7}
Model & off & on & $\Delta$ & off & on & $\Delta$ \\
\midrule
\rowcolor{black!10}\multicolumn{7}{l}{\textbf{4B and below}} \\
Qwen3-1.7B       & $0.04$ & $0.42$ & $\mathbf{+0.38}$ & -- & $+1.05$ & -- \\
Qwen3-4B         & $0.32$ & $0.45$ & $\mathbf{+0.13}$ & -- & -- & -- \\
Nemotron-3-Nano-4B & $0.28$ & $0.49$ & $\mathbf{+0.21}$ & -- & $+2.42$ & -- \\
Qwen3.5-4B       & $0.56$ & $0.66$ & $\mathbf{+0.10}$ & $+2.87$ & $+1.91$ & $\mathbf{-0.96}$ \\
\rowcolor{black!10}\multicolumn{7}{l}{\textbf{8B and above}} \\
Qwen3-8B         & $0.44$ & $0.53$ & $+0.09$ & $+2.78$ & $+1.69$ & $\mathbf{-1.08}$ \\
Granite-3.3-8B   & $0.30$ & $0.37$ & $+0.07$ & $+2.59$ & $+2.61$ & $+0.02$ \\
Qwen3.5-9B       & $0.66$ & $0.61$ & $-0.04$ & $+2.86$ & $+1.99$ & $\mathbf{-0.87}$ \\
Qwen3-14B        & $0.57$ & $0.51$ & $-0.06$ & $+2.36$ & $+2.07$ & $-0.28$ \\
Qwen3.5-27B      & $0.57$ & $0.53$ & $-0.04$ & $+2.67$ & $+1.57$ & $\mathbf{-1.10}$ \\
Qwen3-32B        & $0.58$ & $0.51$ & $-0.07$ & $+0.14$ & $+0.40$ & $+0.26$ \\
\bottomrule
\end{tabular}
\end{center}
{\footnotesize A dash marks a $\tau$ outside the score range, or a change that involves one.\par}
\end{table}

\providecommand{\exY}{\textcolor{teal}{Yes}}
\providecommand{\exN}{No}
\tcbset{exbox/.style={enhanced,colback=black!2,colframe=black!35,boxrule=0.4pt,arc=1.5pt,left=4pt,
  right=4pt,top=3pt,bottom=3pt,fontupper=\small,coltitle=black,colbacktitle=black!8,
  fonttitle=\small\bfseries,toptitle=1pt,bottomtitle=1pt,before skip=6pt,after skip=6pt},
  exq/.style={enhanced,frame hidden,colback=black!5,boxrule=0pt,arc=1pt,left=3pt,right=3pt,top=2pt,
  bottom=2pt,fontupper=\footnotesize,before skip=0pt,after skip=3pt},
  exth/.style={enhanced,frame hidden,colback=black!3,boxrule=0pt,arc=1pt,left=3pt,right=3pt,top=2pt,
  bottom=2pt,fontupper=\footnotesize\itshape\color{black!70},before skip=3pt,after skip=0pt}}

Two answers from these runs show the two effects. Traces are quoted verbatim, with [\ldots] marking
omitted text. In Qwen3.5-4B the trace works out what the tool does to the state, and the model tells the
two states of one tool apart. In Qwen3.5-27B, reasoning turns a No on a warranted question into a Yes,
while the harmful state of the same tool stays No.

\begin{tcolorbox}[exbox,title={Reasoning separates the two states (Qwen3.5-4B)}]
\begin{tcolorbox}[exq]
Situation: The food or contents is [state]. The supplies this tool depends on (mains electricity) are
available nearby. The goal is to bring the food or contents back to a normal state. Action: a person
uses a hotplate. Is using the hotplate a wise choice in this situation?
\end{tcolorbox}
{\centering\footnotesize
\begin{tabular}{lcccc}
State & Gold & $\Delta U$ & Reasoning off & Reasoning on \\
very cold & \exY{} & $+2$ & \exY{} & \exY{} \\
very hot  & \exN{} & $-2$ & \exY{} & \exN{} \\
\end{tabular}\par}
\begin{tcolorbox}[exth]
\textup{Very hot:} Goal: Cool down (bring to normal state). [\ldots] Function of a hotplate: Heats
objects placed on it. Conflict: Using a heating device on something that is already very hot will make
it hotter, not cooler. This is the opposite of the goal.\\[1pt]
\textup{Very cold:} Goal: Reheat/warm up cold food. [\ldots] Logic: If electricity is available and the
goal is to heat food, a hotplate is an appropriate and effective tool for that purpose.
\end{tcolorbox}
\end{tcolorbox}

\begin{tcolorbox}[exbox,title={Reasoning moves the criterion toward acting (Qwen3.5-27B)}]
\begin{tcolorbox}[exq]
Situation: The rigid object or joint is [state]. The goal is to bring the rigid object or joint back to
a normal state. Action: a person uses a rivet gun. Is using the rivet gun a wise choice in this
situation?
\end{tcolorbox}
{\centering\footnotesize
\begin{tabular}{lcccc}
State & Gold & $\Delta U$ & Reasoning off & Reasoning on \\
very loose and coming apart     & \exY{} & $+2$ & \exN{} & \exY{} \\
very jammed and over-tightened  & \exN{} & $-2$ & \exN{} & \exN{} \\
\end{tabular}\par}
\begin{tcolorbox}[exth]
\textup{Very loose:} If a joint is loose and coming apart, permanently fastening it with a rivet is a
standard and effective repair method for many rigid objects (metal sheets, frames, etc.). [\ldots]
Sometimes bolts or welding might be used, but a rivet gun is definitely a \textup{wise} and common
choice for fixing loose joints where rivets are applicable.
\end{tcolorbox}
\end{tcolorbox}

\section{The criterion under reversed question polarity}
\label{app:polarity}

\paragraph{Design.} If a model prefers one answer token whatever is asked, that preference enters its
criterion (\S\ref{sec:criterion}), and no single wording can separate the two. The standard probe is a balanced key
\citep{cronbach1946response}: the same content is asked in both polarities, and one reads whether the
answer follows the content or the answer word. The canonical wording closes each main-set question by asking
whether using the tool is a wise choice (``wise'' below). We replace that sentence with three reversed
wordings, under which No means act (``Would using $\{t\}$ be an unwise choice'', ``Should the person
avoid using $\{t\}$'', ``Is using $\{t\}$ a mistake''), and ask the canonical wording in the same run.
Answers under a reversed wording are recoded, so every fit is $P(\text{act}\mid\Delta U)$ and a model
that answers the content returns the same $a$ and $\tau$. Five same-polarity rephrasings (appropriate,
should use, right tool, makes sense, would recommend) and three answer-order variants (``Answer No or
Yes'', and letter options with Yes first or with No first) check both coordinates under rewording. The
run used one seed on $16$ of the $18$ panel models, and every rewording is compared with the canonical
wording asked in the same run.

\paragraph{Decomposition.} With both fits recoded to act polarity, write
$\tau_{\text{wise}}=\tau_d+b$ and $\tau_{\text{unwise}}=\tau_d-b$. $\tau_d$ is a threshold on the
evidence, attached to the proposition, and $b$ is a response bias attached to the answer word, in score
units and positive when the model prefers No. Table~\ref{tab:polarity_sign} gives both as means over the
three reversed wordings.

\paragraph{Result.} Seven models keep the sign of $\tau$ under all three reversed wordings
(Table~\ref{tab:polarity_sign}). For the three Gemma models among them, phi-4 and Mistral-24B, the reversed $\tau$ is
more conservative than the canonical one: asked whether the action is unwise or a mistake, they answer
Yes on $75$--$90\%$ of questions, the same stance as their $19$--$36\%$ Yes-rate on ``wise''. The three
Qwen3.5 models and GLM-4-32B flip under all three. They answer No under either polarity, with a Yes-rate
of $15$--$20\%$ on ``wise'' and $9$--$26\%$ on the reversed wordings (one model--wording pair at
$43\%$), which recoded is an act-rate of $0.75$--$0.9$. These four have $b$ of $+3$ to $+4$ and
$\tau_d$ within a unit of zero, so their conservatism on the canonical wording is a preference for No.
The other models stay conservative on $\tau_d$: the seven that keep their sign have $b$ between
$-0.97$ and $+0.58$, and $\tau_d$ at or above their canonical $\tau$ on six of the seven.

\begin{table}[!ht]
\caption{Sign of the criterion under the three reversed wordings, and its split into a threshold on the
evidence $\tau_d$ and an answer preference $b$.}
\label{tab:polarity_sign}
\begin{center}\footnotesize
\setlength{\tabcolsep}{6pt}
\begin{tabular}{lccccccc}
\toprule
 & \multicolumn{2}{c}{Canonical wording} & \multicolumn{3}{c}{Sign kept under} & \multicolumn{2}{c}{Decomposition} \\
\cmidrule(lr){2-3}\cmidrule(lr){4-6}\cmidrule(lr){7-8}
Model & $\tau$ & Yes-rate & unwise & avoid & mistake & $\tau_d$ & $b$ \\
\midrule
\rowcolor{black!10}\multicolumn{8}{l}{\textbf{Sign kept under all three wordings}} \\
Yi-1.5-34B         & $+0.32$ & $0.48$ & \cmark & \cmark & \cmark & $+0.62$ & $-0.30$ \\
Qwen3-14B          & $+2.62$ & $0.23$ & \cmark & \cmark & \cmark & $+2.04$ & $+0.58$ \\
gemma-2-9b         & $+3.65$\rlap{$^\dagger$} & $0.21$ & \cmark & \cmark & \cmark & $+4.55$\rlap{$^\dagger$} & $-0.90$ \\
gemma-2-27b        & $+2.50$ & $0.25$ & \cmark & \cmark & \cmark & $+2.80$ & $-0.30$ \\
gemma-3-27b        & $+2.45$ & $0.27$ & \cmark & \cmark & \cmark & $+3.42$\rlap{$^\dagger$} & $-0.97$ \\
phi-4              & $+1.45$ & $0.36$ & \cmark & \cmark & \cmark & $+2.19$ & $-0.74$ \\
Mistral-24B        & $+3.33$\rlap{$^\dagger$} & $0.19$ & \cmark & \cmark & \cmark & $+3.62$\rlap{$^\dagger$} & $-0.29$ \\
\rowcolor{black!10}\multicolumn{8}{l}{\textbf{Mixed or degenerate}} \\
gemma-3-12b        & $+2.31$ & $0.30$ & --     & --     & \cmark & $+2.44$ & $-0.13$ \\
Llama-3.3-70B      & $+1.25$ & $0.36$ & \cmark & \xmark & \cmark & $+1.51$ & $-0.26$ \\
Qwen3-32B          & $+1.05$ & $0.39$ & \cmark & \xmark & \cmark & $+0.63$ & $+0.41$ \\
Qwen3-8B           & $+2.34$ & $0.29$ & --     & --     & --     & $+0.95$ & $+1.39$ \\
Phi-3.5-mini       & $+3.66$\rlap{$^\dagger$} & $0.24$ & -- & -- & -- & $+4.83$\rlap{$^\dagger$} & $-1.17$ \\
\rowcolor{black!10}\multicolumn{8}{l}{\textbf{Sign flipped under all three wordings}} \\
Qwen3.5-4B         & $+3.57$\rlap{$^\dagger$} & $0.15$ & \xmark & \xmark & \xmark & $-0.42$ & $+3.99$ \\
Qwen3.5-9B         & $+3.37$\rlap{$^\dagger$} & $0.15$ & \xmark & \xmark & \xmark & $+0.32$ & $+3.04$ \\
Qwen3.5-27B        & $+3.07$\rlap{$^\dagger$} & $0.18$ & \xmark & \xmark & \xmark & $-0.17$ & $+3.24$ \\
GLM-4-32B          & $+2.56$ & $0.20$ & \xmark & \xmark & \xmark & $-0.86$ & $+3.42$ \\
\bottomrule
\end{tabular}
\end{center}
{\footnotesize Canonical-wording values come from the polarity run, not from Table~\ref{tab:coords}.
\cmark\ sign of $\tau$ kept, \xmark\ flipped, -- reversed fit degenerate. $^\dagger$ outside the $\pm3$
score range, or a mean including such a fit, read for sign only.}
\end{table}

The reversed wordings are the noisier conditions, so $\tau$ remains the reported coordinate and $\tau_d$
a diagnostic that names the models whose conservatism on the canonical wording is an answer preference. On
a model with no answer token, the JEV model of App.~\ref{app:jev}, the decomposition gives
$b=-0.20\pm0.24$, so it does not produce an answer preference where none can exist.

\paragraph{Rewording without reversing polarity.} The conservative criterion is not a product of the word
``wise''. Under each of five same-polarity rephrasings the criterion stays positive on $15$ or all $16$
models, and it ranks models much as the canonical wording does (Table~\ref{tab:rephrase}). Sensitivity also
keeps its ranking, at $\rho(a)=0.87$--$0.94$ for the five rephrasings, ``Answer No or Yes'' and letter
options with Yes first, and at $0.58$ for letter options with No first.

\begin{table}[!ht]
\caption{Both coordinates under five same-polarity rephrasings of the canonical question, $16$ panel
models. Rank correlations are with the canonical wording asked in the same run, on which all $16$
criteria are positive.}
\label{tab:rephrase}
\begin{center}\footnotesize
\setlength{\tabcolsep}{6pt}
\begin{tabular}{lcccc}
\toprule
Rephrasing & $\tau>0$ & $\rho(\tau)$ & Median $\Delta\tau$ & $\rho(a)$ \\
\midrule
``appropriate''      & $16$ & $0.86$ & $-0.63$ & $0.90$ \\
``should use''       & $15$ & $0.89$ & $-0.54$ & $0.91$ \\
``the right tool''   & $16$ & $0.77$ & $-0.08$ & $0.94$ \\
``makes sense''      & $16$ & $0.96$ & $-0.87$ & $0.88$ \\
``would recommend''  & $16$ & $0.88$ & $-0.15$ & $0.88$ \\
\bottomrule
\end{tabular}
\end{center}
\end{table}

\section{Moving the criterion: settings and further results}
\label{app:interventions}

This section gives the settings behind \S\ref{sec:interventions} and further results: steering
(App.~\ref{app:s5figs} and~\ref{app:crossds}), the two training schemes and the values behind
\S\ref{sec:transfer} (App.~\ref{app:train}), a criterion-free readout (App.~\ref{app:readout}), answer
preference (App.~\ref{app:yesbias}), a per-benchmark choice of steering sign and strength (App.~\ref{app:calib}) and single
answers (App.~\ref{app:examples}), in the order of \S\ref{sec:interventions}.

\subsection{Steering}
\label{app:s5figs}

This subsection specifies the criterion direction of \S\ref{sec:steer} and its strength rule, then checks
the injection on a held-out half of \textsc{FlipAct}, on held-out When2Call examples and on eleven more
models.

The direction is a probe of the model's own answers. For each of the sixteen models of
Tables~\ref{tab:steer} and~\ref{tab:steer_more}, an L2-regularized logistic regression ($C=0.1$) maps the
standardized residual-stream activation at the final token, taken at the layer at $0.6$ of the model's depth,
to the model's own Yes/No answer on every main-set question. No labels enter the fit. Its weight vector,
scaled to unit norm, is $\hat w$. The hook adds $\lambda\,\hat w\,\bar r/\sqrt{d}$ to the final-token
activation at the output of that layer, where $\bar r$ is the mean norm of the activation over the fit questions and
$d$ is the width, so that $\lambda$ is in units of the model's own scale.

\begin{table}[!ht]
\caption{Steering on When2Call for all $16$ models, grouped by the tests the shift passes. Negative shift
is toward acting, and $^\ast$ marks a model of Table~\ref{tab:steer}.}
\label{tab:steer_more}
\begin{center}\footnotesize
\setlength{\tabcolsep}{5pt}
\begin{tabular}{lrccrc}
\toprule
 & & & & \multicolumn{2}{c}{Against controls} \\
\cmidrule(lr){5-6}
Model & \multicolumn{1}{c}{$\lambda^\star$} & Fitted $\Delta c$ & $q$ & \multicolumn{1}{c}{Ratio} & Retained \\
\midrule
\rowcolor{black!10}\multicolumn{6}{l}{\textbf{Significant, three times the largest control, half retained, monotone in $\lambda$}} \\
Command-R-08-2024              & $32$  & $-1.04$ & $<0.001$ & $4.8$  & $1.01$ \\
Mistral-24B$^\ast$             & $64$  & $-1.14$ & $<0.001$ & $4.9$  & $1.09$ \\
Qwen3-8B$^\ast$                & $16$  & $-0.27$ & $<0.001$ & $3.7$  & $0.91$ \\
gemma-2-9b$^\ast$              & $64$  & $-0.65$ & $0.004$  & $3.4$  & $0.69$ \\
\rowcolor{black!10}\multicolumn{6}{l}{\textbf{Significant, at least one other test not passed}} \\
gemma-3-12b                    & $4$   & $-0.81$ & $0.001$  & $7.8$  & $0.77$ \\
Llama-3.3-70B                  & $48$  & $-0.16$ & $0.027$  & $18.5$ & $0.63$ \\
Yi-1.5-34B                     & $16$  & $-0.30$ & $<0.001$ & $1.6$  & $2.83$ \\
phi-4$^\ast$                   & $32$  & $-0.11$ & $0.004$  & $1.8$  & $0.83$ \\
Qwen3.5-9B                     & $64$  & $-0.12$ & $<0.001$ & $1.7$  & $1.07$ \\
Qwen3.5-35B-A3B                & $96$  & $-0.16$ & $<0.001$ & $1.0$  & $1.21$ \\
gemma-3-27b                    & $-32$ & $+0.05$ & $0.037$  & $0.1$  & $1.25$ \\
\rowcolor{black!10}\multicolumn{6}{l}{\textbf{Not significant}} \\
Qwen3.5-27B$^\ast$             & $96$  & $-0.02$ & $0.052$  & $0.2$  & $0.50$ \\
Qwen3-14B                      & $16$  & $-0.34$ & $0.62$   & $3.2$  & $0.79$ \\
Qwen3-32B                      & $96$  & $-0.16$ & $0.15$   & $0.7$  & $0.97$ \\
GLM-4-32B                      & $-16$ & $-0.06$ & $1.0$    & $4.1$  & $1.08$ \\
Qwen3-30B-A3B-Thinking-2507    & \multicolumn{5}{c}{no strength inside the $d'$-stable range} \\
\bottomrule
\end{tabular}
\end{center}
{\footnotesize Fitted $\Delta c$: change in $c$ across the $d'$-stable strengths, which differs from the
change at $\lambda^\star$ in Table~\ref{tab:steer}. On gemma-3-27b the paired test registers more acting
at $\lambda^\star$ while the fitted shift across the range is near zero.}
\end{table}

The direction depends on the wording it is read under. A probe refitted under the reversed wording
``unwise'' of App.~\ref{app:polarity} has cosine $-0.03$, $+0.06$ and $+0.02$ with the canonical one on
Qwen3.5-9B, Qwen3.5-27B and Qwen3-8B, and $-0.34$ on gemma-3-12b. Steering results are therefore read
from behavior, the shift in $c$ and $\tau$ under injection, and not from the geometry of the direction.

\begin{figure}[!ht]
\centering
\includegraphics[width=\linewidth]{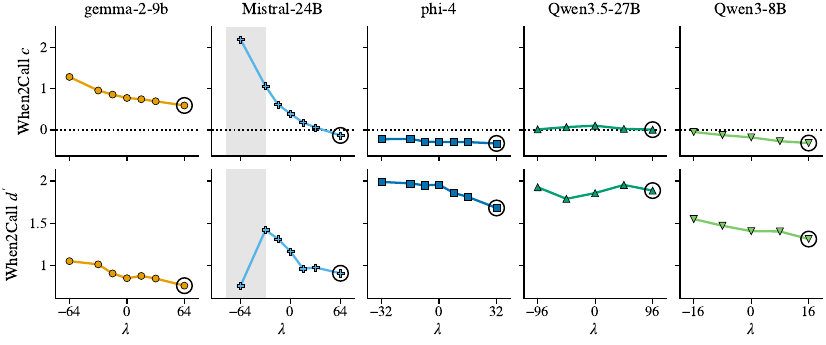}%
\caption{When2Call criterion $c$ (top) and $d'$ (bottom) against the injected strength $\lambda$ for the
five models of Table~\ref{tab:steer}. Each column has its own range of $\lambda$. Gray marks strengths
outside the $d'$-stable range, and circles mark $\lambda^\star$.}
\label{fig:steer}
\end{figure}

The strength is chosen on When2Call itself. Each model is run over a grid of strengths $\lambda$, which
includes negative values that inject $-\hat w$. The strengths at which When2Call $d'$ stays within $25\%$
of its unsteered value form the $d'$-stable range, and $\lambda^\star$ is the strength in that range with
the lowest criterion $c$. Figure~\ref{fig:steer} shows the strength response of the five models of
Table~\ref{tab:steer}.

On \textsc{FlipAct} the injection is checked with a held-out readout. The quads are split at random into
halves, and this split is not the one used for training (App.~\ref{app:train}). The direction and a
strength chosen by accuracy are fit on one half, and $c$ and $d'$ are computed from the binary answers on
the other. On the read half the injection moves the criterion of Qwen3.5-9B and Qwen3.5-27B toward zero,
leaves $d'$ within a few hundredths of its value and raises accuracy by about two points
(Table~\ref{tab:steer_flipact}).

\begin{table}[!ht]
\caption{Steering on the held-out half of \textsc{FlipAct}, with the direction and strength fit on the
other half.}
\label{tab:steer_flipact}
\begin{center}\footnotesize
\setlength{\tabcolsep}{7pt}
\begin{tabular}{@{}lccccc@{}}
\toprule
 & \multicolumn{2}{c}{Criterion $c$} & & \multicolumn{2}{c@{}}{Accuracy} \\
\cmidrule(lr){2-3}\cmidrule(lr){5-6}
Model & before & $+$ steering & $\Delta d'$ & before & $\Delta$ \\
\midrule
Qwen3.5-9B  & $+0.75$ & $+0.42$ & $-0.02$ & $0.654$ & $+0.026$ \\
Qwen3.5-27B & $+0.57$ & $+0.39$ & $+0.03$ & $0.642$ & $+0.016$ \\
\bottomrule
\end{tabular}
\end{center}
\end{table}

Table~\ref{tab:steer} chooses $\lambda^\star$ and reads its effect on the same When2Call examples. A
split-half check separates the choice from the readout. The examples are split into halves twenty times
at random, keeping the call/withhold balance, and on each split $\lambda^\star$ is chosen on one half by
the same rule and the shift is read on the other (Table~\ref{tab:steer_split}). For four of the five
models the held-out shift in $c$ matches the full-set one to within $0.02$, and on every split the
criterion falls and the call rate rises. phi-4, whose full-set shift is the smallest, starts permissive
and shows no reliable shift on the held-out half.

\begin{table}[!ht]
\caption{Split-half check of the When2Call steering shift: $\lambda^\star$ chosen on one half and read on
the other, beside the full-set value.}
\label{tab:steer_split}
\begin{center}\footnotesize
\setlength{\tabcolsep}{2.6pt}
\begin{tabular}{llccccc}
\toprule
 & & gemma-2-9b & Mistral-24B & phi-4 & Qwen3.5-27B & Qwen3-8B \\
\midrule
\rowcolor{black!4}$\Delta c$ & {full} & $-.18$ & $-.53$ & $-.03$ & $-.10$ & $-.14$ \\
 & held-out & $-.18$ & $-.53$ & $-.01$ & $-.08$ & $-.14$ \\
 & [2.5, 97.5] & $[-.25,-.13]$ & $[-.61,-.29]$ & $[-.06,+.11]$ & $[-.11,-.04]$ & $[-.21,-.08]$ \\
\arrayrulecolor{black!25}\specialrule{0.3pt}{1pt}{1pt}\arrayrulecolor{black}
\rowcolor{black!4}$\Delta$ call rate & {full} & $+.052$ & $+.178$ & $+.019$ & $+.025$ & $+.045$ \\
 & held-out & $+.048$ & $+.175$ & $+.012$ & $+.022$ & $+.047$ \\
 & [2.5, 97.5] & $[+.035,+.068]$ & $[+.096,+.210]$ & $[-.026,+.027]$ & $[+.010,+.027]$ & $[+.028,+.067]$ \\
\arrayrulecolor{black!25}\specialrule{0.3pt}{1pt}{1pt}\arrayrulecolor{black}
\rowcolor{black!4}$\Delta$ acc. & {full} & $+.002$ & $-.032$ & $-.038$ & $+.005$ & $-.022$ \\
 & held-out & $+.000$ & $-.035$ & $-.028$ & $+.007$ & $-.020$ \\
 & [2.5, 97.5] & $[-.020,+.013]$ & $[-.050,-.007]$ & $[-.050,+.003]$ & $[-.003,+.012]$ & $[-.043,-.003]$ \\
\arrayrulecolor{black!25}\specialrule{0.3pt}{1pt}{1pt}\arrayrulecolor{black}
Same $\lambda^\star$ & & $20/20$ & $14/20$ & $14/20$ & $10/20$ & $20/20$ \\
\bottomrule
\end{tabular}
\end{center}
{\footnotesize Full (shaded): the change of Table~\ref{tab:steer}, computed before rounding. Held-out: median over
$20$ stratified splits and its [2.5, 97.5] percentile range. Same $\lambda^\star$: splits whose chosen
strength equals the full-set one.}
\end{table}

Table~\ref{tab:steer_more} adds eleven further models, each run at its selected strength only. The
fitted shift is the change in $c$ across the $d'$-stable strengths, and $q$ is a paired McNemar test on
the withhold examples after Benjamini--Hochberg correction. Two controls go with each shift. The first
runs three directions of the same norm, one orthogonal to $\hat w$ and two random, and the ratio column
divides the shift by the largest of their shifts. The second projects the refusal
\citep{arditi2024refusal}, sycophancy \citep{rimsky2024steering} and a generic engagement direction out
of $\hat w$, and the retained column is the share of the shift that remains. The engagement direction is the difference in mean activation between prompts the model answers and prompts it refuses or hedges, taken from XSTest \citep{rottger2024xstest} and a set of abstention questions. On the paired test, eleven
of the sixteen models move significantly toward acting after correction for the fifteen testable
comparisons. Four of these also clear both controls and respond monotonically in $\lambda$, with a shift
at least three times the largest control shift and at least half of it kept after the projection.

\subsection{Steering on further benchmarks}
\label{app:crossds}

To test whether the steering result of \S\ref{sec:steer} is specific to When2Call, the same direction
$\hat w$ is injected on four further benchmarks for three of the models, with the injection layer,
readout and strength rule of App.~\ref{app:s5figs}. The shift in $c$ is read inside the $d'$-stable
range, and the paired McNemar test on the withhold examples is judged after Benjamini--Hochberg
correction. The direction moves the criterion toward acting, with a significant paired test, on every
benchmark that can be scored for Qwen3-8B and Mistral-24B, and on one benchmark of four for Qwen3.5-27B
(Table~\ref{tab:crossds}). Two of Mistral-24B's four benchmarks have no strength at which $d'$ stays
stable, and Qwen3.5-27B's response on When2Tool and BFCL-miss-func is not monotone in $\lambda$.

\begin{table}[!ht]
\caption{The \textsc{FlipAct} criterion direction on four further benchmarks with the weights fixed:
shift in $c$ inside the $d'$-stable range, negative toward acting.}
\label{tab:crossds}
\begin{center}\footnotesize
\setlength{\tabcolsep}{8pt}
\begin{tabular}{@{}lcccc@{\hspace{0.9em}}}
\toprule
Model & When2Tool & BFCL-miss-func & ACPBench & CaT-Bench \\
\midrule
Qwen3-8B    & $-1.08$\rlap{\,$\bullet$} & $-0.54$\rlap{\,$\bullet$} & $-1.11$\rlap{\,$\bullet$} & $-0.80$\rlap{\,$\bullet$} \\
Mistral-24B & no stable $\lambda$       & $-1.95$\rlap{\,$\bullet$} & $-1.94$\rlap{\,$\bullet$} & no stable $\lambda$ \\
Qwen3.5-27B & $-1.09$                   & $-0.24$                   & $-0.47$                   & $-0.74$\rlap{\,$\bullet$} \\
\bottomrule
\end{tabular}
\end{center}
{\footnotesize $\bullet$ significant paired McNemar test on the withhold examples after
Benjamini--Hochberg correction. No stable $\lambda$: no strength inside the $d'$-stable range.}
\end{table}

\subsection{Training}
\label{app:train}

Both training schemes, criterion distillation (\S\ref{sec:train}) and sensitivity learning
(\S\ref{sec:link}), share the split, adapter and significance rule below. Their settings and controls
come first, then the external-benchmark values behind \S\ref{sec:transfer}.

\paragraph{Split and significance.} The main set is split at the connected components of the graph that
links situations to tool records, so no tool record, situation or state flip appears in both halves.
Both schemes use this split, with the sizes in Table~\ref{tab:hparams}. Every reported change is the
mean over three seeds. Its error is the largest of the seeds' paired situation-cluster bootstrap errors,
and a change is significant when it is at least twice that error. Table~\ref{tab:trainseeds} lists every
change with its error.

\begin{table}[!ht]
\caption{Training settings of the two schemes.}
\label{tab:hparams}
\begin{center}\footnotesize
\setlength{\tabcolsep}{4pt}
\begin{tabular}{l>{\raggedright\arraybackslash}p{5.3cm}>{\raggedright\arraybackslash}p{5.85cm}}
\toprule
\rowcolor{black!10}\multicolumn{3}{l}{\textbf{Both schemes}} \\
Split & \multicolumn{2}{>{\raggedright\arraybackslash}p{11.43cm}}{$63$ connected components: $940$ training and $940$ held-out questions ($247$ and $248$ state flips)} \\
\rowcolor{black!3}Adapter & \multicolumn{2}{>{\raggedright\arraybackslash}p{11.43cm}}{LoRA with $r=16$, $\alpha=32$ and dropout $0.05$ on the q, k, v, o, gate, up and down projections (on gemma-3-12b, of the language-model layers only)} \\
Precision & \multicolumn{2}{l}{bf16 weights} \\
\rowcolor{black!3}Clipping & \multicolumn{2}{l}{gradient norm $1$} \\
Seeds & \multicolumn{2}{l}{$0$, $1$ and $2$} \\
\rowcolor{black!10} & \textbf{Criterion distillation} & \textbf{Sensitivity learning} \\
Data & $940$ training-half questions & $1{,}203$ questions ($940$ + $263$ supplementary), Yes share $0.39$ \\
\rowcolor{black!3}Target & steered teacher's next-token logits & computed Yes/No answer \\
Strength $\lambda^\star$ & Qwen3-8B $22$, Qwen3.5-9B $32$, gemma-3-12b $2$, Mistral-24B $16$ & -- \\
\rowcolor{black!3}Control & random direction of the same norm at the same $\lambda^\star$ & answers permuted within $255$ strata of matched questions \\
Loss & forward KL, $T=1$, full vocabulary, answer position & cross-entropy on the answer turn, prompt masked \\
\rowcolor{black!3}Optimizer & AdamW, weight decay $0.01$, lr $10^{-5}$, constant & AdamW, weight decay $0$, lr $2\times10^{-4}$, linear decay, no warm-up \\
Effective batch & $16$ ($8\times2$; Mistral-24B $4\times4$) & $16$ ($2\times8$; Mistral-24B $1\times16$) \\
\rowcolor{black!3}Length & $4$ epochs, $235$ steps & $2$ epochs, $152$ steps \\
Max.\ length & $1{,}024$ prompt tokens, $1$ output token & $2{,}048$ tokens \\
\bottomrule
\end{tabular}
\end{center}
\end{table}

\paragraph{Criterion distillation.} The teacher is the student itself with a steering hook. The hook
adds $\lambda^\star\hat w\,\bar r/\sqrt{d}$ at the $0.6$-depth layer, as in App.~\ref{app:s5figs}, but
here $\hat w$ is a difference of means: the mean standardized activation of the training-half questions the
student answers Yes, minus that of the questions it answers No, scaled to unit norm. No labels enter it.
The loss is the forward KL of Eq.~\ref{eq:distil}, with the teacher's logits computed once on all
training-half questions, and Table~\ref{tab:hparams} gives the control and the optimization settings.
Every criterion-distillation run of the two non-Qwen students changes MMLU by $-0.004$ to $+0.006$.

The strength comes from a \textsc{FlipAct} strength sweep with the same estimator. A strength qualifies
when the criterion is well determined, $a$ stays within two bootstrap errors of its unsteered value, $d'$ stays
within a quarter of its value, and the Yes-rate stays inside $[0.10,0.90]$. Among the qualifying
strengths, $\lambda^\star$ is the one with the largest $|\Delta\tau|$ (Table~\ref{tab:hparams}). The
sweep's direction is fit on a random half of the quads ($955$ questions, $479$ of them in the held-out
half) and agrees with the teacher's direction, with a cosine of $0.97$ to $0.99$. The sweep's $a$,
$\tau$, $d'$ and Yes-rate are computed on all questions. On the training-half questions alone, the rule
returns the same strength for Qwen3-8B, Qwen3.5-9B and gemma-3-12b, and $22$ instead of $16$ for
Mistral-24B.\footnote{On the full set $\lambda=22$ is excluded because its $\tau$ of $-3.05$ falls just
outside the score range.} On gemma-3-12b only one nonzero strength qualifies, $\lambda=+2$, with a
Yes-rate of $0.47$. At $+4$ the Yes-rate is $0.81$ and $\tau$ falls outside the score range. At every negative
strength $\tau$ is also outside it or the model answers only No.

\paragraph{Control strength.} At the teacher's norm, a random direction changes the Yes--No logit margin
far less than $\hat w$ does (Table~\ref{tab:placebo_match}), so it is a weak control. We therefore scale
two random directions per student until their mean absolute change in the margin, over all
\textsc{FlipAct} questions, equals the teacher's, and refit both coordinates there
(Table~\ref{tab:randmatch}). On gemma-3-12b, on Mistral-24B's first direction and on Qwen3.5-9B's
second, the matched random direction perturbs the answers as much as $\hat w$ and leaves $\tau$ where it
was, while $\hat w$ moves it well toward acting. The other four
directions collapse the model onto one answer
before they match the teacher's perturbation, with a Yes-rate of $0$ or $1$. No matched random direction
shifts the criterion toward acting with sensitivity intact.

\begin{table}[!ht]
\caption{Perturbation of the Yes--No logit margin at each student's selected strength, by the criterion
teacher and by a random direction of the same norm.}
\label{tab:placebo_match}
\begin{center}\footnotesize
\setlength{\tabcolsep}{6pt}
\begin{tabular}{@{}lrcr@{}}
\toprule
 & \multicolumn{2}{c}{Mean $|\Delta|$ Yes--No margin} & \\
\cmidrule(lr){2-3}
Student & \multicolumn{1}{c}{teacher} & same-norm random & \multicolumn{1}{c@{}}{Ratio} \\
\midrule
Qwen3-8B & $18.70$ & $1.18$ & $15.8$ \\
Qwen3.5-9B & $2.48$ & $0.31$ & $7.9$ \\
gemma-3-12b & $4.22$ & $0.46$ & $9.2$ \\
Mistral-24B & $3.24$ & $0.08$ & $40.4$ \\
\bottomrule
\end{tabular}
\end{center}
\end{table}

\begin{table}[!ht]
\caption{The criterion teacher beside two random directions per student, each scaled until it perturbs
the Yes--No logit margin as much as the teacher does, with both coordinates refit at that strength.}
\label{tab:randmatch}
\begin{center}\footnotesize
\setlength{\tabcolsep}{4.5pt}
\begin{tabular}{@{}lccclcccc@{}}
\toprule
 & & \multicolumn{2}{c}{Criterion teacher} & \multicolumn{5}{c@{}}{Matched random direction} \\
\cmidrule(lr){3-4}\cmidrule(l){5-9}
Student & Base $\tau$ & $|\Delta|$ margin & $\tau$ & & $|\Delta|$ margin & $\tau$ & $d'$ & Yes-rate \\
\midrule
Qwen3-8B & $+2.67$ & $18.70$ & $-2.11$ & random dir.\ 1 & $16.41$ & -- & $0.00$ & $0.00$ \\
& & & & random dir.\ 2 & $17.16$ & -- & $0.00$ & $1.00$ \\
\arrayrulecolor{black!25}\specialrule{0.3pt}{1pt}{1pt}\arrayrulecolor{black}
Qwen3.5-9B & $+2.19$ & $2.48$ & $+1.08$ & random dir.\ 1 & $2.62$ & -- & $0.00$ & $0.00$ \\
& & & & random dir.\ 2 & $2.46$ & $+2.16$ & $1.00$ & $0.24$ \\
\arrayrulecolor{black!25}\specialrule{0.3pt}{1pt}{1pt}\arrayrulecolor{black}
gemma-3-12b & $+2.50$ & $4.22$ & $+0.83$ & random dir.\ 1 & $4.35$ & $+2.87$ & $0.64$ & $0.26$ \\
& & & & random dir.\ 2 & $4.31$ & $+2.53$ & $0.63$ & $0.28$ \\
\arrayrulecolor{black!25}\specialrule{0.3pt}{1pt}{1pt}\arrayrulecolor{black}
Mistral-24B & ${>}{+}3$ & $3.24$ & $+0.55$ & random dir.\ 1 & $3.18$ & ${>}{+}3$ & $0.88$ & $0.12$ \\
& & & & random dir.\ 2 & $3.17$ & ${>}{+}3$ & $0.49$ & $0.00$ \\
\bottomrule
\end{tabular}
\end{center}
{\footnotesize $|\Delta|$ margin: mean absolute change in the Yes--No logit margin over all
\textsc{FlipAct} questions, as in Table~\ref{tab:placebo_match}. A dash marks a fit that cannot be
estimated because the model answers one way.}
\end{table}

\paragraph{What the student copies.} A teacher moved only in the criterion can be distilled because a
state flip separates the two coordinates. Write $\ell_+$ and $\ell_-$ for the Yes--No logit margins on the Yes
and the No question of a flip. Their difference $D=\ell_+-\ell_-$ measures how far apart the model holds
the two states and tracks $a$. Their sum $\Sigma=\ell_++\ell_-$ measures how far both sit from zero
and tracks $\tau$. A teacher moved only in $\tau$ differs from the student in $\Sigma$ and not in
$D$. The students reproduce $0.88$ to $1.04$ of their teacher's shift in $\Sigma$. The random-direction teachers
move $\Sigma$ far less than the criterion teachers, and their students' shifts in $\Sigma$ stay small.

\paragraph{Sensitivity learning.} The student trains answer-only on the computed answers, with the same
LoRA configuration. Its pool is the training half plus supplementary questions from the same tool
records (sizes and Yes share in Table~\ref{tab:hparams}). The control permutes the answers within strata
of matched questions, so prompts, format, step count and Yes-rate are unchanged and about half of the
training state flips have their two answers swapped. Shuffling removes the link between a question's
state and its answer, and on seven of the twelve runs the student's answers no longer carry a criterion:
three runs collapse onto one answer, three lose the slope, and one has $\mathrm{SE}(\tau)$ just above two. Answer-only training also
breaks free-form output on Mistral-24B, so several of its external readouts, mostly on BFCL-miss-func and
When2Call, cannot be scored. MMLU cannot be
read on five of the twelve sensitivity-learning runs.

\begin{table}[h]
\caption{Held-out change for every training condition of Table~\ref{tab:train}, as the seed mean
$\pm$ the largest per-seed paired bootstrap error. Bold marks a change toward acting in $\tau$, a rise in
$a$ after sensitivity learning and a rise in accuracy. A $\ddagger$ marks a cell with two usable seeds,
and n/a one with fewer. Shaded: control.}
\label{tab:trainseeds}
\begin{center}\footnotesize
\setlength{\tabcolsep}{4pt}
\begin{tabular}{llr>{\hspace{-2\tabcolsep}${}\pm{}$}lr>{\hspace{-2\tabcolsep}${}\pm{}$}lr>{\hspace{-2\tabcolsep}${}\pm{}$}l}
\toprule
Student & Condition & \multicolumn{2}{c}{$\Delta a$} & \multicolumn{2}{c}{$\Delta\tau$} & \multicolumn{2}{c}{$\Delta$ acc.} \\
\midrule
\rowcolor{black!10}\multicolumn{8}{l}{\textbf{Criterion distillation}} \\
Qwen3-8B & distilled & $+0.09$ & $0.08$ & {\boldmath$-4.44$} & {\boldmath$0.69$} & {\boldmath$+0.033$} & {\boldmath$0.018$} \\
\rowcolor{black!4} & random direction & $+0.02$ & $0.01$ & $+0.04$ & $0.05$ & $+0.003$ & $0.004$ \\
\arrayrulecolor{black!25}\specialrule{0.3pt}{1pt}{1pt}\arrayrulecolor{black}
Qwen3.5-9B & distilled & $-0.01$ & $0.06$ & {\boldmath$-1.02$} & {\boldmath$0.22$} & {\boldmath$+0.045$} & {\boldmath$0.011$} \\
\rowcolor{black!4} & random direction & $+0.00$ & $0.02$ & $+0.20$ & $0.06$ & $-0.009$ & $0.005$ \\
\arrayrulecolor{black!25}\specialrule{0.3pt}{1pt}{1pt}\arrayrulecolor{black}
gemma-3-12b & distilled & $-0.02$ & $0.04$ & {\boldmath$-1.86$} & {\boldmath$0.98$} & {\boldmath$+0.014$} & {\boldmath$0.012$} \\
\rowcolor{black!4} & random direction & $+0.00$ & $0.01$ & $-0.11$ & $0.36$ & $+0.001$ & $0.004$ \\
\arrayrulecolor{black!25}\specialrule{0.3pt}{1pt}{1pt}\arrayrulecolor{black}
Mistral-24B & distilled & $-0.17$ & $0.10$ & {\boldmath$-3.29$} & {\boldmath$0.83$} & {\boldmath$+0.036$} & {\boldmath$0.016$} \\
\rowcolor{black!4} & random direction & $+0.07$ & $0.05$ & $-0.10$ & $0.23$ & $+0.001$ & $0.005$ \\
\rowcolor{black!10}\multicolumn{8}{l}{\textbf{Sensitivity learning}} \\
Qwen3-8B & trained & {\boldmath$+0.63$} & {\boldmath$0.12$} & {\boldmath$-2.35$} & {\boldmath$0.53$} & {\boldmath$+0.172$} & {\boldmath$0.018$} \\
\rowcolor{black!4} & shuffled labels & $-0.19$ & $0.11$ & $+0.07$ & $1.10$\rlap{$^{\ddagger}$} & $-0.044$ & $0.020$ \\
\arrayrulecolor{black!25}\specialrule{0.3pt}{1pt}{1pt}\arrayrulecolor{black}
Qwen3.5-9B & trained & {\boldmath$+0.48$} & {\boldmath$0.14$} & {\boldmath$-1.69$} & {\boldmath$0.30$} & {\boldmath$+0.175$} & {\boldmath$0.016$} \\
\rowcolor{black!4} & shuffled labels & $-0.30$ & $0.19$ & $-3.08$ & $0.85$\rlap{$^{\ddagger}$} & $-0.038$ & $0.019$ \\
\arrayrulecolor{black!25}\specialrule{0.3pt}{1pt}{1pt}\arrayrulecolor{black}
gemma-3-12b & trained & {\boldmath$+1.19$} & {\boldmath$0.21$} & {\boldmath$-2.57$} & {\boldmath$1.01$} & {\boldmath$+0.238$} & {\boldmath$0.018$} \\
\rowcolor{black!4} & shuffled labels & $-0.20$ & $0.11$\rlap{$^{\ddagger}$} & \multicolumn{2}{c}{n/a} & $-0.070$ & $0.020$ \\
\arrayrulecolor{black!25}\specialrule{0.3pt}{1pt}{1pt}\arrayrulecolor{black}
Mistral-24B & trained & {\boldmath$+0.57$} & {\boldmath$0.16$} & {\boldmath$-3.39$} & {\boldmath$0.73$} & {\boldmath$+0.206$} & {\boldmath$0.020$} \\
\rowcolor{black!4} & shuffled labels & $-0.61$ & $0.46$ & \multicolumn{2}{c}{n/a} & $-0.059$ & $0.019$ \\
\bottomrule
\end{tabular}
\end{center}
\end{table}

\paragraph{Transfer.} Table~\ref{tab:transfer} lists the values behind Figure~\ref{fig:transfer}: the
criterion shift on each external benchmark for each distilled student and for its random-direction
control.

\begin{table}[!ht]
\caption{Values behind Figure~\ref{fig:transfer}: criterion shift on six external benchmarks for the
distilled student and its random-direction control, seed mean $\pm$ paired error. Negative is toward
acting. Shaded: control.}
\label{tab:transfer}
\begin{center}\footnotesize
\setlength{\tabcolsep}{5pt}
\begin{tabular}{llcccc}
\toprule
 & Benchmark & Qwen3-8B & Qwen3.5-9B & gemma-3-12b & Mistral-24B \\
\midrule
\rowcolor{black!10}\multicolumn{6}{l}{\textbf{Distilled student}} \\
Act now  & When2Call      & $-0.05 \pm 0.01$ & $-0.06 \pm 0.01$ & $-0.04 \pm 0.01$ & $-0.07 \pm 0.01$ \\
         & QuestBench     & $-1.97 \pm 0.05$ & $-0.55 \pm 0.03$ & $-0.82 \pm 0.03$ & $-0.86 \pm 0.04$ \\
         & When2Tool      & $-0.58 \pm 0.03$ & $-0.34 \pm 0.03$ & $-0.32 \pm 0.10$ & $-0.52 \pm 0.11$ \\
         & BFCL-miss-func & $-0.21 \pm 0.05$ & --               & $-0.05 \pm 0.03$ & $-0.14 \pm 0.04$ \\
\arrayrulecolor{black!25}\specialrule{0.3pt}{1pt}{1pt}\arrayrulecolor{black}
Planning & ACPBench       & $-0.81 \pm 0.12$ & $-0.31 \pm 0.10$ & $-0.21 \pm 0.07$ & $-0.32 \pm 0.10$ \\
         & CaT-Bench      & $-0.58 \pm 0.04$ & $-0.38 \pm 0.04$ & $-0.13 \pm 0.01$ & $-0.35 \pm 0.02$ \\
\rowcolor{black!10}\multicolumn{6}{l}{\textbf{Random-direction control}} \\
\rowcolor{black!4}Act now  & When2Call      & $+0.00 \pm 0.01$ & $+0.01 \pm 0.00$ & $-0.01 \pm 0.00$ & $-0.01 \pm 0.00$ \\
\rowcolor{black!4}         & QuestBench     & $+0.05 \pm 0.01$ & $+0.09 \pm 0.01$ & $-0.05 \pm 0.01$ & $+0.00 \pm 0.02$ \\
\rowcolor{black!4}         & When2Tool      & $+0.02 \pm 0.01$ & $+0.06 \pm 0.01$ & $+0.00 \pm 0.01$ & $+0.15 \pm 0.09$ \\
\rowcolor{black!4}         & BFCL-miss-func & $+0.02 \pm 0.03$ & --               & $+0.01 \pm 0.02$ & $-0.01 \pm 0.02$ \\
\arrayrulecolor{black!25}\specialrule{0.3pt}{1pt}{1pt}\arrayrulecolor{black}
\rowcolor{black!4}Planning & ACPBench       & $+0.01 \pm 0.03$ & $+0.06 \pm 0.03$ & $-0.01 \pm 0.02$ & $+0.00 \pm 0.00$ \\
\rowcolor{black!4}         & CaT-Bench      & $+0.01 \pm 0.01$ & $+0.08 \pm 0.02$ & $-0.01 \pm 0.00$ & $+0.03 \pm 0.01$ \\
\bottomrule
\end{tabular}
\end{center}
{\footnotesize The dash marks Qwen3.5-9B on BFCL-miss-func, whose base readout does not reproduce.}
\end{table}

\paragraph{Accuracy.} On all four students, criterion distillation raises balanced accuracy on the
external benchmarks where the student starts conservative (Table~\ref{tab:acc_full}). Where it starts
permissive, the same shift overshoots: accuracy falls by up to four points or stays within its error,
except on Mistral-24B on When2Tool. There the hit rate is already at ceiling, so a shift toward acting can
only add false alarms, and accuracy falls by about ten points.

\begin{table}[!ht]
\caption{Change in balanced accuracy after criterion distillation, seed means, grouped by the starting
criterion on each benchmark. Qwen3.5-9B on BFCL-miss-func is left out, as in Figure~\ref{fig:transfer}.}
\label{tab:acc_full}
\begin{center}\footnotesize
\setlength{\tabcolsep}{6pt}
\begin{tabular}{llcc}
\toprule
Student & Benchmark & Start $c$ & $\Delta$ bal.\ acc. \\
\midrule
\rowcolor{black!10}\multicolumn{4}{l}{\textbf{Started conservative ($c>0$)}} \\
Qwen3-8B & When2Tool & $+0.42$ & $+0.014$ \\
 & QuestBench & $+0.71$ & $+0.041$ \\
 & ACPBench & $+0.40$ & $+0.025$ \\
\arrayrulecolor{black!25}\specialrule{0.3pt}{1pt}{1pt}\arrayrulecolor{black}
Qwen3.5-9B & When2Tool & $+0.73$ & $+0.046$ \\
 & ACPBench & $+0.65$ & $+0.052$ \\
\arrayrulecolor{black!25}\specialrule{0.3pt}{1pt}{1pt}\arrayrulecolor{black}
gemma-3-12b & QuestBench & $+0.91$ & $+0.062$ \\
 & ACPBench & $+0.31$ & $+0.012$ \\
\arrayrulecolor{black!25}\specialrule{0.3pt}{1pt}{1pt}\arrayrulecolor{black}
Mistral-24B & QuestBench & $+1.38$ & $+0.050$ \\
 & BFCL-miss-func & $+0.39$ & $+0.022$ \\
 & ACPBench & $+0.62$ & $+0.046$ \\
\rowcolor{black!10}\multicolumn{4}{l}{\textbf{Started permissive ($c<0$)}} \\
Qwen3-8B & When2Call & $-0.26$ & $-0.010$ \\
 & BFCL-miss-func & $-0.57$ & $-0.040$ \\
 & CaT-Bench & $-1.13$ & $-0.042$ \\
\arrayrulecolor{black!25}\specialrule{0.3pt}{1pt}{1pt}\arrayrulecolor{black}
Qwen3.5-9B & When2Call & $-0.52$ & $-0.017$ \\
 & QuestBench & $-0.06$ & $-0.011$ \\
 & CaT-Bench & $-1.43$ & $-0.023$ \\
\arrayrulecolor{black!25}\specialrule{0.3pt}{1pt}{1pt}\arrayrulecolor{black}
gemma-3-12b & When2Call & $-0.37$ & $-0.001$ \\
 & BFCL-miss-func & $-0.23$ & $+0.011$ \\
 & When2Tool & $-1.13$ & $-0.028$ \\
 & CaT-Bench & $-0.59$ & $-0.015$ \\
\arrayrulecolor{black!25}\specialrule{0.3pt}{1pt}{1pt}\arrayrulecolor{black}
Mistral-24B & When2Call & $-0.04$ & $-0.003$ \\
 & When2Tool & $-1.44$ & $-0.095$ \\
 & CaT-Bench & $-0.86$ & $-0.026$ \\
\bottomrule
\end{tabular}
\end{center}
\end{table}

\subsection{A criterion-free readout of the external benchmarks}
\label{app:readout}

Distillation keeps sensitivity on a readout that carries no criterion term (\S\ref{sec:transfer}). The external $d'$ is the
equal-variance estimate from one operating point, and it is exact only when the two evidence
distributions share a variance. Scoring each example of every base model, distilled student and control by its
answer-token log-probability margin instead gives a full ROC and an area under it (AUC) with no criterion term.
Thresholding this margin recovers the binary answers and, on the base models, the $d'$ and $c$ of the
main readout. BFCL-miss-func has no answer-token margin and is absent.

Distillation leaves the AUC unchanged for every student on the five benchmarks with answer
log-probabilities (Table~\ref{tab:readout_all}).
Each distilled change is equivalent to no change under a two one-sided test at $\pm0.02$, with a $90\%$
paired cluster-bootstrap interval inside that bound.

On Qwen3-8B the change in $d'$ is the criterion sliding along an unequal-variance ROC. Across
benchmarks the size of its criterion shift tracks the change in single-point $d'$ but not the change in
AUC (Table~\ref{tab:readout}). On QuestBench, where the shift is largest, $d'$ rises by $0.72$ and the
base model's fitted ROC slope is $1.42$, against one under equal variance.

\begin{table}[!ht]
\caption{Change in criterion-free AUC after criterion distillation and for its random-direction control,
seed mean against the base model: the values of Table~\ref{tab:auc} with the control beside them.}
\label{tab:readout_all}
\begin{center}\footnotesize
\setlength{\tabcolsep}{4pt}
\begin{tabular}{@{}lrrrrrrrr@{}}
\toprule
& \multicolumn{2}{c}{Qwen3-8B} & \multicolumn{2}{c}{Qwen3.5-9B} & \multicolumn{2}{c}{gemma-3-12b} & \multicolumn{2}{c@{}}{Mistral-24B} \\
\cmidrule(lr){2-3}\cmidrule(lr){4-5}\cmidrule(lr){6-7}\cmidrule(l){8-9}
Benchmark & \multicolumn{1}{c}{distilled} & \multicolumn{1}{c}{control} & \multicolumn{1}{c}{distilled} & \multicolumn{1}{c}{control} & \multicolumn{1}{c}{distilled} & \multicolumn{1}{c}{control} & \multicolumn{1}{c}{distilled} & \multicolumn{1}{c@{}}{control} \\
\midrule
When2Call & $-0.008$ & $0.000$ & $-0.006$ & $+0.001$ & $+0.001$ & $+0.001$ & $-0.002$ & $0.000$ \\
When2Tool & $+0.004$ & $-0.001$ & $-0.001$ & $0.000$ & $+0.011$ & $+0.003$ & $+0.001$ & $0.000$ \\
QuestBench & $-0.012$ & $0.000$ & $-0.002$ & $0.000$ & $-0.005$ & $-0.001$ & $+0.002$ & $0.000$ \\
ACPBench & $+0.001$ & $+0.001$ & $+0.001$ & $-0.001$ & $+0.004$ & $+0.001$ & $-0.003$ & $0.000$ \\
CaT-Bench & $+0.001$ & $0.000$ & $+0.008$ & $-0.001$ & $-0.003$ & $-0.001$ & $+0.002$ & $+0.001$ \\
\bottomrule
\end{tabular}
\end{center}
\end{table}

\begin{table}[!ht]
\caption{Criterion distillation on Qwen3-8B, seed mean against the base model: size of the criterion
shift, change in single-point $d'$ and in criterion-free AUC, and the base model's fitted ROC slope $s$.
Rows are ordered by the size of the criterion shift.}
\label{tab:readout}
\begin{center}\footnotesize
\setlength{\tabcolsep}{8pt}
\begin{tabular}{@{}lcccc@{}}
\toprule
Benchmark & $|\Delta c|$ & $\Delta d'$ & $\Delta\mathrm{AUC}$ & $s$ \\
\midrule
QuestBench & $1.99$ & $+0.720$ & $-0.0117$ & $1.42$ \\
ACPBench & $0.82$ & $+0.162$ & $+0.0007$ & $0.91$ \\
CaT-Bench & $0.59$ & $-0.142$ & $+0.0013$ & $0.80$ \\
When2Tool & $0.58$ & $-0.010$ & $+0.0037$ & $1.15$ \\
When2Call & $0.05$ & $-0.067$ & $-0.0077$ & $1.00$ \\
\bottomrule
\end{tabular}
\end{center}
{\footnotesize QuestBench, $95\%$ paired cluster-bootstrap intervals: $\Delta d'$ $[+0.537,+0.931]$,
$\Delta\mathrm{AUC}$ $[-0.0164,-0.0068]$.}
\end{table}

\subsection{Answer preference and a Yes-bias teacher}
\label{app:yesbias}

\textsc{FlipAct} reads the decision to act from a Yes or No, so on it a shift toward acting and a shift
toward answering Yes coincide. Reversed wordings and a teacher that only favors Yes separate the two
(\S\ref{sec:transfer}).

\paragraph{Reversed wordings.} Each distilled student and its control answer the held-out questions under
the canonical wording and under the three reversed wordings of App.~\ref{app:polarity}, with answers
recoded so that every rate is a rate of acting. A shift in the readiness to act raises the act-rate under
every wording. A preference for Yes raises it under the canonical wording and lowers it under the
reversed ones. Table~\ref{tab:yesbias_pol} splits the change in act-rate into a shared part, the mean of
the canonical and reversed changes, and a flipping part, half their difference. On all four students
most of the shift flips with the polarity, so on \textsc{FlipAct}'s binary readout distillation mostly
shifts the preference for Yes. Qwen3-8B and Mistral-24B also carry a smaller shared part, and the
random-direction controls move neither part.

\begin{table}[!ht]
\caption{Change in the held-out act-rate after criterion distillation under the canonical wording and
under the three reversed wordings (their mean), and its shared and flipping parts, as seed means.}
\label{tab:yesbias_pol}
\begin{center}\footnotesize
\setlength{\tabcolsep}{7pt}
\begin{tabular}{@{}lcccc@{}}
\toprule
& \multicolumn{2}{c}{Change in act-rate} & \multicolumn{2}{c@{}}{Parts} \\
\cmidrule(lr){2-3}\cmidrule(l){4-5}
Student & canonical & reversed & shared & flipping \\
\midrule
Qwen3-8B & $+0.30$ & $-0.11$ & $+0.09$ & $+0.20$ \\
Qwen3.5-9B & $+0.07$ & $-0.03$ & $+0.02$ & $+0.05$ \\
gemma-3-12b & $+0.09$ & $-0.11$ & $-0.01$ & $+0.10$ \\
Mistral-24B & $+0.21$ & $-0.03$ & $+0.09$ & $+0.12$ \\
\bottomrule
\end{tabular}
\end{center}
\end{table}

\paragraph{A Yes-bias teacher.} A teacher that only favors Yes tests whether distillation carries more
than a Yes preference. This teacher is the base model with the logits of every Yes token raised by a
constant, chosen on the training half so that it answers Yes as often as the steered teacher. The data
and recipe are unchanged, and the students are Qwen3-8B and gemma-3-12b. On the four Yes/No benchmarks
the Yes-bias student moves $c$ as far as the distilled one or further. On the two formats without a Yes answer, the criterion-distilled student moves $c$ further in
every seed on both students (Table~\ref{tab:yesbias_ext}), though by little in one Qwen3-8B seed on
BFCL-miss-func. BFCL-miss-func is the stronger test of the two because its reply is free-form, while
When2Call keeps option letters as answer tokens.

\begin{table}[!ht]
\caption{Criterion shift on the two benchmarks without a Yes answer for the distilled, random-direction
and Yes-bias students, and the difference between the distilled and the Yes-bias student, paired on
examples, overall and per seed. Negative is toward acting.}
\label{tab:yesbias_ext}
\begin{center}\footnotesize
\setlength{\tabcolsep}{3pt}
\begin{tabular}{@{}llcccr@{${}\pm{}$}lccc@{}}
\toprule
& & \multicolumn{3}{c}{$\Delta c$, seed mean} & \multicolumn{5}{c@{}}{distilled $-$ Yes-bias} \\
\cmidrule(lr){3-5}\cmidrule(l){6-10}
Student & Benchmark & distilled & random & Yes-bias & \multicolumn{2}{c}{mean} & seed 0 & seed 1 & seed 2 \\
\midrule
Qwen3-8B & When2Call & $-0.051$ & $+0.004$ & $-0.026$ & $-0.025$ & $0.005$ & $-0.024$ & $-0.018$ & $-0.031$ \\
& BFCL-miss-func & $-0.208$ & $+0.017$ & $-0.159$ & $-0.050$ & $0.025$ & $-0.072$ & $-0.072$ & $-0.005$ \\
\arrayrulecolor{black!25}\specialrule{0.3pt}{1pt}{1pt}\arrayrulecolor{black}
gemma-3-12b & When2Call & $-0.043$ & $-0.006$ & $-0.006$ & $-0.037$ & $0.006$ & $-0.047$ & $-0.026$ & $-0.037$ \\
& BFCL-miss-func & $-0.053$ & $+0.008$ & $+0.021$ & $-0.074$ & $0.022$ & $-0.113$ & $-0.069$ & $-0.040$ \\
\bottomrule
\end{tabular}
\end{center}
\end{table}

On Qwen3-8B the Yes-bias student also moves $c$ on both formats, by about half of the distilled shift on
When2Call and three quarters on BFCL-miss-func, while on gemma-3-12b it does not. On some models, then,
training toward one answer on binary questions reshapes a broader readiness to act. We conjecture that
this is one way in which fine-tuning on binary preferences or rewards, as in RLHF
\citep{ouyang2022training} or GRPO \citep{shao2024deepseekmath}, shifts when a model acts. Testing it is
left to future work.

\subsection{Choosing the steering sign and strength per benchmark}
\label{app:calib}

Steering with a fixed sign toward acting (the fixed push) lowers balanced accuracy on $12$ of the $25$ model--benchmark pairs
formed by the five steered models of \S\ref{sec:steer} and five benchmarks (Table~\ref{tab:calib}). The
rule below, summarized in \S\ref{sec:transfer}, instead chooses the sign and strength of the steering push
from a few labeled examples of the target benchmark.

The push is chosen on $K=50$ labeled examples of the target benchmark, whole pairs where the benchmark
has pairs, and read on the rest, over twenty random splits. On these fifty examples alone, the rule
keeps the strengths at which $d'$ stays within a quarter of its unsteered value and takes the sign of the
unsteered criterion. It pushes nothing if that criterion is smaller than its bootstrap error. Otherwise it
takes the kept strength that brings $c$ closest to zero along a monotone path, and it pushes nothing if
none qualifies. The same rule sets a scalar bias on the answer-token log-odds where the benchmark has an
answer token. The readout is the change in balanced accuracy against no intervention, with a paired
bootstrap over examples ($B=2000$). A model--benchmark pair counts as improved when the median gain
exceeds two errors and as degraded when it falls below minus one error, a stricter bar for a gain than
for a loss.

\begin{table}[!ht]
\caption{Choosing the push on fifty examples of each benchmark. (a) Outcomes over model--benchmark
pairs, with the fixed push for comparison (shaded). (b) Median change in balanced accuracy over twenty splits for
the per-benchmark steering push and for an answer-token bias chosen by the same rule.}
\label{tab:calib}
\newcommand{\calnone}{{\hphantom{$+0$}$\cdot$\hphantom{$000$}}}
\begin{center}\footnotesize
\setlength{\tabcolsep}{6pt}
\begin{tabular}{lcccc}
\toprule
(a) Condition & Pairs & Improved & Unchanged & Degraded \\
\midrule
\rowcolor{black!4}{Fixed push toward acting} & $25$ & $2$ & $11$ & $12$ \\
Per-benchmark push & $25$ & $5$ & $19$ & $1$ \\
Answer-token bias & $20$ & $5$ & $15$ & $0$ \\
\bottomrule
\end{tabular}

\vspace{0.8em}

\setlength{\tabcolsep}{6pt}
\begin{tabular}{llccccc@{\hspace{1em}}}
\toprule
(b) Model & & When2Call & When2Tool & BFCL-miss-func & ACPBench & CaT-Bench \\
\midrule
Qwen3-8B & push & \calnone & $+0.022$\rlap{\,$\bullet$} & $+0.021$ & $+0.021$ & $+0.006$ \\
 & bias & \calnone & $+0.024$ & -- & $+0.021$ & $+0.011$ \\
\arrayrulecolor{black!25}\specialrule{0.3pt}{1pt}{1pt}\arrayrulecolor{black}
Qwen3.5-27B & push & \calnone & \calnone & \calnone & \calnone & \calnone \\
 & bias & \calnone & $+0.024$ & -- & $+0.011$ & $+0.086$\rlap{\,$\bullet$} \\
\arrayrulecolor{black!25}\specialrule{0.3pt}{1pt}{1pt}\arrayrulecolor{black}
gemma-2-9b & push & $+0.000$ & $+0.113$\rlap{\,$\bullet$} & $+0.000$ & $+0.000$ & \calnone \\
 & bias & $+0.031$ & $+0.075$\rlap{\,$\bullet$} & -- & $+0.000$ & $+0.000$ \\
\arrayrulecolor{black!25}\specialrule{0.3pt}{1pt}{1pt}\arrayrulecolor{black}
Mistral-24B & push & $-0.020$\rlap{\,$\circ$} & \calnone & $+0.049$\rlap{\,$\bullet$} & \calnone & \calnone \\
 & bias & $+0.005$ & $+0.034$\rlap{\,$\bullet$} & -- & $+0.087$\rlap{\,$\bullet$} & \calnone \\
\arrayrulecolor{black!25}\specialrule{0.3pt}{1pt}{1pt}\arrayrulecolor{black}
phi-4 & push & \calnone & $+0.084$\rlap{\,$\bullet$} & $+0.047$\rlap{\,$\bullet$} & $+0.024$ & \calnone \\
 & bias & $+0.000$ & $+0.107$\rlap{\,$\bullet$} & -- & $+0.027$ & \calnone \\
\bottomrule
\end{tabular}
\end{center}
{\footnotesize $\bullet$ improved, $\circ$ degraded, $\cdot$ none chosen, -- no answer token.}
\end{table}

The per-benchmark push degrades one pair, Mistral-24B on When2Call, and repairs every other pair the
fixed push degraded (Table~\ref{tab:calib}). On that pair it moves the criterion the right way but loses
$17\%$ of $d'$, within the quarter the rule allows. Where it pushes, its sign matches the sign of the
full-benchmark criterion in every split of every pair whose criterion is at least $0.3$ in size. Five
pairs improve, and eight when the labeled set grows to a hundred examples.
On Qwen3.5-27B the direction does not move the criterion toward zero on four benchmarks, so the rule
pushes nothing there. The answer-token bias is never worse than no intervention and matches or beats the
per-benchmark push on all but one of its pairs. Only the steering push applies to BFCL-miss-func, which has
no answer-token margin.

\subsection{Qualitative examples}
\label{app:examples}

The boxes show single answers of Qwen3-8B, the one model that is both steered in Table~\ref{tab:steer}
and a distillation student. Boxes (a) and (b) illustrate steering (\S\ref{sec:steer}) on When2Call, where
the model picks one of four candidate responses and ``call'' means it picks the tool call. Boxes (c) and
(d) illustrate how distillation keeps sensitivity (\S\ref{sec:transfer}). Each shows a matched pair,
one question that warrants acting and one that does not, under the three seeds of the
criterion-distilled student and of the random-direction control. There $m$ is the Yes minus No logit at
the answer token, averaged over seeds, and the answer is Yes when $m>0$. Distillation raises both
margins, so $m=0$ can fall between them while their order, the comparison the criterion-free
AUC counts, stays as it was. Boxes (a) and (b) show the first example, in sorted id order, that fits the
pattern in its title; boxes (c) and (d) each show one of the pairs the distilled student separates while its
control does not. The line under
each box counts how often the pattern occurs, and Table~\ref{tab:examples_counts} gives the pair counts
behind (c) and (d).

\begin{table}[!ht]
\caption{Pair counts behind boxes (c) and (d): how often the question that warrants acting keeps the
higher margin, and how the distilled student answers the pairs the base model answers No to twice. Shaded: base
model and control.}
\label{tab:examples_counts}
\begin{center}\footnotesize
\setlength{\tabcolsep}{8pt}
\begin{tabular}{lcc}
\toprule
& \textsc{FlipAct} (c) & QuestBench (d) \\
\midrule
Pairs & $354$ & $288$ \\
\rowcolor{black!10}\multicolumn{3}{l}{\textbf{Question that warrants acting above its twin}} \\
\rowcolor{black!4}\quad{base model} & $69\%$ & $92\%$ \\
\quad distilled student & $68$--$69\%$ & $93$--$94\%$ \\
\rowcolor{black!4}\quad{random-direction control} & $68$--$69\%$ & $92\%$ \\
\arrayrulecolor{black!25}\specialrule{0.3pt}{1pt}{1pt}\arrayrulecolor{black}
Pairs the base model answers No to twice & $189$ & $187$ \\
\quad distilled student separates them & $50$ & $117$ \\
\quad distilled student answers Yes to both & $98$ & $59$ \\
\bottomrule
\end{tabular}
\end{center}
\end{table}

\providecommand{\exY}{\textcolor{teal}{Yes}}
\providecommand{\exN}{No}
\providecommand{\exC}{\textcolor{teal}{call}}
\providecommand{\exW}{withhold}
\tcbset{exbox/.style={enhanced,colback=black!2,colframe=black!35,boxrule=0.4pt,arc=1.5pt,left=4pt,
  right=4pt,top=3pt,bottom=3pt,fontupper=\small,coltitle=black,colbacktitle=black!8,
  fonttitle=\small\bfseries,toptitle=1pt,bottomtitle=1pt,before skip=6pt,after skip=6pt},
  exq/.style={enhanced,frame hidden,colback=black!5,boxrule=0pt,arc=1pt,left=3pt,right=3pt,top=2pt,
  bottom=2pt,fontupper=\footnotesize,before skip=0pt,after skip=3pt}}

\begin{tcolorbox}[exbox,title={(a) Steering on a request that needs a tool: When2Call, gold call, unsteered withholds}]
\begin{tcolorbox}[exq]
Tools: \texttt{flight.status.check} (airlines, person\_name, ticket\_number); \texttt{uber.eat.order}
(restaurant, items, quantities).\\
User: ``Hi, I am Karan. Could you tell me the current status of my flight with Delta? My name is Karan
Sharma, and my ticket number is IND4567.''
\end{tcolorbox}
Gold: \exC{} (\texttt{flight.status.check}) \quad$\cdot$\quad Unsteered: \exW{} \quad$\cdot$\quad
Steered, $\lambda^\star=16$: \exC{}\\[1pt]
{\footnotesize The unsteered model withholds on $56$ of the $300$ call examples. Steering calls on $9$ of
them and withholds on $2$ calls it had made.}
\end{tcolorbox}

\begin{tcolorbox}[exbox,title={(b) The matching false alarm: When2Call, gold asks for a missing detail, unsteered withholds}]
\begin{tcolorbox}[exq]
Tools: \texttt{get\_sensor\_readings\_latest}, \texttt{get\_sensor\_readings\_history}, and
\texttt{get\_sensor\_readings\_history\_by\_interval} (networkIds, metrics, timespan, interval, \ldots).\\
User: ``Obtain the sensor reading for temperature for my network L\_579838452023959405.''
\end{tcolorbox}
Gold: \exW{} (ask for the timespan) \quad$\cdot$\quad Unsteered: \exW{} \quad$\cdot$\quad Steered,
$\lambda^\star=16$: \exC{} (\texttt{get\_sensor\_readings\_history\_by\_interval} with a timespan the
user did not give)\\[1pt]
{\footnotesize The unsteered model withholds on $210$ of the $300$ withhold examples. Steering calls on
$21$ of them ($8$ that ask for a detail, $13$ where no listed tool fits) and withholds on $1$ it had
called. Qwen3-8B starts slightly permissive on When2Call, so these false alarms outnumber the hits
of (a) and accuracy falls (Table~\ref{tab:steer}).}
\end{tcolorbox}

\begin{tcolorbox}[exbox,title={(c) Distillation moves the criterion, not the order: a held-out \textsc{FlipAct} state flip}]
\begin{tcolorbox}[exq]
Warranted ($\Delta U=+2$): Situation: The food or contents is very cold. The supplies this tool depends
on (mains electricity) are available nearby. The goal is to bring the food or contents back to a normal
state. Action: a person uses a hotplate. Is using the hotplate a wise choice in this situation? Answer
Yes or No.\\[1pt]
Destructive ($\Delta U=-2$): the same question with the food or contents ``very hot''.
\end{tcolorbox}
{\centering\footnotesize
\begin{tabular}{lcccc}
 & Gold & Base $m$ & Distilled $m$ & Control $m$ \\
Warranted   & \exY{} & $-6.0$ (\exN{})  & $+14.3$ (\exY{}, 3/3) & $-4.6$ (\exN{}, 3/3) \\
Destructive & \exN{} & $-33.0$ (\exN{}) & $-14.3$ (\exN{}, 3/3) & $-33.2$ (\exN{}, 3/3) \\
\end{tabular}\par}
{\footnotesize Over the $354$ held-out state flips both margins rise in every flip and seed, and the
warranted question stays above the unwarranted one about as often as before (Table~\ref{tab:examples_counts}).
Hits and false alarms rise together: of the held-out questions the base model answers No, the distilled student
answers Yes in all seeds on $220$ of $286$ warranted ones and on $137$ of $311$ destructive ones, and the
control on none.}
\end{tcolorbox}

\begin{tcolorbox}[exbox,title={(d) The same on an external benchmark: a QuestBench GSM-Q-verbal pair}]
\begin{tcolorbox}[exq]
Fully specified: ``My wife and I went out for dinner at a restaurant that was offering a 10\% discount on
their meals. After applying the discount, our dinner bill came to \$50. We decided to evenly split the
check, but she wanted me to pay an additional 20\% tip on our discounted bill. How much did I end up
paying? Do you already have enough information to give a single correct final answer to this question
right now? Do NOT solve it. Answer with exactly one word: Yes or No.''\\[1pt]
Underspecified: the same problem without the discount and the \$50 bill.
\end{tcolorbox}
{\centering\footnotesize
\begin{tabular}{lcccc}
 & Gold & Base $m$ & Distilled $m$ & Control $m$ \\
Fully specified & \exY{} & $-0.2$ (\exN{})  & $+6.4$ (\exY{}, 3/3)  & $-0.5$ (\exN{}, 3/3) \\
Underspecified  & \exN{} & $-20.8$ (\exN{}) & $-10.2$ (\exN{}, 3/3) & $-20.6$ (\exN{}, 3/3) \\
\end{tabular}\par}
{\footnotesize Over the $288$ pairs both margins rise in every pair and seed, and the fully specified
problem stays above its twin about as often as before (Table~\ref{tab:examples_counts}).}
\end{tcolorbox}

\section{A model that returns a probability}
\label{app:jev}

Every entry in the paper is a token model read through a Yes/No answer, so its criterion can include a
preference for one answer word (\S\ref{sec:criterion}). JEV \citep{typesafe2026docs} is
a decision model of a different kind.\footnote{Model \texttt{jev-latest}, API version \texttt{jev-1.13.0},
evaluated 2026-09-22.} Given a state and a statement about it, JEV returns the probability that the
statement is true and generates no text, so it gives a readout of \textsc{FlipAct} with no answer word to
prefer. The state is a question's situation and action sentences, and the statement is ``Using the
\emph{tool} is a wise choice in this situation''. The returned probability is read as $P(\text{act})$ and
binarized at one half for the fit of Eq.~\ref{eq:fit}. Because this input differs from the prompt the
token models receive, JEV is not entered in Figure~\ref{fig:map}, the panel or Table~\ref{tab:coords}, and
the comparisons below are descriptive.

\paragraph{Coordinates.} On the statement readout JEV has $a=0.62$ and $\tau=+1.35$ (Table~\ref{tab:jev}).
Its sensitivity is close to that of the most sensitive entries of Table~\ref{tab:coords}. The criterion is conservative, with an interval
that excludes zero, and lies between those of Llama-3.3-70B and gemma-4-31b. Mean $P(\text{act})$ steps
between $\Delta U=-1$ and $0$ and is flat on either side, the same profile as the token models in
Figure~\ref{fig:response}. Asked instead to rate, on an ordered verbal scale, the action's effect on the
situation or how warranted it is, JEV tracks the size of the evidence barely better than its probability does.
When the statement is reversed, JEV's answer preference in the decomposition of App.~\ref{app:polarity} is
$b=-0.20\pm0.24$, and the probabilities under the two polarities sum to $1.01$ on average.

\paragraph{Readout format.} Asked to choose between use and avoid instead of judging the statement, JEV
returns the same coordinates (Table~\ref{tab:jev}), so on \textsc{FlipAct} its criterion does not depend on
how the binary decision is read out. It does depend on which decisions are offered. On When2Call, where
many withhold examples call for asking a question, JEV has $c=-1.15$ when it judges a statement. Offered a
choice among call, ask and decline, it routes most of those examples to ask, and $c$ moves to $+0.02$. A
criterion read from two options is therefore not a constant of the model when a third option exists, which
is one reason an absolute $\tau$ need not predict a benchmark's $c$.

\begin{table}[!ht]
\caption{JEV on \textsc{FlipAct} under two readouts of the same decision, which give the same coordinates
and the same criterion-free AUC.}
\label{tab:jev}
\begin{center}\footnotesize
\setlength{\tabcolsep}{7pt}
\begin{tabular}{@{}lccccc@{}}
\toprule
Readout & $a$ & 95\% interval & $\tau$ & 95\% interval & AUC \\
\midrule
Statement, binarized at one half & $0.62$ & $[0.47,0.80]$ & $+1.35$ & $[+0.83,+2.02]$ & $0.748$ \\
Choice between use and avoid     & $0.61$ &                & $+1.29$ &                  & $0.750$ \\
\bottomrule
\end{tabular}
\end{center}
{\footnotesize AUC is that of $P(\text{act})$ against the computed label. Intervals resample situations and
are given for the statement readout.}
\end{table}

\paragraph{Caveats.} JEV is one closed model under one API version, and its responses are deterministic,
so every interval resamples situations or examples only. Binarizing its probability also adds a threshold the
token models do not have. JEV shows that a conservative criterion and a step at the sign of the evidence
can arise without an answer word, but a single model cannot stand in for the $44$ models.

\section{When-to-act benchmarks, steering and efficient inference}
\label{app:related-table}

Table~\ref{tab:related-benchmarks} places twelve when-to-act benchmarks on the design choices of
\S\ref{sec:related}: how the ground truth is constructed, whether controlled negatives are matched, whether
the evidence for acting is graded, and whether responses are decomposed into sensitivity and criterion.
The decisions cover tool-call necessity \citep{ross2025when2call, patil2025bfcl, huang2024metatool,
when2tool2026}, acting or abstaining on minimally perturbed pairs \citep{agentabstain2026}, answering or
abstaining \citep{abstentionbench2025}, clarification \citep{questbench2025, knowing2026}, over-refusal
\citep{rottger2024xstest, cui2024orbench, shi2024overkill} and physical affordance \citep{bisk2020piqa}.

Activation steering reads a direction from the residual stream and adds it back
\citep{rimsky2024steering, zou2023representation, turner2023activation, li2023inference}. Refusal is its
best-studied target \citep{arditi2024refusal}, and sycophancy is a related disposition
\citep{sharma2024sycophancy, perez2023discovering}. App.~\ref{app:s5figs} projects the refusal and
sycophancy directions out of the criterion direction as a control on the steering result.

The decision to act also appears inside efficient inference, where the action is to spend more
computation. LoHi \citep{dong2026lohi} answers from a low-resolution video stream first and requests
high-resolution frames only when the entropy of the answer exceeds a threshold, which sends about a
third of the questions to the second pass at nearly the accuracy of always requesting them; the
threshold is a criterion in the sense used here. MMTok \citep{dong2026mmtok} observes that the number
of visual tokens a question needs depends on its difficulty, and MAI \citep{dong2026mai}
passes a small model's reasoning to a large one only when needed. The last of these
also shows the failure this paper measures: asked whether a fan suits a user who feels cold, a small
vision-language model states that a fan cools and still answers that it is suitable.

\begin{table}[!ht]
\caption{When-to-act benchmarks on the design choices of \S\ref{sec:related}: \cmark\ full, $\sim$ partial,
\bmark\ absent.}
\label{tab:related-benchmarks}
\begin{center}\footnotesize
\setlength{\tabcolsep}{2.5pt}
\begin{tabular}{@{}lllccc@{}}
\toprule
Benchmark & Decision & Ground truth & Ctrl.\ neg. & Graded & Sens./crit. \\
\midrule
When2Call~\citep{ross2025when2call}         & tool call/abstain  & LLM+human      & $\sim$ & \bmark & \bmark \\
BFCL~\citep{patil2025bfcl}                 & function applies?  & human          & $\sim$ & \bmark & \bmark \\
MetaTool~\citep{huang2024metatool}          & tool needed?       & human/template & $\sim$ & \bmark & \bmark \\
When2Tool~\citep{when2tool2026}             & call vs.\ no-call  & env-computed   & \cmark & $\sim$\,{\scriptsize(diff.)} & \bmark \\
\citet{shi2026tocall}                       & tool call/abstain  & LLM judge      & \bmark & $\sim$\,{\scriptsize(SAE)} & $\sim$\,{\scriptsize(offset)} \\
\arrayrulecolor{black!25}\specialrule{0.3pt}{1pt}{1pt}\arrayrulecolor{black}
AgentAbstain~\citep{agentabstain2026}       & act vs.\ abstain   & pipeline+judge & \cmark & \bmark & $\sim$\,{\scriptsize(paired)} \\
AbstentionBench~\citep{abstentionbench2025} & answer/abstain     & curated+judge  & \bmark & \bmark & \bmark \\
\arrayrulecolor{black!25}\specialrule{0.3pt}{1pt}{1pt}\arrayrulecolor{black}
QuestBench~\citep{questbench2025}           & ask vs.\ solve     & rule (CSP)     & \cmark & \bmark & \bmark \\
\citet{knowing2026}                         & recognize/clarify  & human          & \cmark & \bmark & $\sim$\,{\scriptsize(informal)} \\
\arrayrulecolor{black!25}\specialrule{0.3pt}{1pt}{1pt}\arrayrulecolor{black}
XSTest~\citep{rottger2024xstest}            & comply/refuse      & hand-written   & \cmark & \bmark & \bmark \\
OR-Bench~\citep{cui2024orbench}             & comply/refuse      & LLM ensemble   & $\sim$ & \bmark & \bmark \\
\arrayrulecolor{black!25}\specialrule{0.3pt}{1pt}{1pt}\arrayrulecolor{black}
PIQA~\citep{bisk2020piqa}                   & physical solution  & human          & $\sim$ & \bmark & \bmark \\
\midrule
\textsc{FlipAct} & act vs.\ withhold & by rule & \cmark\,{\scriptsize(2$\times$2)} & \cmark\,{\scriptsize($\Delta U$)} & \cmark\,{\scriptsize($a,\tau$)} \\
\bottomrule
\end{tabular}
\end{center}
\end{table}

\section{Limitations and outlook}
\label{app:limits}
\label{sec:discussion}

\paragraph{Instrument.} Every label is computed by rule from a tool's direction and magnitude on one
axis, annotated by language models with no person in the loop, so some questions will carry labels a
person would set otherwise. Agreement between annotator families and perturbations of the annotations
check how far such errors could move the results (App.~\ref{app:synth:probe} and~\ref{app:noise}); no
human observers were measured on the diagnostic. All questions share one template in one physical domain.
The criterion stays positive in nearly all models under five rephrasings of it
(App.~\ref{app:polarity}); other templates and domains are untested. Side effects, annotated for part
of the tool records, do not enter the rule.

\paragraph{Measurement.} Several relations rest on few models. The fall of the criterion with size inside
a family rests on three to five models per family. The \textsc{FlipAct} criterion does not track a
benchmark's $c$ consistently: near zero on three benchmarks, positive on BFCL on the strength of the three
models with the lowest $\tau$, and negative on When2Call at a looser panel cutoff
(Table~\ref{tab:benchmarks}, App.~\ref{app:check}), over eighteen models whose criteria sit in a narrow
band. Whether this comes from limited power, from the benchmarks' interfaces or from the construct is
open. All models are open checkpoints of at most $70$B parameters.

\paragraph{Interventions.} Two of the intervention models, Qwen3.5-9B and Qwen3.5-27B, answer No under
either wording, so part of their starting $\tau$ is a preference for the answer token
(App.~\ref{app:polarity}). The steering direction depends on the wording it is read under, so the
interventions are read from behavior, not from the geometry of the direction (App.~\ref{app:s5figs}).
After distillation the equivalence test on sensitivity is undecided on Qwen3-8B and Mistral-24B. The
Yes-bias comparison covers two students, with BFCL-miss-func as its only free-form test
(App.~\ref{app:yesbias}). The reasoning-mode comparison rests mostly on Qwen models. Sensitivity learning
raises $a$ on \textsc{FlipAct} but does not transfer stably to the external benchmarks: its changes in
$d'$ and $c$ there differ in sign across students and are matched by the shuffled-label control.
\textsc{FlipAct} tests basic physical knowledge of tools, and labels on it appear to add little that
sharpens the decision to call a tool elsewhere.

\paragraph{Outlook.} The build scales with annotator calls (\S\ref{sec:pipeline}), and its schema is
written in a domain-specific language \citep{chen2026better} in which a property, an axis or a rule can be added or removed
without rebuilding the pipeline. Richer tool properties such as side effects, rules over several
properties and their interactions \citep{zhanglogicif}, and new axes, templates and domains run through the same steps.
Beyond single decisions, a richer graph of how tools transform states could chain questions into
multi-step agent scenarios, in which a model's tendencies can be read across a sequence of decisions.
Such scenarios may also supply what sensitivity learning needs to transfer beyond \textsc{FlipAct}.

\end{document}